%% file: main.tex
\documentclass{article}

\usepackage[nonatbib, final]{neurips_2024}

\usepackage[utf8]{inputenc} 
\usepackage[T1]{fontenc}    
\usepackage{xcolor}         

\usepackage[colorlinks=true,linkcolor=red!90!black, citecolor=blue!50!black,pagebackref]{hyperref}%
\renewcommand*{\backref}[1]{}
\renewcommand*{\backrefalt}[4]{
    \ifcase #1
          \or [Cited on page~#2.]
          \else [Cited on pages~#2.]
    \fi
}

\usepackage{url}            
\usepackage{booktabs}       
\usepackage{amsfonts}       
\usepackage{nicefrac}       
\usepackage{microtype}      
\usepackage{tcolorbox}

\usepackage[toc,page,header]{appendix}
\usepackage{wrapfig}
\usepackage{tikz}
\usepackage{subcaption}
\usepackage{marvosym}

\usetikzlibrary{bayesnet, calc, arrows.meta, bending}
\usetikzlibrary{decorations.pathreplacing,angles,quotes,shapes.multipart, calligraphy}

\usepackage{amsmath}

\usepackage{enumerate}

\usepackage{amssymb}
\usepackage{mathtools}
\usepackage{amsthm}
\usepackage[capitalize,noabbrev]{cleveref}
\usepackage{minitoc}

\usepackage{fontawesome5}
\usepackage{graphicx, graphbox}

\theoremstyle{plain}
\newtheorem{theorem}{Theorem}

\newtheorem{lemma}{Lemma}
\newtheorem{corollary}[theorem]{Corollary}
\theoremstyle{definition}
\newtheorem{definition}{Definition}
\newtheorem{assumption}{Assumption}
\theoremstyle{remark}
\newtheorem{remark}{Remark}
\usepackage{printlen}

\usepackage[inline]{enumitem}

\usepackage{algorithm}
\usepackage{algorithmic}

\usepackage{booktabs}
\usepackage{multirow}

\newenvironment{sproof}{\proof}{\endproof}

\input{math_commands.tex}

\title{Identifiable Object-Centric Representation Learning via Probabilistic Slot Attention}

\author{%
  Avinash Kori$^{1*}$ 
  \And
  Francesco Locatello$^2$ 
  \And
  Ainkaran Santhirasekaram$^1$ 
  \AND
  Francesca Toni$^1$ 
  \And
  Ben Glocker$^1$ 
  \And 
Fabio De Sousa Ribeiro$^{1}\thanks{Equal Contribution.}$ \\
  \AND
  \\
  $^1$ Imperial College London, UK \\
  $^2$ Institute of Science and Technology, Austria \\
  \texttt{a.kori21@imperial.ac.uk}
}

\begin{document}
\maketitle

\begin{abstract}
Learning modular object-centric representations is crucial for systematic generalization. 
Existing methods show promising object-binding capabilities empirically, but theoretical identifiability guarantees remain relatively underdeveloped. Understanding when object-centric representations can theoretically be identified is crucial for scaling slot-based methods to high-dimensional images with correctness guarantees.
To that end, we propose a probabilistic slot-attention algorithm that imposes an \textit{aggregate} mixture prior over object-centric slot representations, thereby providing slot identifiability guarantees without supervision, up to an equivalence relation. We provide empirical verification of our theoretical identifiability result using both simple 2-dimensional data and high-resolution imaging datasets.
\end{abstract}

\input{sections/1_introduction}
\input{sections/2_related_works}
\input{sections/3_background}
\input{sections/4_gmm}
\input{sections/5_identifiability_results}
\input{sections/6_experiments}

\input{sections/7_conclusion}

\section*{Acknowledgements}
A. Kori is supported by UKRI (grant number EP/S023356/1), as part of the UKRI Centre for Doctoral Training in Safe and Trusted AI. B. Glocker and F.D.S. Ribeiro acknowledge the support of the UKRI AI programme, and the Engineering and Physical Sciences Research Council, for CHAI - EPSRC Causality in Healthcare AI Hub (grant number EP/Y028856/1).

\bibliography{main}
\bibliographystyle{plain}


\appendix
\include{sections/appendix}

\end{document}

%% file: math_commands.tex
\usepackage{amsmath,amsfonts,bm}

\def\eqref#1{equation~\ref{#1}}

\def\1{\bm{1}}

\def\bmu{\boldsymbol{\mu}}
\def\bsigma{\boldsymbol{\sigma}}

\def\bpi{\boldsymbol{\pi}}

\def\rvc{{\mathbf{c}}}

\def\rvu{{\mathbf{i}}}

\def\rvk{{\mathbf{k}}}

\def\rvq{{\mathbf{q}}}

\def\rvs{{\mathbf{s}}}

\def\rvu{{\mathbf{u}}}
\def\rvv{{\mathbf{v}}}

\def\rvx{{\mathbf{x}}}
\def\rvy{{\mathbf{y}}}
\def\rvz{{\mathbf{z}}}

\def\mA{{\bm{A}}}

\def\mH{{\bm{H}}}

\def\mP{{\bm{P}}}

\def\mW{{\bm{W}}}

\DeclareMathAlphabet{\mathsfit}{\encodingdefault}{\sfdefault}{m}{sl}
\SetMathAlphabet{\mathsfit}{bold}{\encodingdefault}{\sfdefault}{bx}{n}

\def\gB{{\mathcal{B}}}

\def\gN{{\mathcal{N}}}

\def\gP{{\mathcal{P}}}

\def\gS{{\mathcal{S}}}

\def\gX{{\mathcal{X}}}

\def\gZ{{\mathcal{Z}}}

\def\bSigma{\boldsymbol{\Sigma}}



%% file: sections/1_introduction.tex
\section{Introduction}
It has been hypothesized that developing machine learning (ML) systems capable of human-level understanding requires imbuing them with notions of \textit{objectness}~\cite{lake2017building,scholkopf2022statistical}. 
Objectness notions can be characterised as physical, abstract, semantic, geometric, or via spaces and boundaries~\cite {yuille2006vision,epstein2017cognitive}. 
Humans can generalise across environments with few examples to learn from~\cite{tenenbaum2011grow}, and this has been attributed to our ability to segregate percepts into object entities~\cite{rock1973orientation, hinton1979some,kulkarni2015deep,behrens2018cognitive}.

Obtaining object-centric representations is deemed to be a key step for achieving true compositional generalization~\cite{bengio2013representation,lake2017building, battaglia2018relational, greff2020binding}, and uncovering causal influence between discrete concepts and their environment~\cite{marcus2003algebraic, gerstenberg2021counterfactual, gopnik2004theory, scholkopf2022statistical,behrens2018cognitive}. Significant progress in learning object-centric representations has been made~\cite{engelcke2019genesis, engelcke2021genesis, kori2023grounded, seitzer2022bridging, chang2022object, singh2021illiterate}, particularly in unsupervised object discovery settings using an iterative attention mechanism known as Slot Attention (SA)~\cite{locatello2020object}. However, most existing work approaches object-centric representation learning empirically, leaving theoretical understanding relatively underdeveloped. Establishing the \textit{identifiability}~\cite{hyvarinen1999nonlinear,hyvarinen2000independent} of representations is important as it clarifies under which conditions object-centric representation learning is theoretically possible~\cite{brady2023provably}. 

A well-known result shows that identifiability of latent variables is fundamentally impossible without assumptions about the data generating process~\cite{hyvarinen1999nonlinear,locatello2019challenging}. \textit{Therefore, understanding when object representations can theoretically be identified is important to scale object-centric methods to high-dimensional images.} Recent works \cite{brady2023provably, lachapelle2024additive} make important advances on this by explicitly
\begin{wraptable}[13]{r}{6.5cm}
    \centering
    \footnotesize
    \caption{
    Identifiability strategies (mixing function $f$ or latent dist. $p(\mathbf{z})$), and assumptions made by object-centric learning methods.}
    \label{tab:OCL_identifiability}
    \begin{tabular}{lcc}
     \toprule
     \textsc{Method} & \textsc{Assumption} & \textsc{Identif.}  \\ 
     \midrule
     \begin{tabular}[c]{@{}l@{}}
      \scriptsize\cite{burgess2019monet, emami2022slot, engelcke2019genesis, greff2019multi, locatello2020object, wang2023slot} 
      \end{tabular} &
      \hypersetup{linkcolor=red}
      \footnotesize\ref{ass:beta_disentanglement}, 
      \footnotesize\ref{ass:additive_decoder} 
      &
      N/A
      \\
      \midrule
      \begin{tabular}[c]{@{}l@{}}
        \scriptsize\cite{elsayed2022savi++,kipf2021conditional, lin2020space, van2020investigating}
      \end{tabular} &
      \hypersetup{linkcolor=red}
      \footnotesize\ref{ass:beta_disentanglement}, 
      \footnotesize\ref{ass:additive_decoder},
      \footnotesize\ref{ass:aux_data}
      &
      N/A
      \\
      \midrule
     \scriptsize\cite{kori2023grounded} & 
     \hypersetup{linkcolor=red}
     \footnotesize\ref{ass:beta_disentanglement},
     \footnotesize\ref{ass:additive_decoder},
     \footnotesize\ref{ass:object_sufficiency} & N/A
     \\
     \midrule
     
     \scriptsize\cite{brady2023provably}&
     \hypersetup{linkcolor=red}
     \footnotesize\ref{ass:beta_disentanglement},
     \footnotesize\ref{ass:additive_decoder},
     \footnotesize\ref{ass:irreducibility},
     \footnotesize\ref{definition:comp},
     \footnotesize\ref{ass:decoder_injectivity}
     & 
     $f$
     \\
     \midrule
     \textbf{Proposed} &
     \hypersetup{linkcolor=red}
     \footnotesize\ref{ass:beta_disentanglement}, 
     \footnotesize\ref{ass:weak_inj} &
     $p(\mathbf{z})$
     \\\bottomrule
    \end{tabular}
\end{wraptable}
stating the set of assumptions necessary for providing theoretical identifiability of object-centric representations. However, they restrict their attention to properties of the \textit{mixing function}, studying a class of models with \textit{additive} decoders. Although there are merits to this approach, there are practical challenges with the so-called \textit{compositional contrast} objective~\cite{brady2023provably}, as it involves computing Jacobians and requires second-order optimization via gradient descent. Consequently, satisfying the identifiability conditions explicitly (e.g. compositional contrast must be zero) is computationally restrictive for moderately high-dimensional data. In this work, we present a probabilistic perspective that is not subject to the same scalability issues while still providing theoretical identifiability of object-centric representations without supervision. In Table~\ref{tab:OCL_identifiability}, we list object-centric learning methods, their (sometimes implicit) modelling assumptions (see \S~\ref{sec:identifiability} for additional information and Appendix~\ref{appendix:assumptions} for a detailed breakdown and discussion), and their respective identifiability guarantees of object representations. Most methods do not guarantee identifiability, and make the $\gB$-disentanglement (\ref{ass:beta_disentanglement}) and \textit{additive decoder} (\ref{ass:additive_decoder}) assumptions. Brady \emph{et. al.}~\cite{brady2023provably} do provide identifiability guarantees and additionally assume \textit{irreducibility}~(\ref{ass:irreducibility}) and \textit{compositionality}~(\ref{definition:comp}). Our method provides identifiability guarantees by introducing latent structure (i.e. via a GMM prior) which generalizes to non-additive decoders. 
This is advantageous as the computational complexity of additive decoders scales linearly with the number of slots $K$ -- whereas our approach is invariant to $K$. 
Moreover, non-additive decoders have been found to significantly improve performance in practice~\cite{seitzer2022bridging,singh2021illiterate,singh2022simple}, though the theoretical basis is underexplored.
Finally, latent structure can reduce the complexity burden on the mixing function $f$ (decoder), making it easier to learn in practice~\cite{falck2021multi,kivva2022identifiability}.

\textbf{Contributions. \,} Our main contributions are the following:
\begin{enumerate*}[label=\textbf{(\roman*)}]
    \item We prove that object-centric representations (i.e. slots) are identifiable without supervision up to an equivalence relation (\S~\ref{sec:identifiability}) under a latent mixture model specification. To that end, we propose a probabilistic slot-attention algorithm (\S~\ref{sec:expectation_maximization}) which imposes an \textit{aggregate} mixture prior over slot representations.
    \item We show that our approach induces a non-degenerate (\textit{global}) Gaussian Mixture Model (GMM) by aggregating per-datapoint (\textit{local}) GMMs, providing a slot prior which: (a) is empirically stable across runs (i.e. identifiable up to affine transformations and slot permutations); (b) can be tractably sampled from.
    \item We provide conclusive empirical evidence of our theoretical object-centric identifiability result, including visual verification on synthetic 2-dimensional data as well as standard imaging benchmarks (\S~\ref{sec:experiments}).
\end{enumerate*}

%% file: sections/2_related_works.tex
\section{Related Work}
\paragraph{Identifiable Representation Learning.} Identifiability of representations stems from early work in independent component analysis (ICA)~\cite{hyvarinen1999nonlinear,hyvarinen2000independent}, and is making a resurgence recently~\cite{hyvarinen2016unsupervised,hyvarinen2019nonlinear,locatello2019challenging,khemakhem2020variational,von2021self,lachapelle2024nonparametric,yao2024multi}. Common strategies for tackling this identifiability problem are: (i) restricting the class of mixing functions; (ii) using non-i.i.d data, interventional data or counterfactuals; and (iii) imposing structure in the latent space via distributional assumptions. Regarding (i), restricting the class of the mixing functions to conformal maps~\cite{buchholz2022function} or volume-preserving transformations~\cite{yang2022nonlinear} has been found to produce identifiable models. For (ii), prior works~\cite{zimmermann2021contrastive, locatello2020weakly, brehmer2022weakly, ahuja2022interventional, von2021self} assume access to contrastive pairs of observations $(\rvx, \tilde{\rvx})$ obtained from either data augmentation, interventions, or approximate counterfactual inference. As for (iii), latent space structure is enforced via either: (a) using auxiliary variables to make latent variables conditionally independent~\cite{pmlr-v89-hyvarinen19a,khemakhem2020variational,Khemakhem2020_ice}; or (b) distributional assumptions such as placing a mixture prior over the latent variables in a VAE~\cite{dilokthanakul2016deep,willetts2021don,kivva2022identifiability}. In this work, we prove an identifiability result via strategy (iii) but within an object-centric learning context, where the latent variables are a set of object \textit{slots}~\cite{locatello2020object}.

\textbf{Object-Centric Learning. \,} Much early work on unsupervised representation learning is based on the Variational Autoencoder (VAE) framework~\cite{kingma2013auto}, and relies on independence assumptions between latent variables to learn so-called \textit{disentangled} representations~\cite{bengio2013representation,higgins2017betavae,pmlr-v80-kim18b,eastwood2018a,pmlr-v97-mathieu19a}. These methods are closely linked to object-centric representation learning~\cite{burgess2019monet, engelcke2019genesis, greff2019multi}, as they also leverage (iterative) variational inference procedures~\cite{marino2018iterative,van2020investigating, lin2020space}. Alternatively, an iterative attention mechanism known as slot attention (SA)~\cite{locatello2020object} has been the focus of much follow-up work recently~\cite{engelcke2021genesis, singh2021illiterate, wang2023slot, singh2022neural, emami2022slot}. 
\begin{figure*}[!t]
    \centering
    \includegraphics[trim=0 0 0 0,clip,width=.95\textwidth]{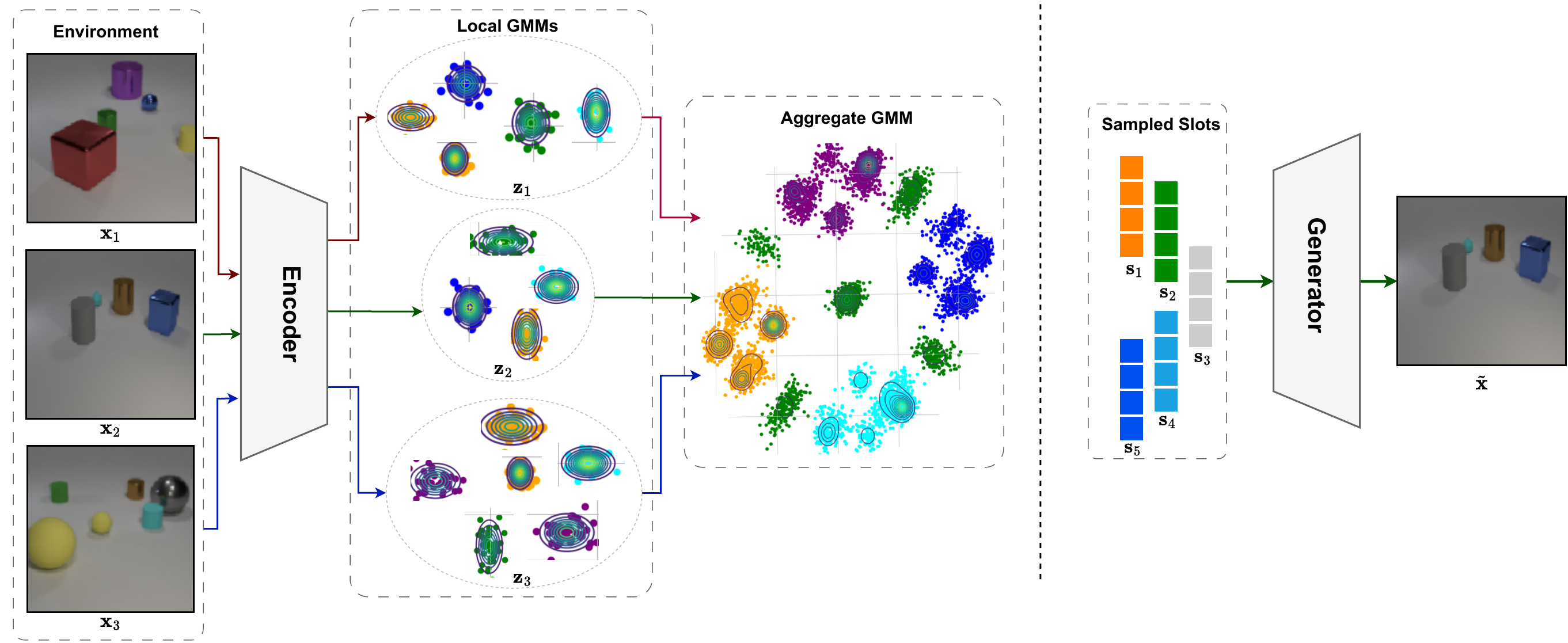}
    \caption{\textbf{Probabilistic slot attention and the identifiable aggregate slot posterior}. (Left) Slot posterior GMMs per datapoint ({local}) and the induced \textit{aggregate} posterior GMM ({global}). (Right) Sampling slot representations from the aggregate slot posterior is tractable.}
    \label{fig:overview}
\end{figure*}
Although slot attention-based methods show promising object \textit{binding}~\cite{greff2020binding} capabilities empirically on select datasets, they do not provide identifiability guarantees on the learned representations. 
Recently,~\cite{mansouri2022object, mansouri2023object} assume the access to interventional data-generating process following~\cite{ahuja2022weakly} to demonstrate the identifiability of object-centric representations, while 
~\cite{brady2023provably, lachapelle2024additive} presented the identifiability results for object representations (i.e. slots), clarifying the necessary assumptions and properties of the \textit{mixing function}
(e.g. additive decoders). 
However, satisfying Brady \emph{et. al.}~\cite{brady2023provably}'s \textit{compositional contrast} identifiability condition explicitly (must be zero) requires computationally restrictive second-order optimization. In contrast, we shift the focus to learning structured object-centric latent spaces via Probabilistic Slot Attention (PSA), bridging the gap between generative model identifiability literature and object-centric representation learning. Notably, our PSA approach is also related to probabilistic capsule routing~\cite{hinton2018matrix,NEURIPS2020_47fd3c87,ribeiro2020capsule,ribeiro2022learning} since slots are equivalent to \textit{universal} capsules~\cite{hinton2022represent}, but like slot attention, offers output permutation symmetry and does not face scalability issues.

%% file: sections/3_background.tex
\section{Background}
\label{sec:background}
\paragraph{Notation.} Let $\gX \subseteq \mathbb{R}^{H \times W \times C}$ denote the input image space, where each image $\mathbf{x}$ is of size $H \times W$ pixels with $C$ channels. 
Let $f_{e}: \gX \rightarrow \gZ$ denote an encoder mapping image space to a latent space $\gZ \subseteq \mathbb{R}^{N \times d}$, where each latent variable $\mathbf{z}$ consists of $N$, $d$-dimensional vectors. Lastly, let $f_{d}: \gS \rightarrow \gX$ denote a decoder mapping from slot representation space $\gS \subseteq \mathbb{R}^{K \times d}$ to image space. 

\paragraph{Slot Attention.} Slot attention~\cite{locatello2020object} receives a set of feature embeddings $\rvz \in \mathbb{R}^{N \times d}$ per input $\mathbf{x}$, and applies an iterative attention mechanism to produce $K$ object-centric representations called slots $\rvs \in \mathbb{R}^{K \times d}$. 
Let $\mW_k, \mW_v$ denote \textit{key} and \textit{value} transformation matrices acting on $\rvz$, and $\mW_q$ the \textit{query} transformation matrix acting on $\mathbf{s}$. To simplify our exposition later on, let $f_s:\gZ \times \gS \rightarrow \gS$  be shorthand notation for the \textit{slot update} function, defined as:
\begin{align}
    &&\rvs^{t+1} \coloneqq f_s(\rvz, \rvs^t) = \hat{\mA}\rvv, && \ \hat{A}_{ij} \coloneqq \frac{A_{ij}}{\sum_{l=1}^N A_{il}}, &&
    \mA \coloneqq \text{softmax}\left(\frac{\rvq \rvk^T}{\sqrt{d}}\right) \in \mathbb{R}^{K \times N}, &&
    \label{eqn:slot_attention_update}
\end{align}
where $\rvq = \mW_q\rvs^t \in \mathbb{R}^{K \times d}$, $\rvk = \mW_k\rvz \in \mathbb{R}^{N \times d}$, and $\rvv=\mW_v\rvz \in \mathbb{R}^{N \times d}$ correspond to the query, key and value vectors respectively and $\mA \in \mathbb{R}^{K \times N}$ is the attention matrix. Unlike self-attention \cite{vaswani2017attention}, the queries $\mathbf{q}$ in slot attention are a function of the slots $\mathbf{s}^t$, and are iteratively refined over $T$ iterations. The initial slots $\mathbf{s}^{t{=}0}$ are randomly sampled from a standard Gaussian. The queries at iteration $t$ are given by $\hat{\rvq}^t = \mW_q \rvs^t$, and the slot update process can be summarized as in Equation \ref{eqn:slot_attention_update}. 

\paragraph{Compositionality.} Compositionality as defined by Brady et al.~\cite{brady2023provably} is a structure imposed on the slot decoder mapping $f_d$ which implies that each image pixel is a function of at most one slot representation, thereby enforcing a local sparsity structure on the Jacobian matrix of $f_d$. 
%
\begin{definition}[Compositional Contrast]
For a differentiable mapping $f_d:\gZ  \rightarrow \gX$, the compositional contrast of $f_d$ at $\mathbf{z}$ is given by:
\begin{align}
    C_{\text{comp}}(f_d, \rvz) = \sum_{n=0}^N \sum_{k=1}^K \sum_{j=k+1}^K  \left \|\frac{\partial f_d(\rvz)_n}{\partial \rvz_k} \right \| \left \|\frac{\partial f_d(\rvz)_n}{\partial \rvz_j} \right \|. \nonumber
\end{align}
\label{dfn:compositional_contrast}
\vspace{-10pt}
\end{definition}
Brady et al.~\cite{brady2023provably}'s main result (Theorem 1) relies on \textit{compositionality} and \textit{invertibility} of $f_d$ to guarantee slot-identifiability when both the compositional contrast and the reconstruction loss equal zero. However, using $C_{\text{comp}}(f_d, \rvz)$ as a metric or as part of an objective function is computationally prohibitive\footnote{E.g. for a CNN with 500K parameters with batch size 32, $\geq$ 125GB of GPU memory is needed}. Our method aims to minimize $C_{\text{comp}}(f_d, \rvz)$ implicitly~\cite{lachapelle2024additive}, without additive decoders.
\begin{figure*}[!t]
    \hfill
    \centering
    \begin{subfigure}{.375\textwidth}    
    \centering
    \begin{tikzpicture}
        \node[obs] (x) {$\mathbf{x}$};
        \node[det, above=25pt of x] (z) {$\mathbf{z}$};
        \node[latent, right=20pt of z] (s1) {$\mathbf{s}^1$};
        \node[latent, below=25pt of s1] (s0) {$\mathbf{s}^0$};
        \node[latent, draw=none, right=20pt of s0] (n) {$\mathcal{N}(0,\mathbf{I})$};
        \node[latent, right=20pt of s1] (s2) {$\mathbf{s}^2$};
        \node[latent, right=25pt of s2] (st) {$\mathbf{s}^T$};
        \node[draw=none, rectangle, inner sep=0pt, right=2pt of s2] (s3) {\footnotesize\hspace{1.3pt}$\dots$};
        \edge{x}{z}
        \edge{z}{s1}
        \edge{s0}{s1}
        \draw [->] (z) to [out=45,in=135] (st);
        \draw [->] (z) to [out=45,in=135] (s2);
        \draw [-] (s2) to [out=0,in=180] (s3);
        \edge{s1}{s2}
        \edge{s3}{st}
        \edge{n}{s0}
    \end{tikzpicture} 
    \caption{Slot Attention Encoder}
    \label{fig:sa_encoder}
    \end{subfigure}
    \hfill
    \begin{subfigure}{.32\textwidth}    
    \centering
    \begin{tikzpicture}
        \node[latent] (z) {$\mathbf{z}$};
        \node[latent, above=of z] (c) {$u$}; 
        \node[latent, draw=none, left=25pt of c] (mu){$\boldsymbol{\mu}$};
        \node[latent, draw=none, left=25pt of z] (sigma){$\boldsymbol{\sigma}$};
        \node[rectangle, draw=none, right=25pt of c] (pi){$\boldsymbol{\pi}$};
        \node[obs, right=25pt of z] (x) {$\mathbf{x}$};
        \plate[] {kplate} {(mu)(sigma)}{$K$};
        \plate[] {nplate} {(c)(z)}{$N$};
        \plate[inner sep=8pt] {mplate} {(mu)(sigma)(c)(z)(x)(pi)}{$M$};
        \draw[->, shorten <=7pt] (mu.center) -- (z);
        \draw[->, shorten <=7pt] (sigma.center) -- (z);
        \edge{c}{z}
        \edge{z}{x}
        \edge{pi}{c}
        \draw [densely dashed,->] (x) to [out=135,in=45] (z);
    \end{tikzpicture}
    \caption{Probabilistic Slot Attention}
    \label{fig:local_gmm}
    \end{subfigure}
    \hfill
    \begin{subfigure}{.29\textwidth}
    \centering
    \begin{tikzpicture}
        \node[obs] (z) {$\mathbf{z}$};
        \node[latent, above=of z] (c) {$u$}; 
        \node[latent, draw=none, left=25pt of c] (mu){$\widehat{\boldsymbol{\mu}}$};
        \node[latent, draw=none, left=25pt of z] (sigma){$\widehat{\boldsymbol{\sigma}}$};
        \node[rectangle, draw=none, right=25pt of c] (pi){$\widehat{\boldsymbol{\pi}}$};
        \plate[] {kplate} {(mu)(sigma)}{$MK$};
        \plate[] {nplate} {(c)(z)}{$N$};
        \plate[draw=none, inner sep=8pt] {mplate} {(mu)(sigma)(c)(z)(x)(pi)}{\color{white}{$M$}};
        \draw[->, shorten <=7pt] (mu.center) -- (z);
        \draw[->, shorten <=7pt] (sigma.center) -- (z);
        \edge{c}{z}
        \edge{pi}{c}
    \end{tikzpicture}
    \caption{Aggregate Posterior Mixture}
    \label{fig:global_gmm}
    \end{subfigure}
    \caption{\textbf{Graphical models of probabilistic slot attention}. \textbf{(a)} Stochastic encoder of standard slot attention~\cite{locatello2020object} with $T$ attention iterations. \textbf{(b)} Proposed model -- each image in the dataset $\{\mathbf{x}_i\}_{i=1}^M$ is encoded into a respective latent representation $\mathbf{z} \in \mathbb{R}^{N\times d}$, to which a (local) Gaussian mixture model with $K$ components is fit via expectation maximisation. The resulting $K$ Gaussians serve as slot posterior distributions: $\mathbf{s}_k \sim \mathcal{N}(\mathbf{s}_k; \boldsymbol{\mu}_k, \boldsymbol{\sigma}_k^2)$, for $k=1,\dots,K$. \textbf{(c)} Aggregate posterior distribution obtained by marginalizing out the data: $q(\mathbf{z}) = \sum_{i=1}^M q(\mathbf{z} \mid \mathbf{x}_i)/M$. We prove that $q(\mathbf{z})$ is a tractable, non-degenerate Gaussian mixture distribution which: (i) serves as the theoretically optimal prior over slots; (ii) is empirically stable across runs (i.e. identifiable up to an affine transformation and slot permutation); (iii) can be tractably sampled from and used for scene composition tasks.
    }
    \hfill
    \label{fig:all_pgms}
\end{figure*}

%% file: sections/4_gmm.tex
\section{Probabilistic Slot Attention}
\label{sec:expectation_maximization}
In this section, we present a probabilistic slot attention framework which imposes a mixture prior structure over the slot latent space. This structure will prove to be instrumental in establishing our main identifiability result in Section~\ref{sec:identifiability}. 
We begin by approaching standard slot attention~\cite{locatello2020object} from a graphical modelling perspective. As shown in Figure~\ref{fig:all_pgms} and explained in Section~\ref{sec:background}, applying slot attention to a deterministic encoding $\mathbf{z} = f_e(\mathbf{x}) \in \mathbb{R}^{N \times d}$ yields a set of $K$ object slot representations $\mathbf{s}_{1:K} \coloneqq \mathbf{s}_1,\dots,\mathbf{s}_K$. This process induces a stochastic encoder $q(\mathbf{s}_{1:K} \mid \mathbf{x})$, where the stochasticity comes from the random initialization of the slots: $\mathbf{s}^{t=0}_{1:K} \sim \mathcal{N}(\mathbf{s}_{1:K}; 0, \mathbf{I}) \in \mathbb{R}^{K \times d}$. Since each slot is a deterministic function of its previous state $\mathbf{s}^{t} \coloneqq f_s(\mathbf{z}, \mathbf{s}^{t-1})$ it is possible to randomly sample initial states $\mathbf{s}^0$ and obtain stochastic estimates of the slots.\footnote{Note that we may use $\mathbf{s}^t$ or $\mathbf{s}(t)$ interchangeably to denote slot representations at slot attention iteration $t$.} However, since each transition depends on $\mathbf{z}$, which in turn depends on the input $\mathbf{x}$, \textit{we do not get a generative model we can tractably sample from}. This can conceivably be remedied by placing a tractable prior over $\mathbf{z}$ and using the VAE framework along the lines of~\cite{pmlr-v202-wang23r}, however, here \textit{we propose an entirely different approach which does not require making additional variational approximations} (see Appendix \ref{appendix:vae_discuss} for further discussion).

\paragraph{Local Slot Mixtures.} Probabilistic slot attention augments standard slot attention by introducing a per-datapoint (i.e. local) Gaussian Mixture Model (GMM) for learning slot distributions. Intuitively, a local GMM can be understood as a way to cluster features within a given image, encouraging the grouping of similar features into object representations. However, unlike regular clustering, here the clustered points are dynamically transformed representations of the actual data. Specifically, we use an encoder function $f_e$ that maps each image $\mathbf{x}_i \in \mathbb{R}^{H \times W \times C}$ in the dataset $\{ \mathbf{x}_i\}_{i=1}^M$, to a latent spatial representation $\mathbf{z}_i \in \mathbb{R}^{N \times d}$. The latent variable $\mathbf{z}$ may be deterministic or stochastic, and we consider the case where $N < HW$ to reflect a modest downscaling with respect to (w.r.t.) $\mathbf{x}$. The goal is to dynamically map each of the $N$, $d$-dimensional vector representations in each $\mathbf{z}$, to one-of-$K$ object slot distributions within a mixture. A \textit{local} GMM can be fit to each posterior latent representation $\mathbf{z} \sim q(\mathbf{z} \mid \mathbf{x}_i)$\footnote{The parametric form of $q$ can be e.g. Gaussian or Dirac delta.} on the fly by maximizing likelihood:
\begin{align}
    p(\rvz \mid \bpi_i, \bmu_i, \bsigma_i) = \prod_{n=1}^N  \sum_{k=1}^K \bpi_{ik} \gN(\rvz_{n}; \bmu_{ik}, \bsigma_{ik}^2),
    \label{eqn:localgmm}
\end{align}
where $\bmu_i = (\bmu_{i1},\dots, \bmu_{iK})$, $\bsigma_i^2= (\bsigma_{i1}^2,\dots, \bsigma_{iK}^2)$ and $\bpi_i = (\pi_{i1},\dots,\pi_{iK})$ are the respective means, diagonal covariances and mixing coefficients of the $i^{\mathrm{th}}$ $K$-component mixture. Figure~\ref{fig:local_gmm} illustrates the resulting probabilistic graphical model (PGM) in more detail.
\begin{wrapfigure}[20]{R}{0.53\textwidth}
\vspace{-24pt}
\begin{minipage}{0.53\textwidth}
\begin{algorithm}[H]
  \caption{Probabilistic Slot Attention}
  \label{alg:the_alg}
  \begin{algorithmic}
    \renewcommand{\baselinestretch}{1.3}\selectfont
    \STATE \textbf{Input:} $\mathbf{z} = f_e(\mathbf{x}) \in \mathbb{R}^{N\times d}$ \hfill {\color{gray}$\triangleright$ representation}
    \STATE $\mathbf{k} \gets \mW_k \mathbf{z} \in \mathbb{R}^{N \times d}$ \hfill {\color{gray}$\triangleright$ compute keys}
    \STATE $\mathbf{v} \gets \mW_v \mathbf{z} \in \mathbb{R}^{N \times d}$ \hfill {\color{gray}$\triangleright$ optional $\mathbf{v} \coloneqq \mathbf{k}$}
    \vspace{3pt}
    \STATE $\forall k$, $\boldsymbol{\pi}(0)_k \gets \frac{1}{K}$, $\boldsymbol{\mu}(0)_k \sim \mathcal{N}(0, \mathbf{I}_d)$, $\boldsymbol{\sigma}(0)_k^2 \gets \mathbf{1}_d$
    \vspace{-8pt}
    \STATE \textbf{for} $t=0,\dots,T-1$
    \vspace{3pt}
    \STATE \quad $\displaystyle{A_{nk} \gets \frac{\boldsymbol{\pi}(t)_k\mathcal{N}\left(\mathbf{k}_n;\mW_q \boldsymbol{\mu}(t)_k, \boldsymbol{\sigma}(t)_k^2\right)}{\sum_{j=1}^{K}\boldsymbol{\pi}(t)_j\mathcal{N}\left(\mathbf{k}_n;\mW_q \boldsymbol{\mu}(t)_j, \boldsymbol{\sigma}(t)_j^2\right)}}$
    \vspace{3pt}
    \STATE \quad $\hat{A}_{nk} \gets A_{nk}/\sum_{l=1}^NA_{lk}$ \hfill {\color{gray}$\triangleright$ normalize}
    \vspace{3pt}
    \STATE \quad $\boldsymbol{\mu}(t+1)_k \gets \sum_{n=1}^N \hat{A}_{nk}\mathbf{v}_n$ \hfill {\color{gray}$\triangleright$ update slots}
    \vspace{3pt}
    \STATE \quad $\boldsymbol{\sigma}(t+1)_k^2 \gets \sum_{n=1}^N \hat{A}_{nk} \left(\mathbf{v}_n - \boldsymbol{\mu}(t+1)_k\right)^2$
    \vspace{3pt}
    \STATE \quad $\boldsymbol{\pi}(t+1)_k \gets \sum_{n=1}^N A_{nk} / N$ \hfill {\color{gray}$\triangleright$ update mixing}
    \vspace{3pt}
    \STATE \textbf{return} $\boldsymbol{\mu}(T), \boldsymbol{\sigma}(T)^2$ \hfill {\color{gray}$\triangleright$ $K$ slot distributions}
  \end{algorithmic}
\end{algorithm}
\end{minipage}
\vspace{-5pt}
\end{wrapfigure}

To maximize the likelihood in Equation~\ref{eqn:localgmm} per datapoint $\mathbf{x}_i$, we present a bespoke expectation-maximisation (EM) algorithm for slot attention, yielding closed-form update equations for the parameters as shown in Algorithm~\ref{alg:the_alg}, and explained next.

\paragraph{Probabilistic Projections.} A powerful property of slot attention and cross-attention more broadly~\cite{vaswani2017attention}, is its ability to decouple the \textit{agreement} mechanism from the representational content. That is, the dot-product is used to measure agreement between each \textit{query} (slot) vector and all the \textit{key} vectors, to dictate how much of each \textit{value} vector (content from $\mathbf{z}_i$) should be represented in each slot's revised representation. To retain this flexibility and decouple the attention computation from the content, we incorporate key-value projections into our probabilistic approach. For brevity, the $i$ subscript is implicit in the following, keeping in mind that these are local quantities (per-datapoint $\mathbf{x}_i$). The parameters of the $K$ Gaussian slot distributions are initialized (at attention iteration $t=0$) as follows:
\begin{align}
    && \forall k, && \boldsymbol{\pi}(0)_k = K^{-1}, &&\boldsymbol{\mu}(0)_k \sim \mathcal{N}(0, \mathbf{I}_d),
    &&\boldsymbol{\sigma}(0)^2_k = \mathbf{1}_d. &&
\end{align}
%
The respective queries $\mathbf{q}$, keys $\mathbf{k}$, and values $\mathbf{v}$ are then given by:
\begin{align}
   && \mathbf{q}(t) = \mW_q \boldsymbol{\mu}(t),&& \mathbf{k} = \mW_k \mathbf{z}, &&\mathbf{v} = \mW_v \mathbf{z},&& 
\end{align}
where $\mathbf{W}_q, \mathbf{W}_k, \mathbf{W}_v \in \mathbb{R}^{d \times d}$, whereas $\mathbf{q}(t)$ denotes the queries at attention iteration $t$. To measure agreement between each input feature (key) and slot (query), we evaluate the normalized probability density of each key under a Gaussian model defined by each slot:
\begin{align}
    &&A_{nk} = \frac{1}{Z}\boldsymbol{\pi}(t)_k\mathcal{N}\left(\mathbf{k}_n; \mathbf{q}(t)_k, \boldsymbol{\sigma}(t)_k^2\right), &&Z = \sum_{j=1}^{K}\boldsymbol{\pi}(t)_j\mathcal{N}\left(\mathbf{k}_n;\mathbf{q}(t)_k, \boldsymbol{\sigma}(t)_j^2\right),&&
\end{align}
where $A_{nk}$ corresponds to the posterior probability that slot (query) $k$ is responsible for input feature (key) $n$. This process yields the slot attention matrix $\mA \in \mathbb{R}^{N \times K}$. As shown in Algorithm~\ref{alg:the_alg}, the mixture parameters $\bpi(t), \bmu(t), \bsigma(t)^2$ are then updated using the attention matrix and the values $\mathbf{v}$. If the values are chosen to be equal to the keys $\mathbf{v} \coloneqq \mathbf{k}$, then the procedure is more in line with standard EM, but the agreement mechanism and the content become entangled. After $T$ probabilistic slot attention iterations, the resulting $K$ Gaussians serve as slot posterior distributions:
\begin{align}
    &&\mathbf{s}(T)_{k} \sim \mathcal{N}\left(\boldsymbol{\mu}(T)_{k}, \boldsymbol{\sigma}(T)_{k}^2\right),  &&\mathrm{for} \ \ k=1,\dots,K,&&
\end{align}
where $\boldsymbol{\mu}(T)$ and $\boldsymbol{\sigma}(T)^2$ are the parameters of all the Gaussians in the mixture given a particular datapoint $\mathbf{x}$. The slots $\mathbf{s}(T)_{1:K}$ are then used for input reconstruction, e.g. by maximizing a (possibly Gaussian) likelihood $p(\mathbf{x} \mid \mathbf{s}(T)_{1:K})$ parameterized by a (possibly additive) decoder $f_d$.

\paragraph{Computational Complexity.}
Probabilistic slot attention (PSA) retains the $\mathcal{O}(TNKd)$ computational complexity of slot attention. The additional operations we introduce for calculating slot mixing coefficients and slot variances (under diagonal slot covariance structure) have complexities of $\mathcal{O}(NK)$ and $\mathcal{O}(NKd)$ respectively, which do not alter the dominant term. When PSA is combined with an additive decoder, it can \textit{lower} computational complexity by eliminating the need to decode inactive slots. In the following, we outline a principled approach for pruning inactive slots.

\paragraph{Automatic Relevance Determination of Slots.}
An open problem in slot-based modelling is the dynamic estimation of the number of slots $K$ needed for each input~\cite{locatello2019challenging,kori2023grounded}. Probabilistic slot attention offers an elegant solution to this problem using the concept of Automatic Relevance Determination (ARD)~\cite{10.5555/525544}. ARD is a statistical framework that prunes irrelevant features by imposing data-driven, sparsity-inducing priors on model parameters to regularize the solution space. Since the output mixing coefficients $\boldsymbol{\pi}(T) \in \mathbb{R}^K$ are input dependent (i.e. local), irrelevant components (slots) will naturally be pruned after $T$ attention iterations, i.e. $\boldsymbol{\pi}(T)_k \to 0$ for any unused slot $k$. We can either use a probability threshold $\tau \in [0, 1)$ to prune unused slots or place a Dirichlet prior over the mixing coefficients to explicitly induce sparsity. For simplicity, we take the former approach: 
\begin{align}
    \rvs_\tau \coloneqq \big\{\rvs(T)_k \mid k \in [K], \bpi(T)_k > \tau \big\},    
\end{align}
where $\rvs_\tau$ denotes the set of \textit{active} slots with mixing coefficient greater than $\tau$, and 
each slot is (optionally) sampled from its Gaussian distribution: 
$\rvs(T)_k \sim \gN\left(\bmu(T)_{k}, \bsigma(T)^2_{k}\right)$.
\begin{figure}[!t]
    \centering
    \includegraphics[trim={10 12 10 10},clip,width=\textwidth]{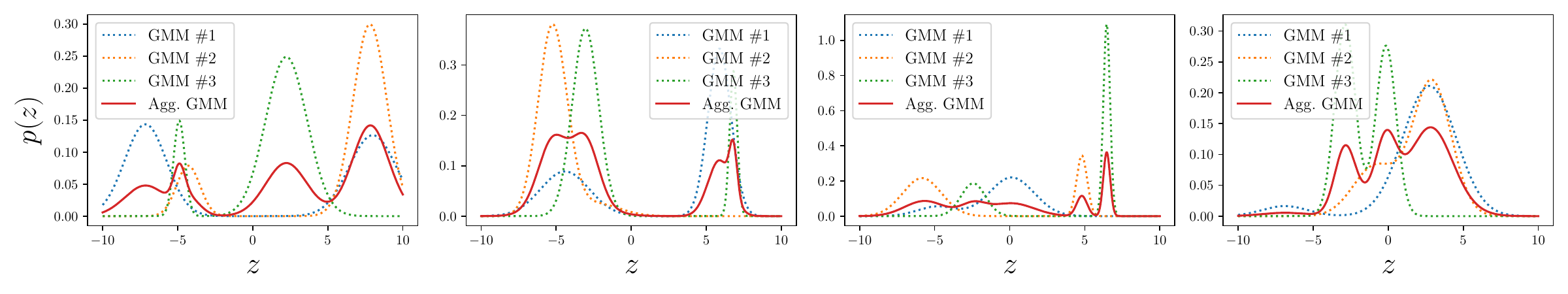}
    \caption{\textbf{Aggregate Gaussian Mixture Density.} Examples of aggregate posterior mixtures. For each plot, we compute the aggregate mixture (red line) based on three random bimodal Gaussian mixtures, and plot the respective densities. The three GMMs here are analogous to the local GMMs obtained from probabilistic slot attention (Algorithm~\ref{alg:the_alg}), and the aggregate GMM represents $q(\mathbf{z})$.}
    \label{fig:agg_post_ex}
\end{figure}
\paragraph{Aggregate Posterior Gaussian Mixture.}
As previously explained, probabilistic slot attention goes beyond standard slot attention by introducing a per-datapoint (i.e. local) GMM to learn \textit{distributions} over slot representations. This imposes structure over the latent space and gives us access to \textit{posterior} slot distributions after the attention iterations. Rather than constraining slot posteriors to be close to a tractable prior -- e.g. via the VAE framework~\cite{kingma2013auto,wang2023slot} which requires further variational approximations -- we leverage our probabilistic setup to compute the optimal (global) prior over slots. 

We propose to compute the \textit{aggregate slot posterior} by marginalizing out the data: $q(\mathbf{z}) = \sum_{i=1}^M q(\mathbf{z} \mid \mathbf{x}_i)/M$, given a pre-trained probabilistic slot attention model (Fig.~\ref{fig:agg_post_ex}). In \S~\ref{sec:identifiability}, we prove that the aggregate posterior is a tractable, non-degenerate Gaussian mixture distribution which:
\begin{enumerate*}[label=\textbf{(\roman*)}]
    \item Serves as the theoretically optimal prior over slots;
    \item Is empirically stable across runs (i.e. identifiable up to an affine transformation and slot permutation, \S~\ref{sec:identifiability});
    \item Can be tractably sampled from and (optionally) used for slot-based scene composition tasks.
\end{enumerate*}
Since GMMs are universal density approximators given enough components (even GMMs with diagonal covariances), the resulting aggregate posterior $q(\mathbf{z})$ is highly flexible and multimodal. It often suffices to approximate it using a sufficiently large subset of the dataset, if marginalizing out the entire dataset becomes computationally restrictive, although we did not observe this to be the case in practice in our set of experiments.

%% file: sections/5_identifiability_results.tex
\section{Theory: Slot Identifiability Result}
\label{sec:identifiability}
In this section, we leverage the properties of the probabilistic slot attention method proposed in Section~\ref{sec:expectation_maximization} to prove a new object-centric identifiability result. We show that object-centric representations (i.e. slots) are identifiable without supervision (up to an equivalence relation) under mixture model-like assumptions about the latent space. This contrasts with existing work, which provides identifiability guarantees within a specific class of mixing functions, i.e. additive decoders~\cite{lachapelle2024additive}. Our result unifies generative model identifiablity~\cite{hyvarinen2019nonlinear,khemakhem2020variational,kivva2022identifiability} and object-centric learning.
\begin{definition}[Identifiability.]
\label{def: identifiability}
Given an observation space $\mathcal{X}$, a probabilistic model $p$ with parameters $\boldsymbol{\theta} \in \boldsymbol{\Theta}$ is said to be identifiable if the mapping $\boldsymbol{\theta} \in \boldsymbol{\Theta} \mapsto p_{\boldsymbol{\theta}}(\mathbf{x})$ is injective:
\begin{equation}
    \left(p_{\boldsymbol{\theta}_1}(\rvx) = p_{\boldsymbol{\theta}_2}(\rvx), \forall \mathbf{x} \in \mathcal{X}\right) \implies \boldsymbol{\theta}_1 = \boldsymbol{\theta}_2.
    \label{eqn:strongidentifiability}
\end{equation}
\end{definition}
\begin{remark}
Definition~\ref{def: identifiability} says that if any two choices of model parameters lead to the same marginal density, they are equal. This is often referred to as \textit{strong} identifiability~\cite{hyvarinen2019nonlinear,khemakhem2020variational}, and it can be too restrictive, as guaranteeing identifiability up to a simple transformation (e.g. affine) is acceptable in practice. To reflect weaker notions of identifiability, we let $\sim$ denote an equivalence relation on $\boldsymbol{\Theta}$, such that a model can be said to be identifiable up to $\sim$, or $\sim$-identifiable. 
\end{remark}
\begin{definition}[$\sim_{s}$-equivalence]
    Let $f_{\boldsymbol{\theta}}: \gS \rightarrow  \gX$ denote a mapping from slot representation space $\mathcal{S}$ to image space $\mathcal{X}$ (satisfying Assumption \ref{ass:weak_inj}), the equivalence relation $\sim_s$ w.r.t. to parameters $\boldsymbol{\theta} \in \boldsymbol{\Theta}$ is defined bellow, where $\mP \in \gP \subseteq \{0, 1\}^{K \times K}$ is a slot permutation matrix, $\mH \in \mathbb{R}^{d \times d}$ is an affine transformation matrix, and $\mathbf{c} \in \mathbb{R}^d$:
    \begin{align}
        &&\forall \rvx \in \gX, &&\boldsymbol{\theta}_1 \sim_{s} \boldsymbol{\theta}_2 \iff \exists \, \mP, \mH, \mathbf{c} : f_{\boldsymbol{\theta}_1}^{-1}(\rvx) = \mP (f_{\boldsymbol{\theta}_2}^{-1}(\rvx)\mH + \rvc).&&
    \end{align}
    \label{dfn:sequivalence}
\end{definition}
\begin{lemma}[Aggregate Posterior Mixture]
    Given that probabilistic slot attention induces a local (per-datapoint $\mathbf{x} \in \{ \mathbf{x}_i\}_{i=1}^M$) GMM with $K$ components, the aggregate posterior $q(\mathbf{z})$ obtained by marginalizing out $\mathbf{x}$ is a non-degenerate global Gaussian mixture with $MK$ components:
    \begin{align}
        q(\mathbf{z}) &= \frac{1}{M} \sum_{i=1}^M \sum_{k=1}^K \widehat{\bpi}_{ik} \gN\left(\rvz; \widehat{\bmu}_{ik}, \widehat{\bsigma}^2_{ik} \right).
    \end{align}
    \label{prp:aggregate_poterior1}
    \vspace{-15pt}
\end{lemma}
\begin{sproof}\renewcommand{\qedsymbol}{}
    The full proof is given in Appendix~\ref{appendix:aggregate_posteriors}. The result is obtained by integrating the product of the latent variable posterior density $q(\mathbf{z} \mid \mathbf{x})$ and the (local) GMM density given $\mathbf{z}$, w.r.t. $\mathbf{x}$. We then proceed by verifying that the mixing coefficients sum to one over all the components in the new mixture (Corollary~\ref{lemma:nondegenerate}), proving aggregated posterior to be a well-defined probability distribution, this can be empirically confirmed in Figure \ref{fig:agg_post_ex}. 
    Next, we use $q(\mathbf{z})$ in our identifiablity result.
\end{sproof}

\begin{theorem}[Mixture Distribution of Concatenated Slots] Let $f_s$ denote a permutation equivariant PSA function such that $f_s( \mathbf{z}, P \mathbf{s}^t) = P f_s(\mathbf{z}, \mathbf{s}^t)$, where $P \in \{0, 1\}^{K \times K}$ is an arbitrary permutation matrix. Let $\mathbf{s} = (\mathbf{s}_1, \dots, \mathbf{s}_K) \in \mathbb{R}^{Kd}$ be a random variable defined as the concatenation of $K$ individual slots, where each slot is Gaussian distributed within a $K$-component mixture: $\mathbf{s}_k \sim \mathcal{N}(\boldsymbol{\mu}_k, \boldsymbol{\Sigma}_k) \in \mathbb{R}^d, \forall k \in \{1, \dots K\}$. Then, $\mathbf{s}$ is also GMM distributed with $K!$ mixture components:
\begin{align}
    &&p(\mathbf{s}) = \sum_{p=1}^{K!} \boldsymbol{\pi}_p \mathcal{N}(\mathbf{s}; \boldsymbol{\mu}_p, \boldsymbol{\Sigma}_p), &&\text{where} &&\boldsymbol{\pi} \in \Delta^{K!-1}, && \boldsymbol{\mu}_p \in \mathbb{R}^{Kd},&& \boldsymbol{\Sigma}_p \in \mathbb{R}^{Kd \times Kd}. &&
\end{align}
\label{thm:concatenated_mixture}
\end{theorem}
\vspace{-20pt}
\begin{sproof}\renewcommand{\qedsymbol}{}
    The proof is in Appendix~\ref{appendix:aggregate_posteriors}. We observe that the permutation equivariance of the PSA function $f_s$ induces $K!$ ways of concatenating sampled slots $\mathbf{s}_k$, where each permutation maps to a different mode with block diagonal covariance structure in a GMM living in $\mathbb{R}^{Kd}$ (e.g. Fig.~\ref{fig:3d},~\ref{fig:concat_slot_mix}).
\end{sproof}
\begin{theorem}[$\sim_s$-Identifiable Slot Representations]
\label{thm:affine_identifiability1}
Given that the aggregate posterior $q(\mathbf{z})$ is an optimal, non-degenerate mixture prior over slot space (Lemma~\ref{prp:aggregate_poterior1}),  $f : \mathcal{S} \to \mathcal{X}$ is a piecewise affine weakly injective mixing function (Assumption~\ref{ass:weak_inj}), and the slot representation, $\mathbf{s} = (\mathbf{s}_1,\dots,\mathbf{s}_K)$ can be observed as a sample from a GMM (Theorem \ref{thm:concatenated_mixture}), then $p(\rvs)$ is identifiable as per Definition~\ref{dfn:sequivalence}.
\vspace{-8pt}
\end{theorem}
\begin{sproof}\renewcommand{\qedsymbol}{}
    The proof is given in Appendix~\ref{appendix:identifiability_proofs}. Lemma~\ref{prp:aggregate_poterior1} and Corollary~\ref{lemma:nondegenerate} state that the optimal latent variable prior in our case is GMM distributed, non-degenerate and equates to the aggregate posterior $q(\mathbf{z})$. This permits us to extend~\cite{kivva2022identifiability}'s result to show that if $q(\mathbf{z})$ is distributed according to a non-degenerate GMM and the mixing function $f_d$ is piecewise affine and weakly injective, then the \textit{slot} distribution representation, $p(\rvs)$ which is also a GMM (Theorem \ref{thm:concatenated_mixture}) is identifiable up to an affine transformation and arbitrary slot permutation.
\end{sproof}
\begin{corollary}
[Individual Slot Identifiability]
\label{cor:slotidentifiability}
If the distribution over concatenated slots $p(\rvs)$, where $\mathbf{s} = (\mathbf{s}_1, \dots, \mathbf{s}_K) \in \mathbb{R}^{Kd}$, is $\sim_s$-identifiable (Theorem~\ref{thm:affine_identifiability1}) then this implies $q(\mathbf{z})$ is identifiable up to an affine transformation and permutation of the slots $\rvs_k$. Therefore, each slot distribution $\mathbf{s}_k \sim \mathcal{N}(\boldsymbol{\mu}_k, \boldsymbol{\Sigma}_k) \in \mathbb{R}^d, \forall k \in \{1, \dots K\}$ is also identifiable up to an affine transformation.
\vspace{-8pt}
\end{corollary}

%% file: sections/6_experiments.tex
\begin{figure*}[t]
    \centering
    \hfill
    \begin{subfigure}{.01\textwidth}
        \rotatebox{90}{\hspace{37pt} \footnotesize $\mathbf{z}_2$}
    \end{subfigure} 
    \hfill
    \begin{subfigure}{.17\textwidth}
        \centering
        {\footnotesize $\mathrm{Run \ \#1}$} \\[1pt]
        \includegraphics[trim={85, 55, 0, 33.2},clip,width=.985\textwidth]{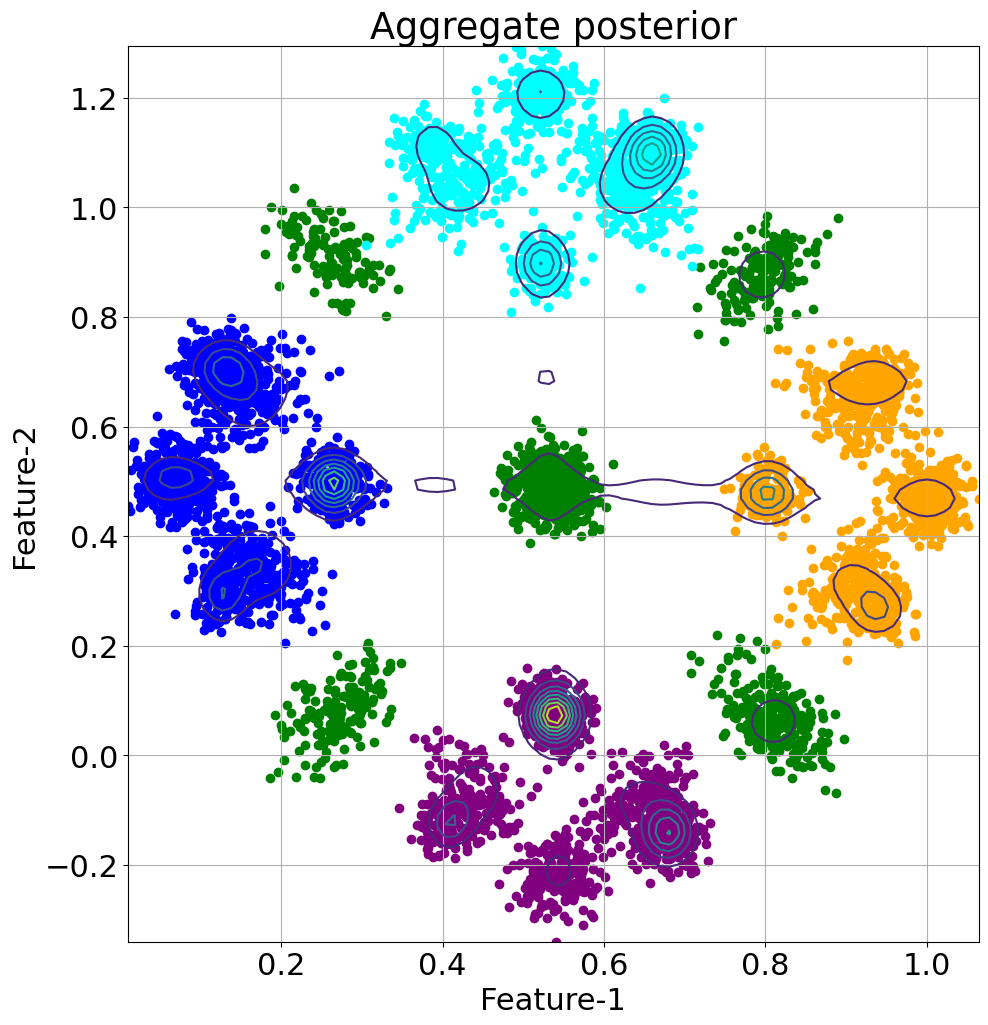}
        \vspace*{-18pt}
        \caption*{\footnotesize$\mathbf{z}_1$}
    \end{subfigure}
    \hfill
    \begin{subfigure}{.01\textwidth}
        \rotatebox{90}{\hspace{37pt} \footnotesize $\mathbf{z}_2$}
    \end{subfigure} 
    \begin{subfigure}{.17\textwidth}
        \centering
        {\footnotesize $\mathrm{Run \ \#2}$} \\[1pt]
        \includegraphics[trim={85, 55, 0, 33.2},clip,width=.985\textwidth]{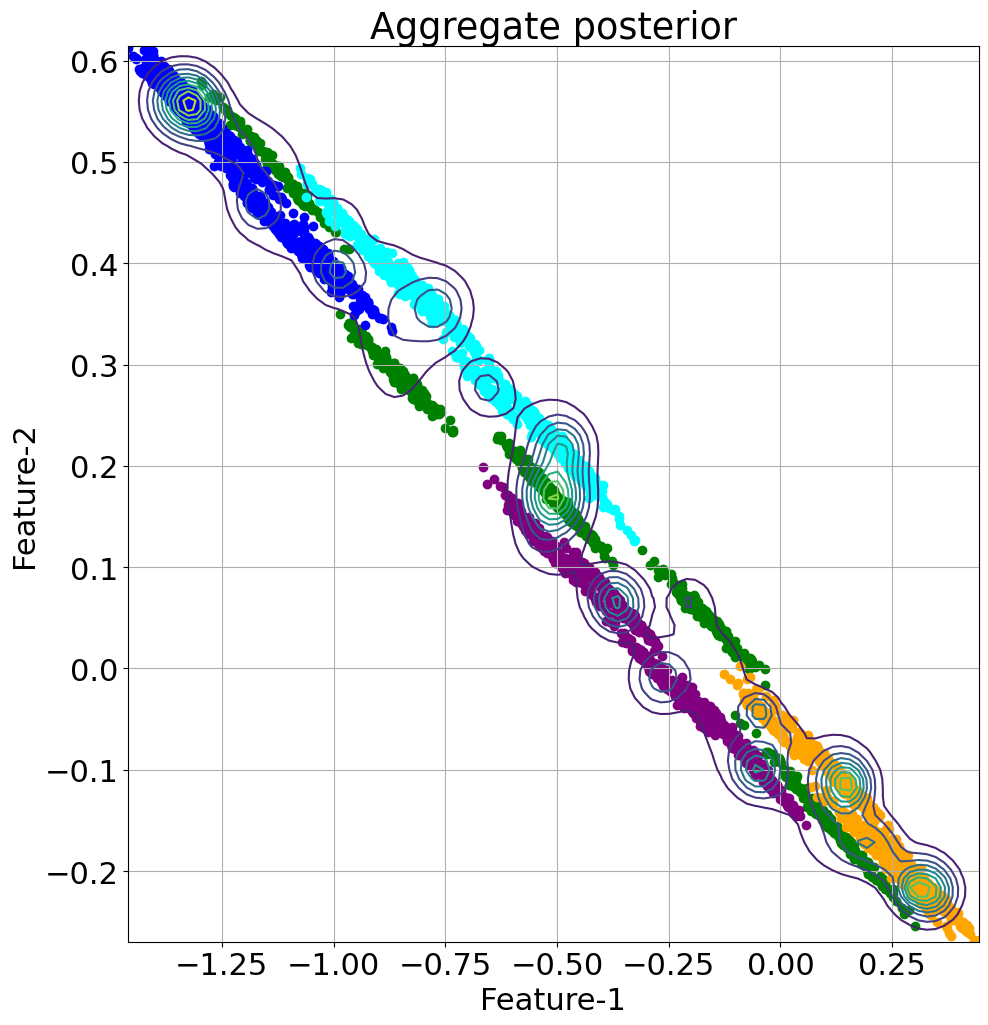}
        \vspace*{-18pt}
        \caption*{\footnotesize$\mathbf{z}_1$}
    \end{subfigure}
    \hfill
    \begin{subfigure}{.01\textwidth}
        \rotatebox{90}{\hspace{37pt} \footnotesize $\mathbf{z}_2$}
    \end{subfigure} 
    \begin{subfigure}{.17\textwidth}
        \centering
        {\footnotesize $\mathrm{Run \ \#3}$} \\[1pt]
        \includegraphics[trim={85, 55, 0, 33.2},clip,width=.985\textwidth]{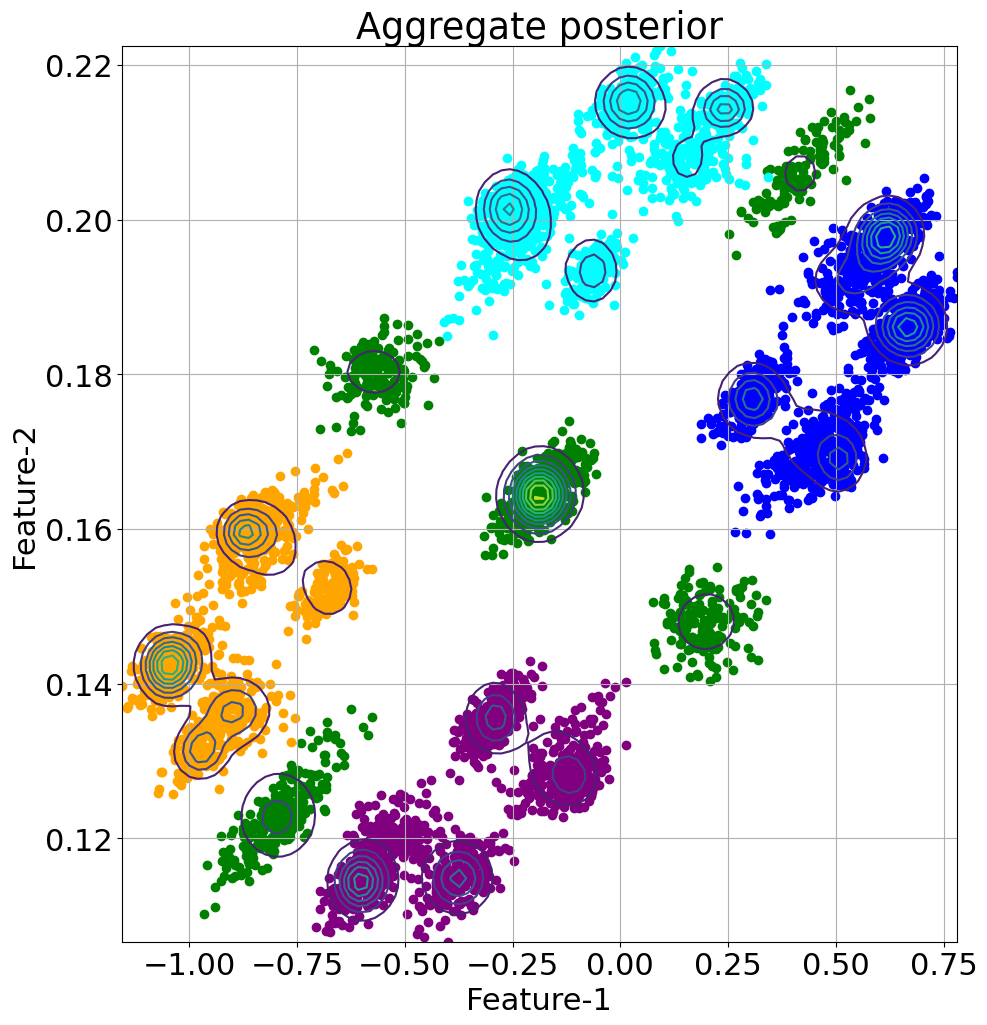}
        \vspace*{-18pt}
        \caption*{\footnotesize$\mathbf{z}_1$}
    \end{subfigure}
    \hfill
    \begin{subfigure}{.01\textwidth}
        \rotatebox{90}{\hspace{37pt} \footnotesize $\mathbf{z}_2$}
    \end{subfigure} 
    \begin{subfigure}{.17\textwidth}
        \centering
        {\footnotesize $\mathrm{Run \ \#4}$} \\[1pt]
        \includegraphics[trim={85, 55, 0, 33.2},clip,width=.985\textwidth]{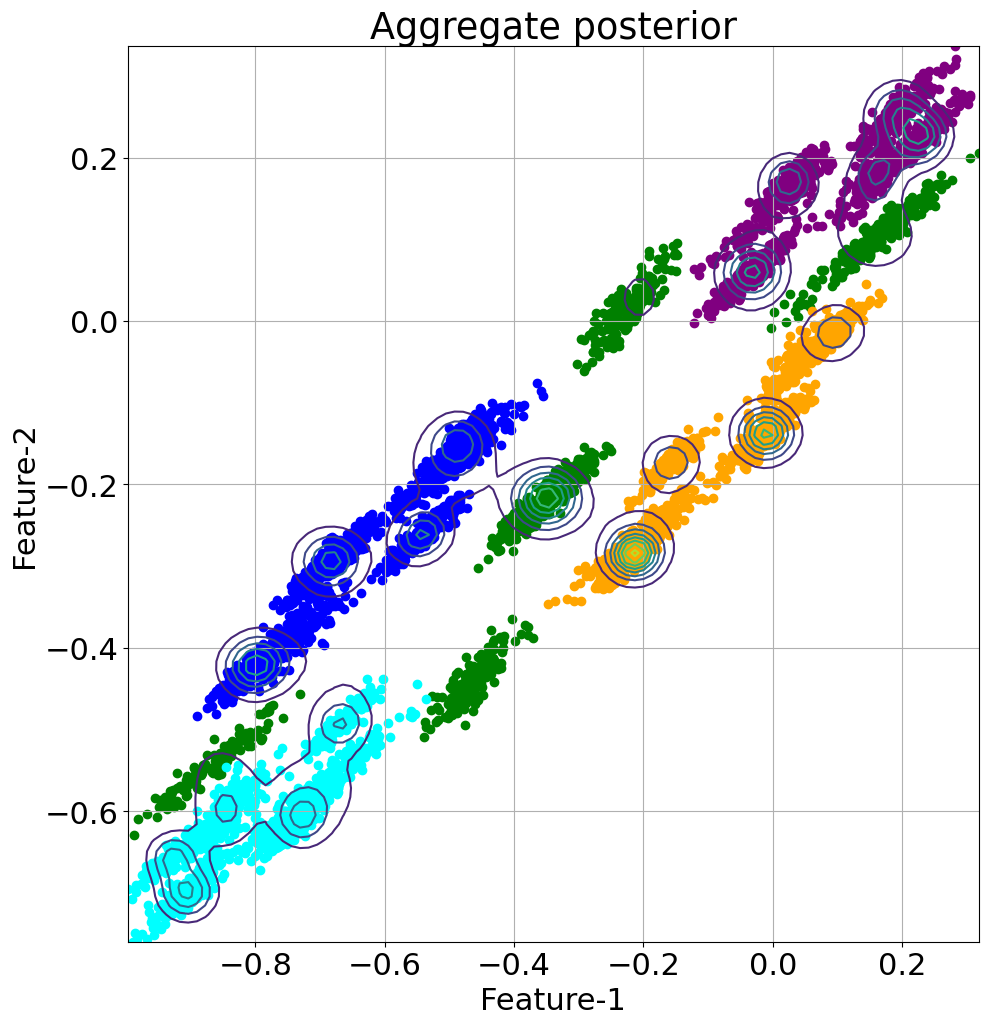}
        \vspace*{-18pt}
        \caption*{\footnotesize$\mathbf{z}_1$}
    \end{subfigure}
    \hfill
    \begin{subfigure}{.01\textwidth}
        \rotatebox{90}{\hspace{37pt} \footnotesize $\mathbf{z}_2$}
    \end{subfigure} 
    \begin{subfigure}{.17\textwidth}
        \centering
        {\footnotesize $\mathrm{Run \ \#5}$} \\[1pt]
        \includegraphics[trim={85, 55, 0, 33.2},clip,width=.985\textwidth]{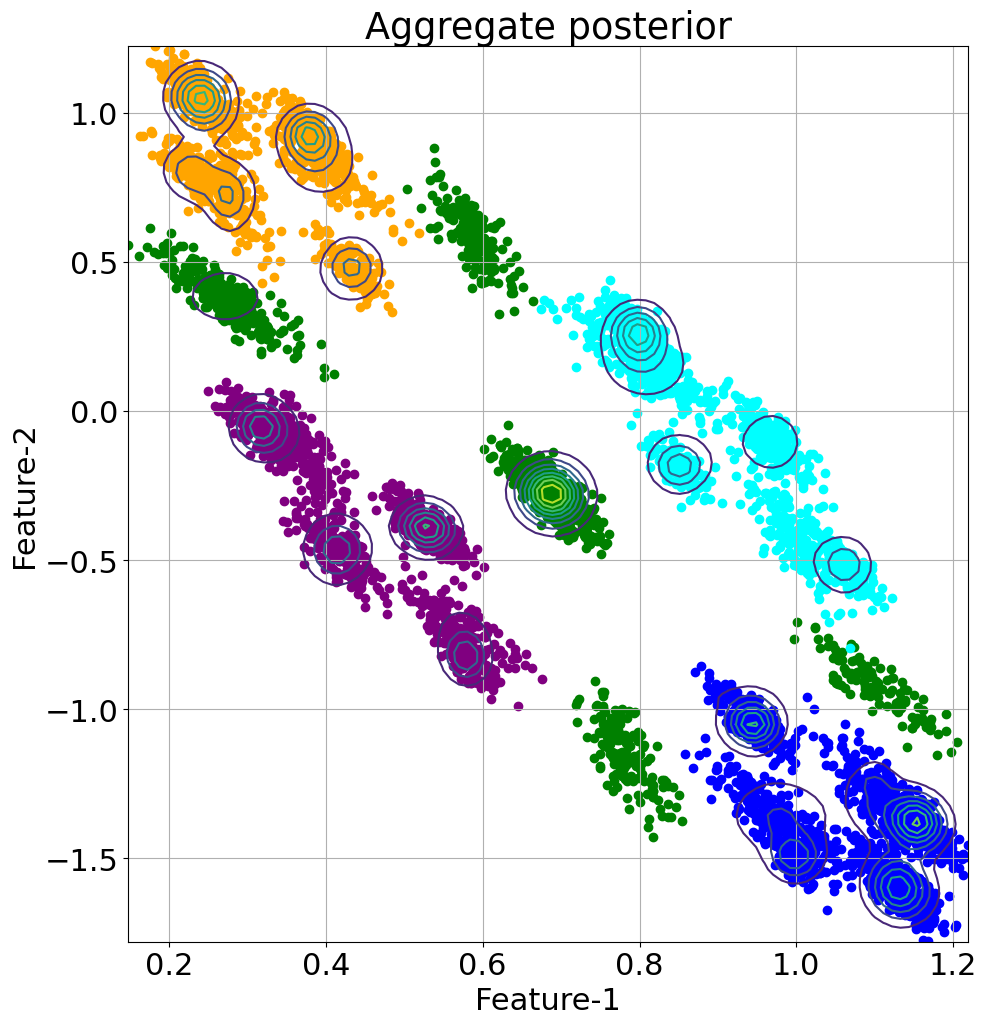}
        \vspace*{-18pt}
        \caption*{\footnotesize$\mathbf{z}_1$}
    \end{subfigure}
    \caption{\textbf{Aggregate posterior identifiability}. Recovered (latent) aggregate posteriors $q(\mathbf{z})$ across 5 runs of our PSA model. As detailed in Section~\ref{subsec:qualitative_analysis}, we used a 2D synthetic dataset with 5 total `object' clusters, with each observation containing at most 3. This provides strong evidence of recovery of the latent space up to affine transformations, empirically verifying our identifiability claim. 
    } 
    \label{fig:agg_posterior}
    \hfill
\end{figure*}
\begin{figure*}[!t]
    \hfill
    \includegraphics[width=.492\textwidth]{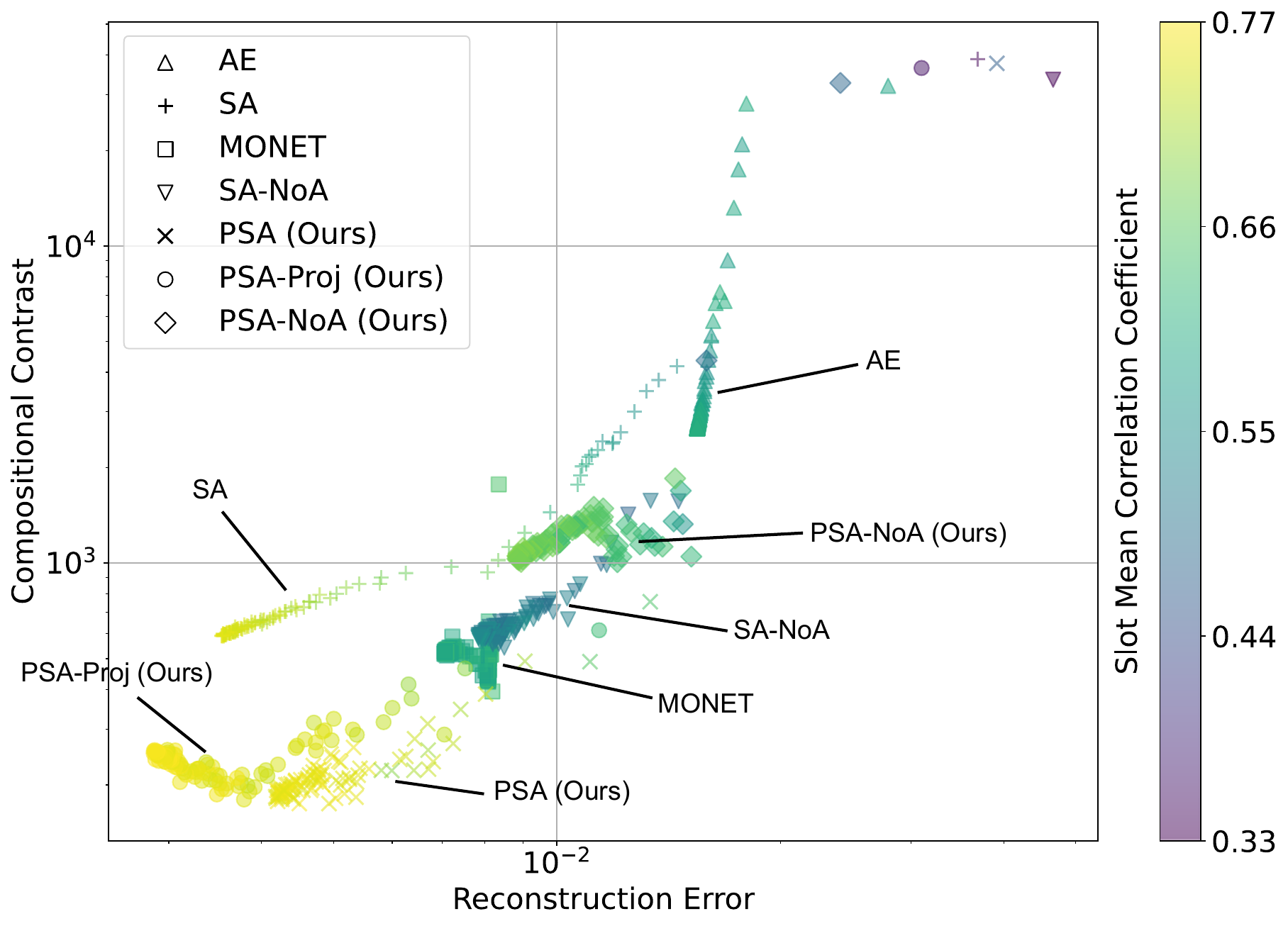} \hfill
    \includegraphics[width=.492\textwidth]{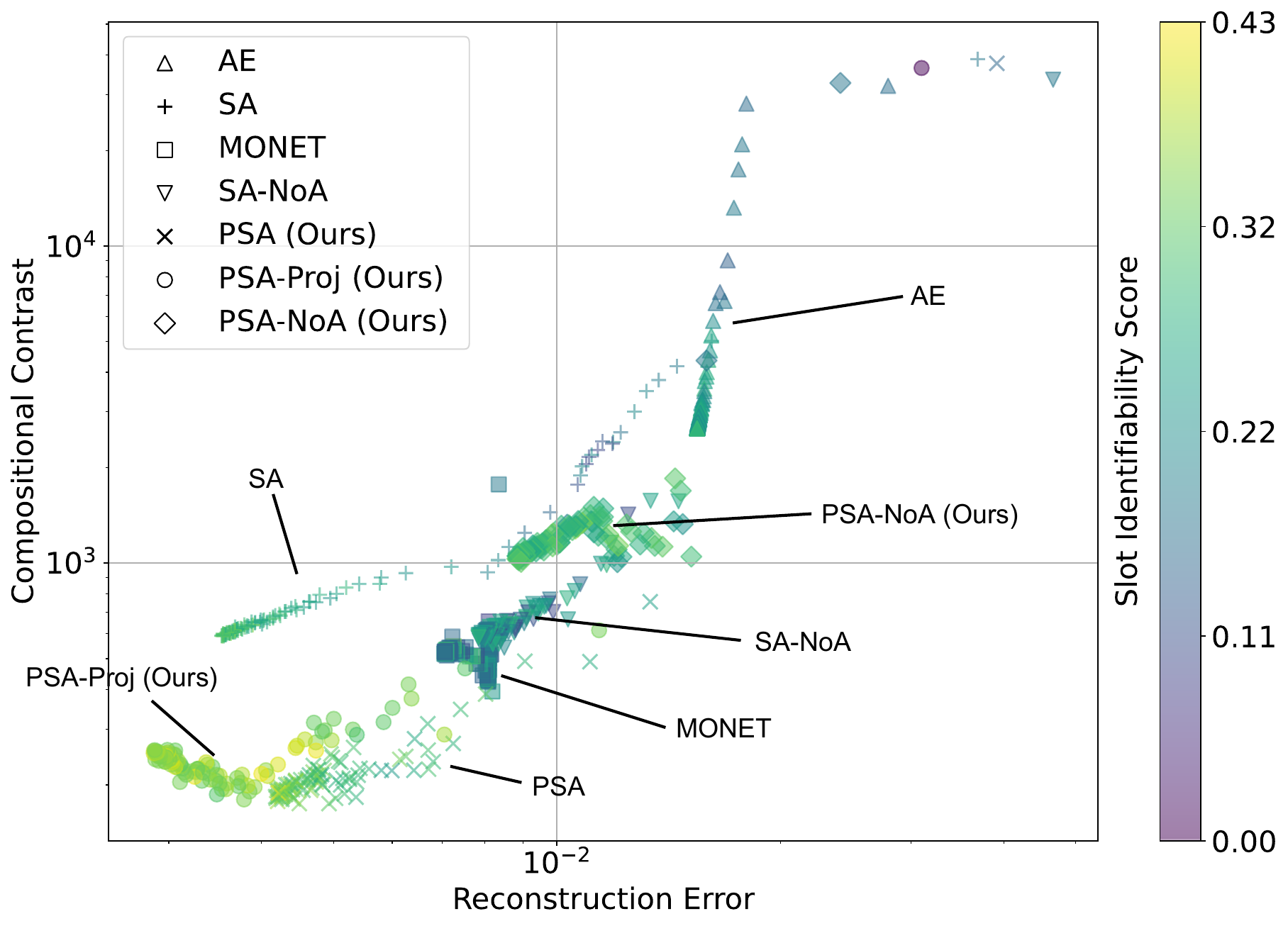} \hfill
    \caption{\textbf{Experiments comparing slot-identifiability scores}. Colour coding represents the level of slot identifiability achieved by each model as measured by the SMCC (left) and SIS (right). SMCC offers a more consistent metric correlating with reconstruction loss.}
    \label{fig:results_sis_smcc}
    \hfill
\end{figure*}
\section{Experiments}
\label{sec:experiments}
\begin{figure}[t]
    \centering
    \begin{minipage}{0.3\textwidth}
        \centering
        \includegraphics[trim={0 0 0 0},clip,width=.7\textwidth]{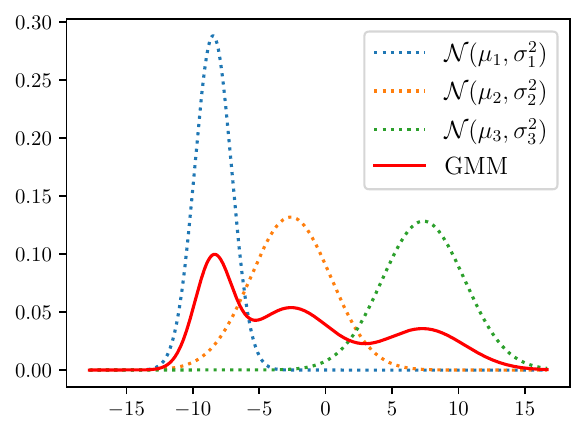}
        \\
        \centering
        \includegraphics[trim={0 0 0 25},clip,width=.95\textwidth]{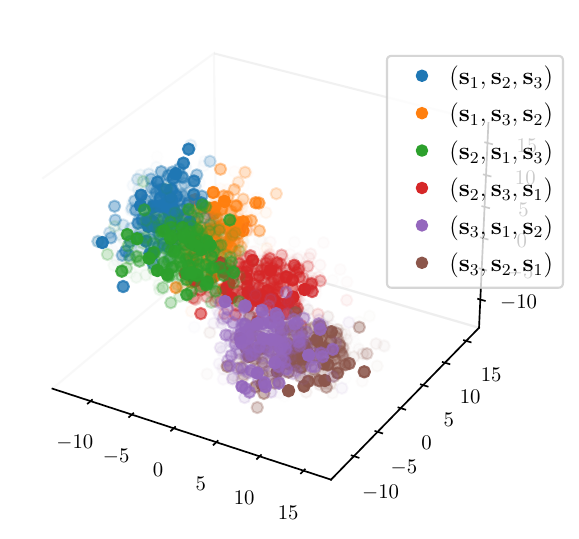}
    \end{minipage}
    \hfill
    \begin{minipage}{0.68\textwidth}
        \centering
        \captionof{table}{Comparing slot identifiability scores (SMCC and slot averaged R2) with existing object-centric learning methods.}
        \footnotesize
        \begin{tabular}{@{}lcccc@{}}
        \toprule
        \textsc{Method} & \multicolumn{2}{c}{CLEVR} & \multicolumn{2}{c}{\textsc{Objects-Room}}
        \\ 
        \cmidrule(l){2-5} 
         & SMCC $\uparrow$   & R2 $\uparrow$   & SMCC $\uparrow$   & R2 $\uparrow$ \\ 
        \midrule
        AE             & $0.43 \pm .02$ &  $0.26 \pm .02$ 
                        & $0.46 \pm .05$ & $0.45 \pm .06$\\ 
        MONET          & $0.32 \pm .01$ &  $0.39 \pm .09$ 
                         & $0.43 \pm .04$ & $0.41 \pm .10$\\
        SA             & $0.56 \pm .02$ & $0.55 \pm .05$  
                        & $0.66 \pm .01$ & $0.54 \pm .00$ \\
        SA-NoA         & $0.23 \pm .03$ & $0.24 \pm .02$  
                        & $0.48 \pm .02$ & $0.47 \pm .01$\\
        \midrule
        \textbf{Ours}: \\[2pt]
        \ \textsc{PSA-NoA} & $0.56 \pm .05$ & $0.42 \pm .06$ 
                          & $0.44 \pm .03$ & $0.45 \pm .05$ \\
        \ \textsc{PSA}     & $0.58 \pm .06$ & $0.48 \pm .02$ 
                          & $0.66 \pm .01$ & $\mathbf{0.64 \pm .02}$ \\
        \ \textsc{PSA-Proj} & $\mathbf{0.66 \pm .06}$ & $\mathbf{0.62 \pm .08}$ 
                                & $\mathbf{0.71 \pm .00}$ & $0.62 \pm .02$ \\
        \bottomrule
        \end{tabular}
        \label{table:objectdiscovery_results}
    \end{minipage}
    \caption{\textbf{Concatenated Slot Gaussian Mixture}. Example of the higher-dim GMM ($\text{1D}\to\text{3D}$) induced by the permutation equivariance of PSA and the $K!$ ways of concatenating sampled slots. 
    }
    \label{fig:3d}
\end{figure}

Given that the focus of this work is theoretical, the primary goal of our experiments is to provide strong empirical evidence of our main identifiability result (ref. Figures~\ref{fig:agg_posterior},~\ref{fig:results_sis_smcc}). With that said, we also extend our experimental study to popular imaging benchmarks to demonstrate that our method scales to higher-dimensional settings (ref. Tables~\ref{table:objectdiscovery_results},~\ref{table:compositionality_results}). 

\textbf{Datasets \& Evaluation Metrics.\;} Our experimental analysis involves standard benchmark datasets from object-centric learning literature including~\textsc{SpriteWorld}~\cite{brady2023provably},~\textsc{CLEVR}~\cite{johnson2017clevr}, and~\textsc{ObjectsRoom}~\cite{multiobjectdatasets19}. We report the foreground-adjusted rand index (FG-ARI) and FID~\cite{heusel2017gans} to quantify both object-level \textit{binding} capabilities and image quality. Our main goal is to measure slot-identifiability, so we use the slot identifiability score~\cite{brady2023provably} and the mean correlation coefficient (MCC) across slot representations -- we call the latter slot-MCC (SMCC). For two sets of slots $\{\rvs_i\}_{i=1}^M$, and $\{\tilde{\rvs}_i\}_{i=1}^M$, where $\rvs_i \in \mathbb{R}^{K \times d}$, $\tilde{\rvs}_i \in \mathbb{R}^{K \times d}$, extracted from $M$ images $\{\rvx_i\}_{i=1}^M$, the SMCC between any $\rvs$ and $\tilde{\rvs}$ is obtained by matching the slot representations and their order. The order is matched by mapping slots in $\tilde{\rvs}$ w.r.t $\rvs$ assigned by $\pi$, followed by a learned affine mapping $\mA$ between aligned $\tilde{\rvs}_{\pi(i)}$ and $\rvs$: 
\begin{align}
    \mathrm{SMCC}(\rvs, \tilde{\rvs}) \coloneqq \frac{1}{K\times d}
    \sum_{i=0}^K \sum_{j=0}^d \rho(\rvs_{ij}, \mA \tilde{\rvs}_{\pi(i)j}).
\end{align}
For more details on the metrics please refer to Appendix~\ref{appendix:metrics}.

\textbf{Models \& Baselines.\;} We consider three ablations on our proposed probabilistic slot attention (PSA) method: (i) \textsc{PSA} base model (Algorithm~\ref{alg:the_alg2}); (ii) \textsc{PSA-Proj} model (Algorithm~\ref{alg:the_alg}); and (iii) \textsc{PSA-NoA} model, which is equivalent to \textsc{PSA-Proj} but without an additive decoder. We experiment with two types of decoders: (i) an additive decoder similar to~\cite{watters2019spatial}'s spatial broadcasting model; and (ii) standard convolutional decoder. In all cases, we use LeakyReLU activations to satisfy the weak injectivity conditions (Assumption~\ref{ass:weak_inj}). In terms of object-centric learning baselines, we compare with standard additive autoencoder setups following~\cite{brady2023provably}, slot-attention (SA)~\cite{locatello2020object}, and MONET~\cite{burgess2019monet}.
\paragraph{Verifying Slot Identifiability: Gaussian Mixture of Objects.}
\label{subsec:qualitative_analysis}
To provide conclusive empirical evidence of our identifiability claim (Theorem~\ref{thm:affine_identifiability1}), we set up a synthetic modelling scenario under which it is possible to visualize the aggregate posterior across different runs. The goal is to show that PSA is $\sim_s$-identifiable (Theorem~\ref{thm:affine_identifiability1}) in the sense that it can recover the same latent space distribution up to an affine transformation and slot order permutation. For the data generating process, we defined a $K{=}5$ component GMM, with differing mean parameters $\{\bmu_1, \dots, \bmu_5\}$, and shared isotropic covariances. The 5 components emulate 5 different object types in a given environment. To create a single data point, we randomly chose 3 of the 5 components and sampled 128 points uniformly at random from each mode. Figure~\ref{fig:datasamples} shows some data samples, where different colours correspond to different objects. We used 1000 data points in total for training our PSA model. As shown in Figure \ref{fig:agg_posterior}, the aggregate posterior is either rotated, translated, skewed, or flipped across different runs as predicted by our theory -- this contrasts with all baselines wherein the aggregate posterior is intractable. We observed an SMCC of $0.93 \pm 0.04$, and R2-score of $0.50 \pm 0.08$.
\paragraph{Case Study: Imaging Data.} Although our focus is primarily theoretical, we now demonstrate that our method generalizes/scales to higher-dimensional imaging modalities. To that end, we first use the \textsc{SpriteWorld}~\cite{spriteworld19} dataset to evaluate the SMCC and SIS w.r.t. ground truth latent variables. Figure~\ref{fig:results_sis_smcc} presents our identifiability results against the baselines.
Similar to \cite{brady2023provably}, we observe higher SIS when compositional contrast and reconstruction error decreases. However, when the mixing function is \textit{not} additive, the compositional contrast does not decrease drastically while maintaining higher SMCC and SIS -- this verifies our identifiability claim using only piecewise affine decoders.
As shown in Figure \ref{fig:results_sis_smcc}, we also observe that PSA routing with additive decoder models achieves lower compositional contrast and reconstruction errors when compared with other methods. This indicates that stronger identifiability of slot representations is achievable when combining both slot latent structure and inductive biases in the mixing function.
Unlike for the \textsc{SpriteWorld} dataset, all the ground truth generative factors are \textit{unobserved} for the \textsc{CLEVR} and \textsc{ObjectsRoom} datasets we use. Therefore, for evaluation in these cases, we train multiple models with different seeds and compute the SMCC and SIS measures across different runs. This is similar to our earlier synthetic experiment and standard practice in identifiability literature. Table \ref{table:objectdiscovery_results} presents our main identifiability results on \textsc{CLEVR} and \textsc{ObjectsRoom}. We observe similar trends in favour of our proposed PSA method as measured by both SMCC and (slot averaged) R2 score relative to the baselines.
\paragraph{Case Study: Complex Decoder Structure.} To empirically test slot identifiability using more complex non-additive decoders, we used transformer decoders following SLATE \cite{singh2021illiterate}, and simply replaced the slot attention module with probabilistic slot attention.
On the CLEVR dataset, we observed a significantly improved SMCC of $\textbf{0.73} \pm .01$ and R2 of 
$\textbf{0.55} \pm .06$ relative to Table \ref{table:objectdiscovery_results}.

\begin{wraptable}[21]{r}{21.5em}
    \centering
    \footnotesize
    \caption{Pascal VOC2012 benchmark results using probabilistic slot attention (PSA). All baselines are standard results from~\cite{lowe2024rotating}. \textsc{SA MLP (w/ DINO)} denotes our replication of the DINOSAUR MLP baseline from \cite{seitzer2022bridging}, whereas $({\ddagger})$ denotes the use of slot attention masks rather than decoder alpha masks for evaluation.}
    \begin{tabular}{@{}lcc@{}}
    \toprule
    \textsc{Model} & $\textsc{mBO}_i$$\uparrow$ & $\textsc{mBO}_c$$\uparrow$ \\ 
    \midrule
    \textsc{Block Masks} & $0.247 {\:\pm\: .000}$ & $0.259 {\:\pm\:  .000}$ \\
    SA & $0.222 {\:\pm\: .008}$ & $0.237 {\:\pm\: .008}$ \\ 
    SLATE & $0.310 {\:\pm\: .004}$ & $0.324 {\:\pm\: .004}$ \\ 
    \textsc{Rotating Features} & $0.282 {\:\pm\: .006}$ & $0.320 {\:\pm\: .006}$ \\ 
    \textsc{DINO k-means} & $0.363 {\:\pm\: .000}$ & $0.405 {\:\pm\: .000}$ \\
    DINO CAE & $0.329 {\:\pm\: .009}$ & $0.374 {\:\pm\: .010}$ \\ 
    DINOSAUR MLP & $0.395 {\:\pm\: .000} $ & $0.409 {\:\pm\: .000}$ \\
    \midrule
    \textbf{Ours:}  \\[2pt]
    \textsc{SA MLP (w/ DINO)}& $0.384 {\:\pm\: .000}$ & $0.397 {\:\pm\: .000}$ \\ 
    \textsc{SA MLP (w/ DINO)}$^{\ddagger}$ & $0.400 {\:\pm\: .000}$ & $0.415 {\:\pm\: .000}$ \\
    \textsc{PSA MLP (w/ DINO)} & $0.389 {\:\pm\: .009}$ & $0.422 {\:\pm\: .009}$ \\
    \textsc{PSA MLP (w/ DINO)}$^\ddagger$ & $\textbf{0.405} {\:\pm\: .010}$ & $\textbf{0.436} {\:\pm\: .011}$ \\ 
    \bottomrule
    \end{tabular}
    \label{table:transformer_results}
\end{wraptable}
To demonstrate that probabilistic slot attention can scale to large-scale real-world data we ran additional experiments on the Pascal VOC2012~\cite{li2018tell} dataset, following the exact \textsc{DINOSAUR} strategies and setups described in~\cite{seitzer2022bridging,kakogeorgiou2024spot} for fairness, then simply swapping out the slot attention module with probabilistic slot attention. As shown in Table~\ref{table:transformer_results}, we find that probabilistic slot attention is competitive with standard slot attention on real-world data. We also tested more complex, \textit{non-additive} decoders based on autoregressive transformers, following the \textsc{DINOSAUR}~\cite{seitzer2022bridging} setup. For our \textsc{PSA Transformer (w/ DINO)} model, we observed an $\textsc{mBO}_i$ of \textbf{0.447}, and $\textsc{mBO}_c$ of \textbf{0.521} which is competitive with the \textsc{DINOSAUR Transformer} baseline~\cite{seitzer2022bridging} of 0.44 and 0.512 respectively. In this case, we found that a lower maximum learning rate of $10^{-4}$ was beneficial for stabilizing PSA training. In summary, our experiments corroborate our theoretical results and suggest why \textit{non-additive} decoder structures can still work well given the appropriate latent structure and inference procedures are in place.
With that said, there is a trade-off between identifiability and expressivity induced by the choice of decoder structure \cite{lachapelle2024additive}, so depending on the use case, it may indeed be advantageous to combine both latent and additive decoder structures in practice.

%% file: sections/7_conclusion.tex
\section{Discussion}
\label{sec:discussion}
Understanding when object-centric representations can theoretically be identified is important for scaling slot-based methods to high-dimensional images with correctness guarantees.
In contrast with existing work, which focuses primarily on properties of the slot \textit{mixing function}, we leverage distributional assumptions about the slot latent space to prove a new slot-identifiability result. Specifically, we prove that object-centric representations are identifiable without supervision (up to an equivalence relation) under mixture model-like distributional assumptions on the latent slots. To that end, we proposed a probabilistic slot-attention algorithm that imposes an \textit{aggregate} mixture prior over slot representations which is both demonstrably stable across runs and tractable to sample from. Our empirical study primarily verifies our theoretical identifiability claim and demonstrates that our framework achieves the lowest compositional contrast without being explicitly trained towards that objective, which is computationally infeasible beyond toy datasets. 
In summary, we show how slot identifiability can be achieved via probabilistic constraints on the latent space and piecewise decoders. These piecewise decoders manifest as e.g. standard MLPs with LeakyReLU activations and are generally less restrictive than additive decoders. When coupling probabilistic and additive decoder structures, we observe further performance improvements relative to either one in isolation.

\textbf{Limitations \& Future Work.} \, We recognize that our assumptions, particularly the \textit{weak injectivity} of the mixing function, may not always hold in practice for different types of architectures (see Appendix~\ref{app: Injective Decoders: Sufficient Conditions} for sufficiency conditions).
Although generally applicable, the piecewise-affine functions we use may not always accurately reflect valid assumptions about real-world problems, \emph{e.g.} when the model is misspecified. 
Like all object-centric learning methods, we also assume that the mixing function is invariant to permutations of the slots in practice, which technically makes it non-invertible. 
We deem this aspect to be an interesting area for future work, as an extension to accommodate permutation invariance would strengthen and generalize the identifiability guarantees we provide. 
Additionally, we do not study cases where objects are occluded, \emph{i.e.} when are shared or bordering other objects. 
This limitation is not unique to our work \cite{locatello2020object,brady2023provably, emami2022slot, engelcke2019genesis, kori2023grounded} and overcoming it requires further investigation by the community.
Nonetheless, our theoretical results capture the important concepts in object-centric learning and represent a valuable extension to the nascent theoretical foundations of the area.
In future work, it would be valuable to further relax slot identifiability requirements/assumptions and study slot compositional properties of probabilistic slot attention.

%% file: sections/appendix.tex
\onecolumn
\begin{figure}[!t]
    \centering
    \includegraphics[align=c, trim={0 0 0 0},clip,width=.23\textwidth]{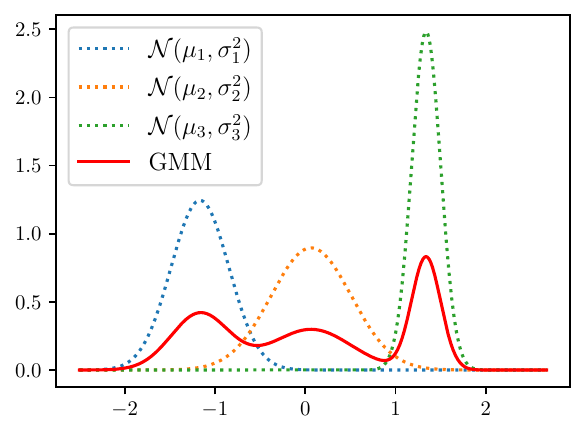}
    \includegraphics[align=c, trim={0 0 0 25},clip,width=.25\textwidth]{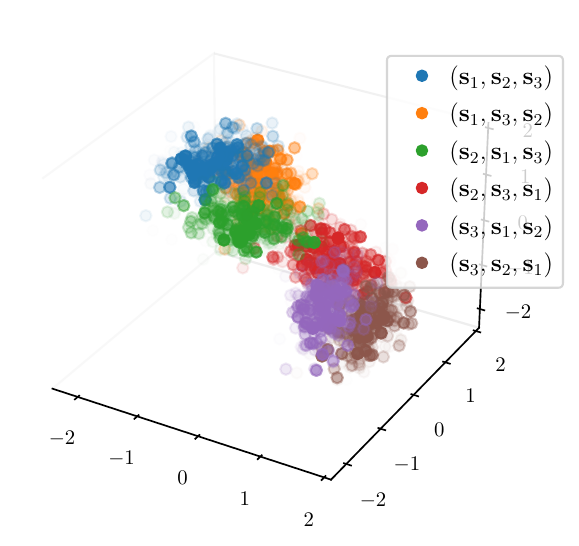}
    \hfill
    \vline
    \hfill
    \includegraphics[align=c, trim={0 0 0 0},clip,width=.23\textwidth]{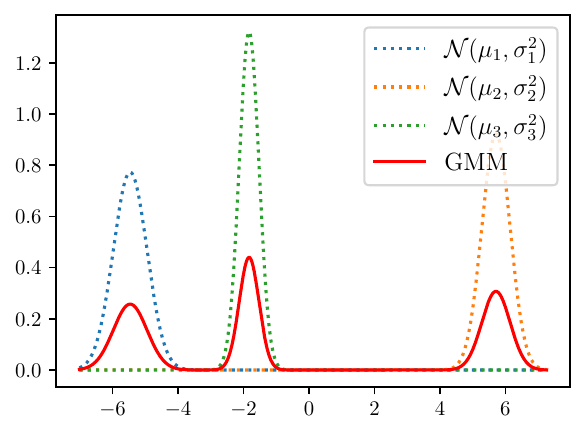}
    \includegraphics[align=c, trim={0 0 0 25},clip,width=.25\textwidth]{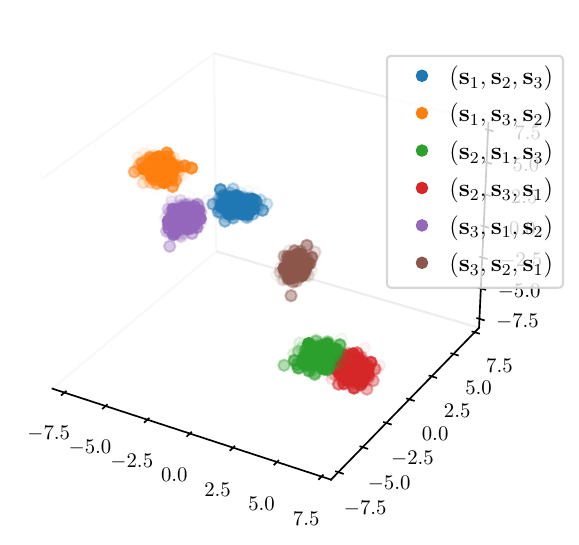}    
    \caption{\textbf{Concatenated Slot Gaussian Mixture.} Examples of the higher dimensional Gaussian mixture induced by the permutation equivariance of slot attention and the $K!$ ways of concatenating sampled slots. Here we start with a random 1D GMM with $K=3$ modes, each representing a different slot distribution, which then induces a respective 3D GMM with $K!=6$ modes.}
    \label{fig:concat_slot_mix}
\end{figure}
\section{List of Assumptions}
\label{appendix:assumptions}
\begin{assumption}[$\gB$- Disentanglement, \cite{lachapelle2024additive}]
Let $\rvs = \{\rvs_B, \forall \, B \in \gB \}$ be a set of features wrt partition set $\gB$. 
The learned mixing function $f_d$ is said to be $\gB$ disentangled wrt true decoder $\tilde{f}_d$ if there exists a permutation respecting diffeomorphism $v_B = \tilde{f}_d^{-1} \circ f_d \, \forall \, B \in \gB$ which for a given feature $\rvs$ can be expressed as $v_B(\rvs) = v_B(\rvs_B)$.  
\label{ass:beta_disentanglement}
\end{assumption}
\begin{assumption}[Additive Mixing Function]
A mixing function $f_d: \gS \rightarrow \gX$, is said to
be additive if there exist a partition set $\gB$ and functions $f_d^B: s \rightarrow \mathbb{R}^{|\gX|}$ such that:
    $f_d(\rvs) = \sum_{B \in \gB} f_d^B(\rvs_{B})$.
\label{ass:additive_decoder}
\end{assumption}
%
%

\begin{assumption}[Irreducibility]
Given an an object $\rvx_k \in \rvx$, a model is considered as irreducible if any subset of an object, $\rvy \subseteq \rvx_k$ is not functionally independent of the complement of the subset contained within the object, $\rvy^c \cap \rvx$ as expressed by the Jacobin rank inequality in equation 5 in \cite{brady2023provably}.
\label{ass:irreducibility}
\end{assumption}
\begin{assumption}[Compositionality]
\label{definition:comp}
Compositionality as defined by is a structure imposed on the slot decoder $f_d$ which implies that each image pixel is a function of at most one slot representation, thereby enforcing a local sparsity structure on the Jacobian matrix of $f_d$~\cite{brady2023provably}. 
\end{assumption}
\begin{remark}
The compositionality is guaranteed by explicitly minimising the compositionality contrast but was empirically observed to be implicitly satisfied in the case of additive decoders~\cite{brady2023provably}. 
Later, \cite{lachapelle2024additive} showed this as the property of additive decoder models. However, the additive decoders studied by \cite{lachapelle2024additive} are not expressive enough to represent the “masked decoders” typically used in object-centric representation learning, which stems from the normalization of the alpha masks. This means some care must be taken in extrapolating the results in \cite{lachapelle2024additive} to the models we use in practice. Additionally, additive decoders scale linearly in the number of slots $K$, so some less significant scalability issues remain relative to state-of-the-art non-additive decoders (e.g. using Transformers).
\end{remark}
\begin{assumption} [$\rvu$ task] Conditioning latent variables on an observed variable to yield identifiable models. The main assumption is that conditioning on a (potentially observed) variable $\mathbf{u}$ renders the latent variables independent of each other~\cite{khemakhem2020variational}.
\label{ass:aux_data}
\end{assumption}
\begin{assumption} [Object Sufficiency]
A model is said to be object sufficient iff there are no additional objects in the original data distributions other than the ones expressed in training data.
\label{ass:object_sufficiency}
\end{assumption}
\begin{assumption}[Decoder Injectivity]
    The function $f_d: \gS \rightarrow \gX$ mapping from slot space to image space is a non-linear piecewise affine injective function. That is, it specifies a unique one-to-one mapping between slots and images.
    \label{ass:decoder_injectivity}
\end{assumption}
\begin{remark}
    In practice, we use a monotonically increasing decoder with leakyReLU activation which should encourage injectivity behaviour \cite{Khemakhem2020_ice, khemakhem2020variational}.
\end{remark}
\begin{assumption}[\textit{Weak} Injectivity~\cite{kivva2022identifiability}] Let $f : \mathcal{Z} \to \mathcal{X}$ be a mapping between latent space and image space, where $\mathrm{dim}(\mathcal{Z}) \leq \mathrm{dim}(\mathcal{X})$. The mapping $f_d$ is weakly injective if there exists $\mathbf{x}_0 \in \mathcal{X}$ and $\delta > 0$ such that $|f^{-1}(\{\mathbf{x}\})| = 1$, $\forall \; \mathbf{x} \in B(\mathbf{x}_0, \delta) \cap f(\mathcal{Z})$, and $\{\mathbf{x} \in \mathcal{X} : \mid f^{-1}(\{\mathbf{x}\}) \mid = \infty\} \subseteq f(\mathcal{Z})$ has measure zero w.r.t. to the Lebesgue measure on $f(\mathcal{Z})$.
\label{ass:weak_inj}
\end{assumption}
%
\begin{remark}
    In words, Assumption~\ref{ass:weak_inj} says that a mapping $f_d$ is weakly injective if: (i) in a small neighbourhood around a specific point $\mathbf{x}_0 \in \mathcal{X}$ the mapping is injective -- meaning each point in this neighbourhood maps to exactly one point in the latent space $\mathcal{Z}$; and (ii) while $f_d$ may not be globally injective, the set of points in $\mathcal{X}$ that map back to an infinite number of points in $\mathcal{Z}$ (non-injective points) is almost non-existent in terms of the Lebesgue measure on the image of $\mathcal{Z}$ under $f_d$.
\end{remark}
%
\section{Aggregate Posterior Proofs}
\label{appendix:aggregate_posteriors}
\paragraph{Lemma \ref{prp:aggregate_poterior1}}(Aggregate Posterior Mixture) \,
    Given that probabilistic slot attention induces a local (per-datapoint $\mathbf{x} \in \{ \mathbf{x}_i\}_{i=1}^M$) GMM with $K$ components, the aggregate posterior $q(\mathbf{z})$ obtained by marginalizing out $\mathbf{x}$ is a non-degenerate global Gaussian mixture with $MK$ components given by:
    \begin{align}
        q(\mathbf{z}) &= \frac{1}{M} \sum_{i=1}^M \sum_{k=1}^K \widehat{\bpi}_{ik} \gN\left(\rvz; \widehat{\bmu}_{ik}, \widehat{\bsigma}^2_{ik} \right).
    \end{align}
\begin{proof}
We begin by noting that the aggregate posterior $q(\rvz)$ is the optimal prior $p(\mathbf{z})$ so long as our posterior approximation $q(\rvz \mid \rvx)$ is close enough to the true posterior $p(\rvz \mid \rvx)$, since for a dataset $\{\mathbf{x}_i\}_{i=1}^M$ we have that:
\begin{align}
    p(\rvz) & = \int p(\rvz \mid \rvx) p(\rvx) d\rvx 
    \\[0pt] & 
    = \mathbb{E}_{\rvx \sim p(\rvx)} \left[p(\rvz\mid \rvx)\right] 
    \\[0pt] & 
    \approx \frac{1}{M}\sum_{i=1}^M p(\rvz\mid \rvx_i)
    \\[0pt] & 
    \approx \frac{1}{M}\sum_{i=1}^M q(\rvz\mid \rvx_i)
    \\[0pt] & 
    \eqqcolon q(\rvz),
\end{align}
where we approximated $p(\mathbf{x})$ using the empirical distribution, then substituted in the approximate posterior and marginalized out $\mathbf{x}$. This observation was first made by~\cite{hoffman2016elbo} and we use it to motivate our setup.

In our case, probabilistic slot attention (Algorithm~\ref{alg:the_alg}) fits a (local) GMM to each latent variable sampled from the approximate posterior: $\rvz \sim q(\rvz \mid \rvx_i)$, for $i=1,\dots,M$. Let $f(\rvz)$ denote the (local) GMM density, its expectation is given by:
\begin{align}
    \mathbb{E}_{p(\rvx), q(\rvz\mid \rvx)} \left[ f(\rvz)\right] 
    & = \iint p(\rvx) q(\rvz \mid \rvx)f(\rvz) d\rvx d\rvz
    \\[5pt] & \approx \iint \frac{1}{M} \sum_{i=1}^M \delta(\rvx - \rvx_i) q(\rvz \mid \rvx)f(\rvz) d\rvx d\rvz
    \\[5pt] & = \int \frac{1}{M} \sum_{i=1}^M q(\rvz \mid \rvx_i)f(\rvz) d\rvz
    \\[5pt] & = \int \frac{1}{M} \sum_{i=1}^M \gN\big(\rvz; \bmu(\rvx_i), \bsigma^2(\rvx_i)\big) \cdot \sum_{k=1}^K \bpi_{ik} \gN\left(\rvz; \bmu_{ik}, \bsigma^2_{ik} \right) d\rvz
    \\[5pt] & \approx \int \frac{1}{M} \sum_{i=1}^M \delta(\rvz - \bmu(\rvx_i)) \cdot \sum_{k=1}^K \bpi_{ik} \gN\left(\rvz; \bmu_{ik}, \bsigma^2_{ik} \right) d\rvz \label{eq: dap}
    \\[5pt] & = \frac{1}{M} \sum_{i=1}^M \sum_{k=1}^K \bpi_{ik} \gN\left(\bmu(\rvx_i); \bmu_{ik}, \bsigma^2_{ik} \right)
    \\[5pt] & \eqqcolon q(\rvz),
\end{align}
where we again used the empirical distribution approximation of $p(\rvx)$, and the following basic identity of the Dirac delta to simplify: $\int \delta(\mathbf{x} - \mathbf{x}') f(\mathbf{x}) d\mathbf{x} = f(\mathbf{x}')$. 

For the general case, however, we must instead compute the product of $q(\mathbf{z} \mid \mathbf{x})$ and $f(\mathbf{z})$ rather than use a Dirac delta approximation as in Equation~\ref{eq: dap}. To that end we may proceed as follows w.r.t. to each datapoint $\mathbf{x}_i$:
\begin{align}
     q(\mathbf{z} \mid \mathbf{x}_i) \cdot f(\mathbf{z}) & = \mathcal{N}\big(\mathbf{z}; \boldsymbol{\mu}(\mathbf{x}_i), \boldsymbol{\sigma}^2(\mathbf{x}_i)\big) \cdot \sum_{k=1}^K \boldsymbol{\pi}_{ik} \mathcal{N}\left(\mathbf{z}; \boldsymbol{\mu}_{ik}, \boldsymbol{\sigma}^2_{ik} \right)
     \\ & = \sum_{k=1}^K \boldsymbol{\pi}_{ik} \left[\mathcal{N}\left(\mathbf{z}; \boldsymbol{\mu}_{ik}, \boldsymbol{\sigma}^2_{ik} \right) \cdot \mathcal{N}\big(\mathbf{z}; \boldsymbol{\mu}(\mathbf{x}_i), \boldsymbol{\sigma}^2(\mathbf{x}_i)\big) \right]
    \\[5pt] & = \sum_{k=1}^K \widehat{\boldsymbol{\pi}}_{ik} \mathcal{N}\left(\mathbf{z}; \widehat{\boldsymbol{\mu}}_{ik}, \widehat{\boldsymbol{\sigma}}^2_{ik} \right),
\end{align}
where the posterior parameters of the resulting mixture are given in closed-form by:
\begin{align}
    &&\widehat{\boldsymbol{\sigma}}^2_{ik} = \left(\frac{1}{{\boldsymbol{\sigma}}^2_{ik}} + \frac{1}{{\boldsymbol{\sigma}}^2(\mathbf{x}_i)}\right)^{-1},
    && \widehat{\boldsymbol{\mu}}_{ik} = \widehat{\boldsymbol{\sigma}}^2_{ik} \left( \frac{\boldsymbol{\mu}(\mathbf{x}_i)}{\boldsymbol{\sigma}^2(\mathbf{x}_i)} + \frac{\boldsymbol{\mu}_{ik}}{\boldsymbol{\sigma}^2_{ik}}\right),&&
\end{align}
which are the standard distributional parameters obtained from a product of two Gaussians. 

For the updated mixture coefficients $\widehat{\boldsymbol{\pi}}_{ik}$, we propose a principled way to include a posterior-weighted contribution of each mode to the new mixture coefficients. First, note that $(\boldsymbol{\pi}_{i1}, \boldsymbol{\pi}_{i2}, \dots, \boldsymbol{\pi}_{iK})$ are parameters of a multinomial distribution as $\sum_{k=1}^K\boldsymbol{\pi}_{ik} = 1$, for each datapoint $\mathbf{x}_i$. Since the Dirichlet distribution is the conjugate prior of the multinomial distribution, we can place a Dirichlet prior over the mixing coefficients for each datapoint, then update it to a posterior using the data. Concretely, we place a symmetric Dirichlet prior over the mixing coefficients as follows:
\begin{align}
    &&(\boldsymbol{\pi}_{i1}, \boldsymbol{\pi}_{i2}, \dots, \boldsymbol{\pi}_{iK}) \sim \mathrm{Dirichlet}\left(\boldsymbol{\alpha}_{i1}, \boldsymbol{\alpha}_{i2}, \dots, \boldsymbol{\alpha}_{iK}\right), && \mathrm{for} \quad i = 1,2,\dots,M, &&
\end{align}
where $\boldsymbol{\alpha}_i \in \mathbb{R}^K$ are the concentration parameters of the $i^{\mathrm{th}}$ Dirichlet distribution, and $\forall i,k : \, \boldsymbol{\alpha}_{ik} = 1$, indicating uniformity over the open standard $(K-1)$-simplex. To compute the posterior Dirichlet distribution we calculate `pseudo-counts' by integrating the product of the posterior density $q(\mathbf{z} \mid \mathbf{x}_i)$ with each one of the $K$ modes of the Gaussian mixture, thereby measuring a posterior-weighted contribution of each mode $k$ to the new aggregate mixture: 
\begin{align}
    && c_{ik} = \int \mathcal{N}\left(\mathbf{z}; \boldsymbol{\mu}_{ik}, \boldsymbol{\sigma}^2_{ik} \right) \cdot \mathcal{N}\big(\mathbf{z}; \boldsymbol{\mu}(\mathbf{x}_i), \boldsymbol{\sigma}^2(\mathbf{x}_i)\big) d\mathbf{z}, && \mathrm{for} \quad i = 1,2,\dots,M, &&
\end{align}
which we can then use as pseudo-counts to compute the Dirichlet posterior:
\begin{align}
    (\widehat{\boldsymbol{\pi}}_{i1}, \widehat{\boldsymbol{\pi}}_{i2}, \dots, \widehat{\boldsymbol{\pi}}_{iK}) \mid (c_{i1}, c_{i2}, \dots, c_{iK}) \sim \mathrm{Dirichlet}\left(\boldsymbol{\alpha}_{i1}+c_{i1}, \boldsymbol{\alpha}_{i2}+c_{i2}, \dots, \boldsymbol{\alpha}_{iK}+c_{iK}\right),
\end{align}
for $i = 1,2,\dots,M$. Each posterior probability is then readily given by the mean estimate
\begin{align}
    \widehat{\boldsymbol{\pi}}_{ik} = \frac{\boldsymbol{\alpha}_{ik} + c_{ik}}{\sum_{j=1}^{K}(\boldsymbol{\alpha}_{ij} + c_{ij})} \implies \sum_{k=1}^K \widehat{\boldsymbol{\pi}}_{ik} = 1. \label{eq: post_pi}
\end{align}
Putting everything together, the aggregated posterior is therefore given by:
\begin{align}
    &&q(\mathbf{z}) = \frac{1}{M} \sum_{i=1}^M \sum_{k=1}^K \widehat{\boldsymbol{\pi}}_{ik} \mathcal{N}\left(\mathbf{z}; \widehat{\boldsymbol{\mu}}_{ik}, \widehat{\boldsymbol{\sigma}}^2_{ik} \right), && \mathrm{where} && \mathbf{z} \sim q(\mathbf{z} \mid \mathbf{x}), &&
\end{align}
which concludes the proof.
\end{proof}
\begin{corollary}
    The aggregate posterior $q(\mathbf{z})$ is a non-degenerate Gaussian mixture, in the sense that it is a well-defined probability distribution, as the updated mixture coefficients sum to 1 over the number of components $M \times K$. 
    \label{lemma:nondegenerate}
\end{corollary}
\begin{proof}
    Recall from Lemma~\ref{prp:aggregate_poterior1} that the aggregate posterior $q(\rvz)$ -- obtained by marginizaling out $\mathbf{x}$ from a probabilistic slot attention model -- is a mixture distribution of $M \times K$ components with the following parameters:
    \begin{align}
        && \left\{\widehat{\boldsymbol{\pi}}_{ik}, \widehat{\boldsymbol{\mu}}_{ik}, \widehat{\boldsymbol{\sigma}}^2_{ik} \right\}, && \mathrm{for} \quad i = 1,2,\dots,M, \quad \mathrm{and} \quad k = 1,2,\dots,K. &&
    \end{align}
    To verify that $q(\mathbf{z})$ is a non-degenerate mixture, we observe the following implication:
    \begin{align}
         &&\sum_{k=1}^K \widehat{\bpi}_{ik} = 1, && \mathrm{for} \quad i = 1,2,\dots,M, &&
    \end{align}
    due to the Dirichlet posterior update in Equation~\ref{eq: post_pi}, and therefore
    \begin{align}
        & \implies \frac{1}{M} \sum_{i=1}^M \sum_{k=1}^K \widehat{\bpi}_{ik} = \frac{1}{M} \sum_{i=1}^M 1 = \frac{1}{M} \cdot M = 1
        \\[5pt] &\implies \sum_{i=1}^M \sum_{k=1}^K \frac{\widehat{\bpi}_{ik}}{M} = 1,
    \end{align}
    which says that the scaled sum of the mixing proportions of all $K$ components in all $M$ GMMs must equal 1, proving that the associated aggregate posterior mixture $q(\mathbf{z})$ is a well-defined probability distribution.
\end{proof}

\textbf{Theorem \ref{thm:concatenated_mixture}} (Mixture Distribution of Concatenated Slots)\textbf{.} \, Let $f_s$ denote a permutation equivariant probabilistic slot attention function such that $f_s( \mathbf{z}, P \mathbf{s}^t) = P f_s(\mathbf{z}, \mathbf{s}^t)$, where $P \in \{0,1\}^{K \times K}$ is an arbitrary permutation matrix. Let $\mathbf{s} = (\mathbf{s}_1, \mathbf{s}_2, \dots, \mathbf{s}_K) \in \mathbb{R}^{Kd}$ be a random variable defined as the concatenation of $K$ individual sampled slots, where each slot is Gaussian distributed within a $K$-component mixture: $\mathbf{s}_k \sim \mathcal{N}(\rvs_k; \boldsymbol{\mu}_k, \boldsymbol{\Sigma}_k) \in \mathbb{R}^d, \forall \, k \in \{1, \dots K\}$. Then, it holds that $\mathbf{s}$ is also Gaussian mixture distributed comprising $K!$ mixture components:
\begin{align}
    &&p(\mathbf{s}) = \sum_{p=1}^{K!} \boldsymbol{\pi}_p \mathcal{N}(\mathbf{s}; \boldsymbol{\mu}_p, \boldsymbol{\Sigma}_p), &&\text{where} &&\boldsymbol{\pi} \in \Delta^{K!-1}, && \boldsymbol{\mu}_p \in \mathbb{R}^{Kd},&& \boldsymbol{\Sigma}_p \in \mathbb{R}^{Kd \times Kd}. &&
\end{align}

\begin{proof} Each slot $\mathbf{s}_k \sim \mathcal{N}(\boldsymbol{\mu}_k, \boldsymbol{\Sigma}_k)$ is sampled independently from $\mathbf{s}_j \sim \mathcal{N}(\boldsymbol{\mu}_j, \boldsymbol{\Sigma}_j)$, for any $j \neq k$, meaning they are conditionally independent given the latent mixture component assignment. Thus, the concatenated slots variable $\mathbf{s} = (\mathbf{s}_1, \mathbf{s}_2, \dots, \mathbf{s}_K)$, can be described by a $Kd$-dimensional multivariate Gaussian distribution 
with a block diagonal covariance structure as follows:
\begin{align}
    \mathbf{s} =
    \begin{bmatrix}
        \mathbf{s}_{\pi(1)} \\
        \mathbf{s}_{\pi(2)} \\
        \vdots \\
        \mathbf{s}_{\pi(K)}
    \end{bmatrix}
    \sim \mathcal{N}\left(
    \begin{bmatrix}
        \boldsymbol{\mu}_{\pi(1)} \\
        \boldsymbol{\mu}_{\pi(2)} \\
        \vdots \\
        \boldsymbol{\mu}_{\pi(K)}
    \end{bmatrix},
    \begin{bmatrix}
    \boldsymbol{\Sigma}_{\pi(1)} & 0 & \cdots & 0 \\
    0 & \boldsymbol{\Sigma}_{\pi(2)} & \cdots & 0 \\
    \vdots & \vdots & \ddots & \vdots \\
    0 & 0 & \cdots & \boldsymbol{\Sigma}_{\pi(K)}
    \end{bmatrix}\right),
\end{align}
where $\pi : [K] \to [K]$ is a permutation function of the set $[K] \coloneqq \{1, 2, \dots, K\}$. Since the slot attention function $f_s$ is permutation equivariant, there exist $K!$ possible ways to concatenate $K$ slots, and each permutation induces a mode within a Gaussian mixture living in $\mathbb{R}^{Kd}$ space. Since each permutation of the slots is equally likely, the mixture coefficients are given by:
\begin{align}
    &\boldsymbol{\pi} = (\boldsymbol{\pi}_1, \boldsymbol{\pi}_2, \dots, \boldsymbol{\pi}_{K!}), && \text{where} &&\boldsymbol{\pi}_p = \frac{1}{K!} && \forall p \in \{1,2, \dots, K!\}&&
    \\ &\implies \sum_{p=1}^{K!} \boldsymbol{\pi}_p = 1,
\end{align}
which concludes the proof.
\end{proof}
\begin{remark}
    Based on the above result, it is evident that concatenating $K\geq2$ unique slots can be viewed as a sample from a GMM with $K!$ components. Constructing a scene requires \textit{at least} two unique slots, one for the background and one for an object, thus supporting our theory regarding slot composition.
\end{remark}

\section{Injective Decoders: Sufficient Conditions}
\label{app: Injective Decoders: Sufficient Conditions}
In this section, we provide a theoretical overview of the decoder architecture we use and offer sufficient conditions for a leaky-ReLU decoder to be \textit{weakly} injective in the sense of Assumption~\ref{ass:weak_inj} as shown and adapted from \cite{kivva2022identifiability}. 
\begin{definition} [Piece-wise Decoders, \cite{kivva2022identifiability}]
\label{def:piecewise_decoders}
Let $\rvs = \{\rvs_1, \rvs_2, \dots, \rvs_K\}$ denote a given set of sampled $z_i$ in each mixture component ($K$ slots) in the GMM, $P(\mathcal{Z})$. Let $\sigma:\mathbb{R}\rightarrow \mathbb{R}$ denote the leaky-ReLU activation function, and let $m = n_0, n_1, n_2, \dots, n_t = n$ and $H(n_1, n_2)$ denote the set of full-rank affine functions $h_i:\mathbb{R}^{n_i}\rightarrow \mathbb{R}^{n_j}$. We consider piece-wise functions mapping each slot representation $\rvs \in \mathcal{S} \in \mathbb{R}^m\times K$ to an image $\rvx \in \mathcal{X} \in \mathbb{R}^n$ in the output space,  $\mathcal{F}_{\sigma}^{mK\rightarrow n}: \mathcal{S} \rightarrow \mathcal{X}$, of the form below:
\begin{equation}
\mathcal{F}^{n_0, \ldots, n_t}_{\sigma} = \Big\{h_t\circ \sigma \circ h_{t-1} \circ \sigma \circ \cdots \sigma\circ h_1 \mid h_i \in H(n_{i-1}, n_i)\Big\}.
\end{equation}
\end{definition}

The following Corollary, corollary and proofs are adapted from \cite{kivva2022identifiability}.
\begin{corollary}\label{lem:invertdecode}
Given $f_d \in \mathcal{F}_{\sigma}^{m\hookrightarrow n}$ where $f_d = h_t\circ \sigma \circ h_{t-1} \circ \sigma \circ \ldots \sigma\circ h_1$, $f_d$ is injective.
\end{corollary}

\begin{proof}
Each affine function $h_i$ has full column rank and is therefore both injective and invertible. Since the activation function $\sigma$ is also injective, we get that $f_d$ is injective and invertible.
\end{proof}

\begin{corollary} 
Let $f_d = h_t\circ \sigma \circ h_{t-1} \circ \sigma \circ \ldots \sigma\circ h_1\in \mathcal{F}^{m\rightarrow n}_{\sigma}$ where $m = n_0\leq n_1\leq \cdots \leq n_k = n$. Given $h_i$ is affine and invertible, then for almost all $x\in f_d(\mathbb{R}^m)$ there exists $\delta$ such that $f_d^{-1}$ is a well-defined affine function, $h_i$ on $B(x, \delta)\cap f_d(\mathbb{R}^m)$.
\end{corollary}
\begin{proof}
We know $f_d$ is a piecewise affine function which is invertible. Therefore, it simply follows,  $\forall y\in B(x, \delta)$ there exists $\delta$ such that $f_d^{-1}$ is an affine function in the domain $B(x, \delta)$. One can therefore deduce there exists some affine function $h_i$ where $f_d^{-1} = h_i^{-1}$ in the domain $B(x, \delta)$.
\end{proof}
\section{Slot Identifiability Proof}
\label{appendix:identifiability_proofs}

\begin{definition}[Slot Identifiability~\cite{brady2023provably}]
\label{def:slot_ident}
    Given a diffeomorphic ground truth function $f:\mathcal{S}\rightarrow \mathcal{X}$, and inference model $\hat{g}:\mathcal{X}\rightarrow \mathcal{S}$, $\hat{g}$ correctly slot identifies every object $\mathbf{x}_j \in \mathcal{X}$ with the ground slot $\mathbf{s}_k \in \mathcal{S}$ via $\hat{\mathbf{s}}_k = \hat{g}(f(\mathbf{s}_k))$ if there exists a unique slot $k \in [K]$ for all $\mathbf{x}_j \in \mathcal{X}$, and there exists an invertible diffeomorphism, or in our case an affine transformation, such that $\mathbf{s}_k = h(\hat{\mathbf{s}}_k)$ for all $\mathbf{s}_k \in \mathcal{S}$.
\end{definition}

\begin{remark}
    \cite{brady2023provably} established that \textit{compositionality} (Assumption~\ref{definition:comp}) and \textit{irreducibility} (Assumption~\ref{ass:irreducibility}) of the decoder $f_d$ is required for slot identifiability in the sense of Definition \ref{def:slot_ident}. 
\end{remark}
\paragraph{Summary \& Intuition.} The following theorems and proof extend the identifiability results of~\cite{kivva2022identifiability} to slot-attention models, we include all the proofs and details for the sake of completion.  
In this work, we do not consider the irreducibility criteria in \cite{brady2023provably} and define slot identifiability only by an injective mapping of each slot to subspaces representing objects in a scene to satisfy compositionality without the use of computationally heavy methods such as additive decoders and compositional contrast. 
We show identifiability of each slot representation up to affine transformation by passing a concatenation of samples from each mixture component which represents slots through a piecewise function in Definition \ref{def:piecewise_decoders} representing the decoder, $f_d$. The trick in this proof lies in our observation of the fact that a concatenation of samples from each slot mixture component is a sample from a high dimensional GMM $p(\mathbf{s})$, as highlighted in Theorem \ref{thm:concatenated_mixture}. 
We then use the identifiability results of~\cite{kivva2022identifiability} to show $p(\mathbf{s})$ is identifiable up to affine transformation. 
This then implies identifiability up to affine transformation of the aggregate posterior $q(\mathbf{z})$, a non-degenerate GMM by Lemma~\ref{prp:aggregate_poterior1} where a sample from a mixture component in $q(\mathbf{z})$ represents an individual slot representation. This contrasts with \cite{kivva2022identifiability} which does not consider identifiability of the aggregate posterior and its mixture components in the context of slot representation learning. 

In order to proceed, we begin by stating three key theorems defined and proven in the work of \cite{kivva2022identifiability} which are essential for our slot identifiability proof.
First, we restate the definition of a \textit{generic point} as outlined by \cite{kivva2022identifiability} below.
\begin{definition}
    A point $\rvx\in f_d(\mathbb{R}^m)\subseteq \mathbb{R}^n$ is generic if there exists $\delta>0$, such that $f_d:B(\rvs, \delta)\rightarrow \mathbb{R}^n$ is affine for every $\rvs\in f_d^{-1}(\{\rvx\})$
\end{definition}
\begin{theorem}[Kivva et al.~\cite{kivva2022identifiability}]
    \label{thm:genericfunctions}
    Given $f_d:\mathbb{R}^m\rightarrow \mathbb{R}^n$ is a piecewise affine function such that $\{\rvx\in \mathbb{R}^n\, :\, |f_d^{-1}(\{\rvx\})| = \infty\}\subseteq f_d(\mathbb{R}^m)$ has measure zero with respect to the Lebesgue measure on $f_d(\mathbb{R}^m)$, this implies $\dim (f_d(\mathbb{R}^m)) = m$ and almost every point in $f_d(\mathbb{R}^m)$ (with respect to the Lebesgue measure on $f_d(\mathbb{R}^m)$) is generic with respect to~$f_d$. 
\end{theorem}
\begin{theorem}[Kivva et al.~\cite{kivva2022identifiability}]
\label{thm:local-gmm-iden}
Consider a pair of finite GMMs  in $\mathbb{R}^m$:
\begin{align}
    &&\rvy =\sum_{j=1}^{J} \bpi_{j}\gN(\rvy; \bmu_{j}, \bSigma_{j}), &&\text{and} &&\rvy'=\sum_{j=1}^{J'} \bpi_{j}'\gN(\rvy'; \bmu'_{j}, \bSigma'_{j}).&&
\end{align}
Assume that there exists a ball $B(\rvx, \delta)$ such that $\rvy$ and $\rvy'$ induce the same measure on $B(\rvx, \delta)$. Then $\rvy \equiv \rvy'$, and for some permutation $\tau$ we have that $\bpi_i = \bpi'_{\tau(i)}$ and $(\bmu_i, \bSigma_i) = (\bmu'_{\tau(i)}, \bSigma'_{\tau(i)})$.
\end{theorem}
\begin{theorem}[Kivva et al.~\cite{kivva2022identifiability}]
    \label{thm:genericfunctions2}
        Given $\rvz~\sim \sum_{i=1}^{J} \bpi_i\gN(\rvz; \bmu_i, \bSigma_i)$ and $\rvz' \sim \sum_{j=1}^{J'} \bpi_j'\gN(\rvz'; \bmu_j', \bSigma_j')$ and $f_d(\rvz)$ and $\tilde{f}_d(\rvz')$ are equally distributed. We can assume for $\rvx \in \mathbb{R}^n$ and $\delta>0$, $f_d$ is invertible on $B(\rvx, 2\delta)\cap f_d(\mathbb{R}^{m})$.
        This implies that there exists $\rvx_1\in B(\rvx, \delta)$ and $\delta_1>0$ such that both $f_d$ and $\tilde{f}_d$ are invertible on $B(\rvx_1, \delta_1)\cap f_d(\mathbb{R}^{m})$.
\end{theorem}

We next propose our slot identifiability result below.

\textbf{Theorem \ref{thm:affine_identifiability1}} ($\sim_s$-Identifiable Slot Representations)\textbf{.}\,
Given that the aggregate posterior $q(\mathbf{z})$ is an optimal, non-degenerate mixture prior over slot space (Lemma~\ref{prp:aggregate_poterior1}),  $f_d : \mathcal{S} \to \mathcal{X}$ is a piecewise affine weakly injective mixing function (Assumption~\ref{ass:weak_inj}), and the slot representation, $\mathbf{s} = (\mathbf{s}_1,\dots,\mathbf{s}_K)$ can be observed as a sample from a GMM (Theorem \ref{thm:concatenated_mixture}), then $p(\rvs)$ is identifiable as per Definition~\ref{dfn:sequivalence}.

\begin{proof}
    The proof extends from~\cite{kivva2022identifiability} to slot-based models.
    Given two piece-wise affine functions $f_d, \tilde{f}_d: \gS \rightarrow \gX$, $\forall k \in [K]$, let $\rvs = (\rvs_1, \dots, \rvs_K), \ni \rvs_k \sim \mathcal{N}(\rvs_k; \bmu_k, \bSigma_k)$ and $\rvs' = (\rvs'_1, \dots, \rvs'_K), \ni \rvs'_k \sim  \mathcal{N}(\rvs'_k; \bmu_k', \bSigma_k')$ be a pair of slot representations constructed by sampling and concatenating each mixture component (i.e. slots) from two distinct GMMs. As proven in Theorem \ref{thm:concatenated_mixture}, given individual sampled slots $\rvs_k$ are conditionally independent given the mixture component $k$, then a concatenated sample is from a higher dimensional GMM in $\mathbb{R}^{Kd}$. 
    Now, suppose for the sake of argument that $f_d(\mathcal{S})$ and $\tilde{f}_d(\mathcal{S}')$ are equally distributed. 
    We assume that there exists $\rvx\in \gX$ and $\delta>0$ such that $f_d$ and $\tilde{f}_d$ are invertible and piecewise affine on $B(\rvx, \delta)\cap f_d(\gS)$, which implies $\dim f_d(\gS) = |\gS|$. 
    
    We now restrict the space $B(\rvx, \delta)$ to a subspace $B(\rvx', \delta')$ where $\rvx \in B(\rvx', \delta')$ such that $f_d$ and $\tilde{f}_d$ are now invertible and affine on $B(\rvx', \delta')\cap f_d(\gS)$.
    Next, we let $L\subseteq \gX$ 
    be an $|\gS|$-dimensional affine subspace (assuming $|\gX| \geq |\gS|$), such that $B(\rvx', \delta')\cap f_d(\gS)
    = B(\rvx', \delta')\cap L$. We also define $h_f, h_{\tilde{f}} : \gS \rightarrow L$ to be a pair of invertible affine functions where $h_f^{-1}(B(\rvx', \delta')\cap L) = f_d^{-1}(B(\rvx', \delta')\cap L)$ and $h_{\tilde{f}}^{-1}(B(\rvx', \delta')\cap L) = \tilde{f}_d^{-1}(B(\rvx', \delta')\cap L)$. 
    
    Therefore, this implies $h_f(\rvs)$ and $h_{\tilde{f}}(\rvs')$ are finite GMMs which coincide on $B(\rvx', \delta')\cap L$ and $h_f(\rvs)\equiv h_{\tilde{f}}(\rvs')$ based on Theorem \ref{thm:local-gmm-iden}. Given, $h = h_{\tilde{f}}^{-1}\circ h_f$ and $h_f(\rvs)$ and $h_{\tilde{f}}(\rvs')$ then $h$ is an affine transformation such that $h(\rvs) = \rvs'$.   

    Given Theorems~\ref{thm:genericfunctions} and \ref{thm:genericfunctions2}, there exists a point $\rvx \in f_d(\gS)$ that is generic with respect $f_d$ and $\tilde{f}_d$ and invertible on $B(\rvx, \delta)\cap f_d(\gS)$. Having established that there is an affine transformation $h(\rvs) = \rvs'$ and two invertible piecewise affine functions $f_d$ and $\tilde{f}_d$ on $B(\rvx, \delta)\cap f_d(\gS)$, this implies that $p(\rvs)$ is identifiable up to an affine transformation and permutation of $\rvs_k \in \rvs$, which concludes the proof. 
\end{proof}

\textbf{Corollary \ref{cor:slotidentifiability}} (Individual Slot Identifiability)\textbf{.}\, 
If the concatenated slot distribution $p(\rvs)$ is $\sim_s$-identifiable (Theorem~\ref{thm:affine_identifiability1}) then this implies $q(\rvz)$ is identifiable up to affine transformation and permutation of the slots, $\rvs_k$ and therefore each slot distribution $\rvs_k \sim \gN(\rvs_k; \bmu_k, \bSigma_k) \in \mathbb{R}^d, \forall \, k \in \{1, \dots K\}$ is also identifiable up to an affine transformation.

\begin{proof}
    Given Theorem \ref{thm:local-gmm-iden}, we know that each higher dimensional mixture component in $p(\rvs)$ induces the same measure on $B(\rvx, \delta)$ and hence for some permutation $\tau$ we have that $(\bmu_{\pi(i)}, \bSigma_{\pi(i)}) = (\bmu'_{\tau(\pi(i))}, \bSigma'_{\tau(\pi(i))})$. Therefore, each mixture component $\rvs_{\pi(i)}$ is identifiable up to affine transformation, and permutation of slots representations in $\rvs$. Now, given sampling $\rvs_k$ is equivalent to obtaining $K$ samples from the GMM, $q(\rvz)$ and concatenating, this makes $q(\rvz)$ identifiable up to affine transformation, $h$ and permutation of slot representations in $\rvs$. 

    It now trivially follows that each slot representation $\rvs_k \sim \gN(\rvs_k; \bmu_k, \bSigma_k) \in \mathbb{R}^d, \forall \, k \in \{1, \dots, K\}$ is identifiable up to affine transformation, $h$ based on the following observed property of GMMs: 
    \begin{align}
    \sum_{k=1}^{K} {\bpi}_{k}h_{\sharp} \left(\gN(\rvs_k; \bmu_{k},\bSigma_{k})\right)
    \sim
    h_{\sharp}\Big(\sum_{k=1}^{K} {\bpi}_{k}\gN(\rvs'_k;\bmu'_{k},\bSigma'_{k})\Big),
\end{align}
which concludes the proof.
\end{proof}

%
\section{Algorithm}
As a result of projection with separate query matrix is considered, the mean and variance updates are coupled with some non-linear affine transformation; for EM to have an exact solution, this transformation needs to be the identity (this can be seen when we derive the update equations for mean and variance). 
The resulting algorithms for these two cases are illustrated in algorithms \ref{alg:the_alg2} and \ref{alg:the_alg3}.
\begin{algorithm}[t]
  \caption{Probabilistic Slot Attention (no $\mathbf{k}$ or $\mathbf{v}$)}
  \label{alg:the_alg2}
  \begin{algorithmic}
    \renewcommand{\baselinestretch}{1.25}\selectfont
    \STATE \textbf{Input:} $\mathbf{z} = f_e(\mathbf{x}) \in \mathbb{R}^{N\times d}$ \hfill {\color{gray}$\triangleright$ input representation}
    \vspace{2pt}
    \STATE $\forall k$, $\boldsymbol{\pi}(0)_k \gets 1/K$, $\boldsymbol{\mu}(0)_k \sim \mathcal{N}(0, \mathbf{I}_d)$, $\boldsymbol{\sigma}(0)_k^2 \gets \mathbf{1}_d$
    \vspace{2pt}
    \STATE \textbf{for} $t=0,\dots,T-1$
    \vspace{3pt}
    \STATE \quad $\displaystyle{A_{nk} \gets \frac{\boldsymbol{\pi}(t)_k\mathcal{N}\left(\mathbf{z}_n;\mW_q \boldsymbol{\mu}(t)_k, \boldsymbol{\sigma}(t)_k^2\right)}{\sum_{j=1}^{K}\boldsymbol{\pi}(t)_j\mathcal{N}\left(\mathbf{z}_n;\mW_q \boldsymbol{\mu}(t)_j, \boldsymbol{\sigma}(t)^2_j\right)}}$
    \vspace{3pt}
    \STATE \quad $\hat{A}_{nk} \gets A_{nk}/\sum_{l=1}^NA_{lk}$ \hfill {\color{gray}$\triangleright$ normalize attention}
    \vspace{2pt}
    \STATE \quad $\boldsymbol{\mu}(t+1)_k \gets \sum_{n=1}^N \hat{A}_{nk}\mathbf{z}_n$ \hfill {\color{gray}$\triangleright$ update slot mean}
    \vspace{2pt}
    \STATE \quad $\boldsymbol{\sigma}(t+1)_k^2 \gets \sum_{n=1}^N \hat{A}_{nk} \left(\mathbf{z}_n - \boldsymbol{\mu}(t+1)_k\right)^2$
    \vspace{2pt}
    \STATE \quad $\boldsymbol{\pi}(t+1)_k \gets \sum_{n=1}^N A_{nk}/N$ \hfill {\color{gray}$\triangleright$ update mixing coeff.}
    \vspace{2pt}
    \STATE \textbf{return} $\boldsymbol{\mu}(T), \boldsymbol{\sigma}(T)^2$ \hfill {\color{gray}$\triangleright$ $K$ slot distributions}
  \end{algorithmic}
\end{algorithm}
\begin{algorithm}[t]
  \caption{Probabilistic Slot Attention V.2 (\textsc{PSA-Proj})}
  \label{alg:the_alg3}
  \begin{algorithmic}
    \renewcommand{\baselinestretch}{1.25}\selectfont
    \STATE \textbf{Input:} $\mathbf{z} = f_e(\mathbf{x}) \in \mathbb{R}^{N\times d}$ 
    \hfill {\color{gray}$\triangleright$ input representation}
    \STATE $\mathbf{k} \gets \mW_k \mathbf{z} \in \mathbb{R}^{N \times d}$ \hfill {\color{gray}$\triangleright$ compute keys}
    \STATE $\forall k$, $\boldsymbol{\pi}(0)_k \gets 1/K$, $\boldsymbol{\mu}(0)_k \sim \mathcal{N}(0, \mathbf{I}_d)$, $\boldsymbol{\sigma}(0)_k^2 \gets \mathbf{1}_d$
    \vspace{2pt}
    \STATE \textbf{for} $t=0,\dots,T-1$
    \vspace{3pt}
    \STATE \quad $\displaystyle{A_{nk} \gets \frac{\boldsymbol{\pi}(t)_k\mathcal{N}\left(\mathbf{k}_n;\mW_q \boldsymbol{\mu}(t)_k, \boldsymbol{\sigma}(t)_k^2\right)}{\sum_{j=1}^{K}\boldsymbol{\pi}(t)_j\mathcal{N}\left(\mathbf{k}_n;\mW_q \boldsymbol{\mu}(t)_j, \boldsymbol{\sigma}(t)_j^2\right)}}$
    \vspace{3pt}
    \STATE \quad $\hat{A}_{nk} \gets A_{nk}/\sum_{l=1}^NA_{lk}$ \hfill {\color{gray}$\triangleright$ normalize attention}
    \vspace{2pt}
    \STATE \quad $\boldsymbol{\mu}(t+1)_k \gets \sum_{n=1}^N \hat{A}_{nk}{\color{blue}\mW_v \mathbf{z}_n}$ \hfill {\color{gray}$\triangleright$ update slot mean}
    \vspace{2pt}
    \STATE \quad $\boldsymbol{\sigma}(t+1)_k^2 \gets \sum_{n=1}^N \hat{A}_{nk} \left({\color{blue}\mW_v \mathbf{z}_n} - \boldsymbol{\mu}(t+1)_k\right)^2$
    \vspace{2pt}
    \STATE \quad $\boldsymbol{\pi}(t+1)_k \gets \sum_{n=1}^N A_{nk}/N$ \hfill {\color{gray} $\triangleright$ update mixing coeff.}
    \vspace{2pt}
    \STATE \textbf{return} $\boldsymbol{\mu}(T), \boldsymbol{\sigma}(T)^2$ \hfill {\color{gray}$\triangleright$ $K$ slot distributions}
  \end{algorithmic}
\end{algorithm}

\section{Metrics}
\label{appendix:metrics}

\begin{figure}[!ht]
    \centering
    \begin{subfigure}{.025\columnwidth}
        \rotatebox{90}{\hspace{8pt} \scriptsize \textbf{Slot Mean Correlation Coefficient}}
    \end{subfigure} \hspace{1pt}
    \begin{subfigure}{.425\columnwidth}
        \includegraphics[trim={27 33 0 0},clip,width=\columnwidth]{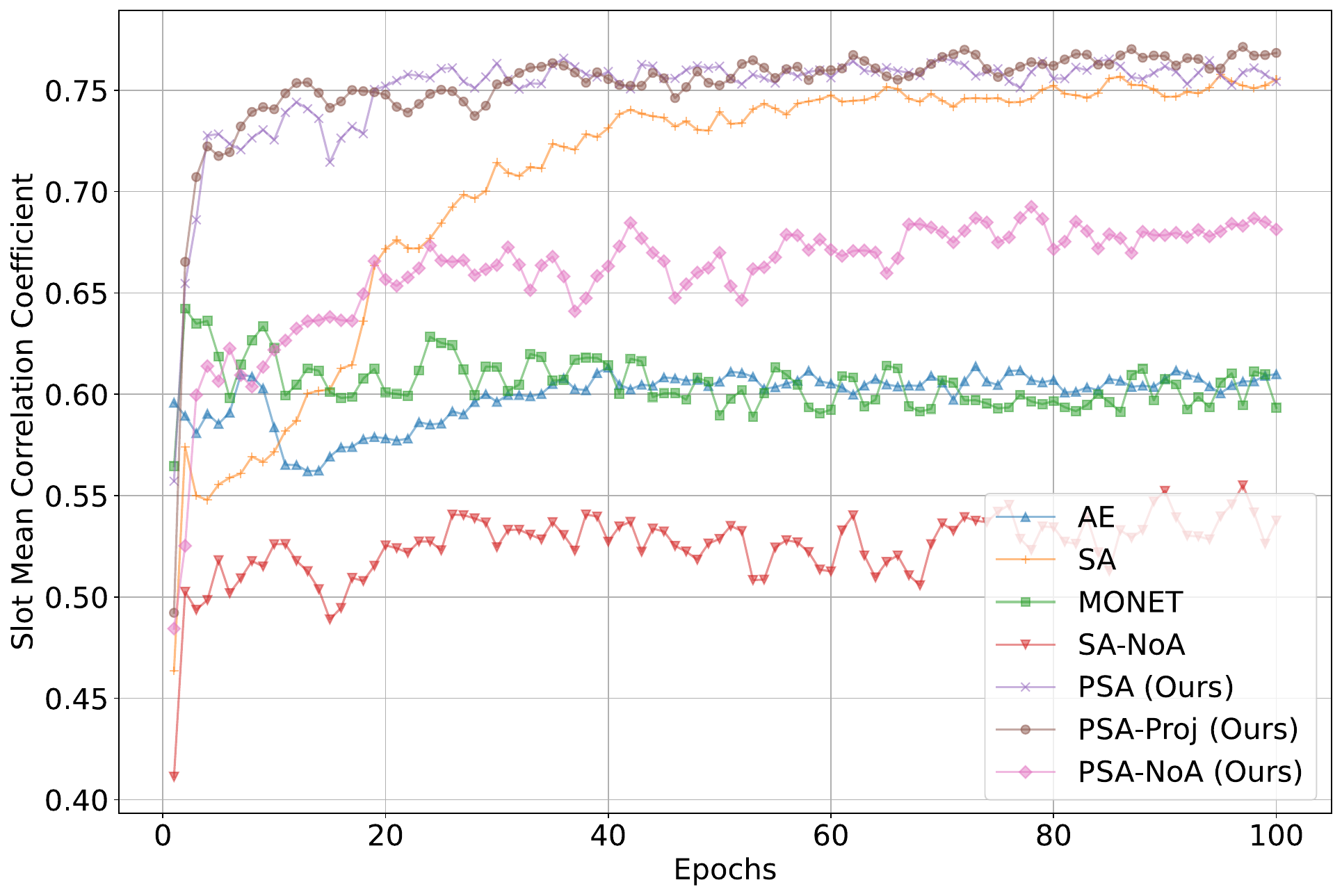} 
    \end{subfigure} 
    \hfill
    \begin{subfigure}{.025\columnwidth}
        \rotatebox{90}{\hspace{25pt} \scriptsize \textbf{Slot Identifiability Score}}
    \end{subfigure} \hspace{1pt}
    \begin{subfigure}{.425\columnwidth}
        \includegraphics[trim={27 33 0 0},clip,width=\columnwidth]{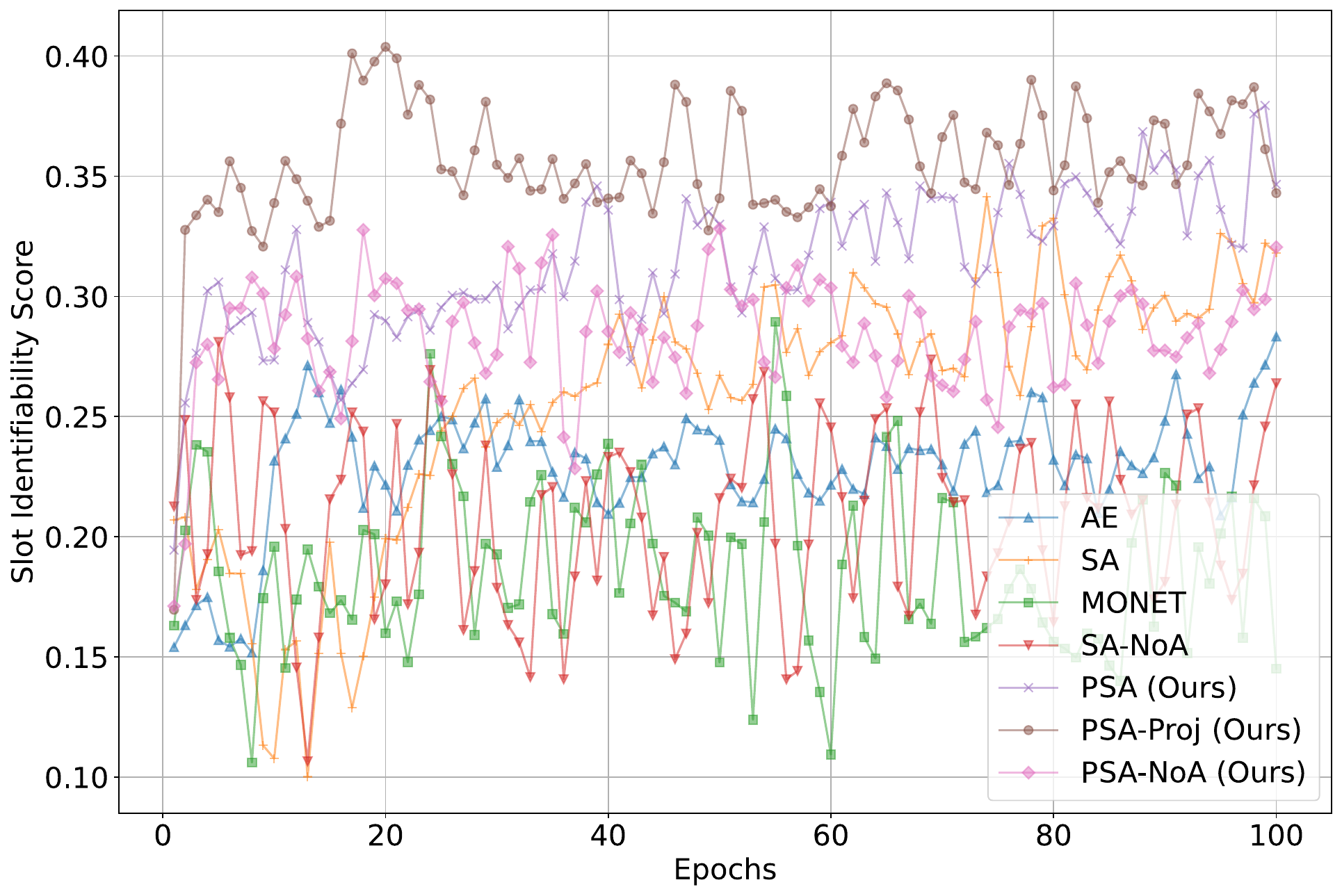}
    \end{subfigure}
    \begin{center}
        \vspace{-3pt}
        \scriptsize  \textbf{Epoch} \hspace{200pt} \textbf{Epoch}
    \end{center}
    \vspace{-5pt}
    \caption{
    \textbf{Identifiability scores throughout training.} Our proposed SMCC metric (top) is much more stable than SIS (bottom) in capturing identifiability, and better in discerning differences between methods, showing substantial improvements for PSA.
    }
    \label{fig:results_sis_smcc_epochs}
\end{figure}

\paragraph{SIS:} Slot identifiability score \cite{brady2023provably}, mainly focus on R2 score between ground-truth and the estimated slot representations wrt to maximum R2 score from the models fit between each and every inferred slots.
By design SIS requires a model fitting at every validation step to compute this relativistic measure, due to which the metric seems to vary quite a bit across runs, as observed in Figure \ref{fig:results_sis_smcc_epochs}.

\paragraph{SMCC:} Mean correlation coefficient is a well studied metric in disentangled representational learning \cite{khemakhem2020ice, khemakhem2020variational}, we extend this with additional permutational invariance on slot dimensions.
SMCC measures the correlation between estimated and ground truth representations (or estimated representations across runs, in the case when ground truth representations are unavailable) once the slots are matched using Hungarian matching. To compute identifiability up to affine transformation along the representational axis and permutation in slots (check definition \ref{dfn:sequivalence}), similar to weak identifiability as per the definition in the paper and in \cite{Khemakhem2020_ice}, we use the MCC up to some affine mapping $\mA$, which we learned by matching slot representations across runs.
In summary, SMCC can be computed with the following three steps:

\begin{itemize}
    \item Matching the order of slots using Hungarian matching;
    \item Affine transformation of slot representation;
    \item Followed by computing mean correlation.
\end{itemize}

Apart from variations described in Figure \ref{fig:results_sis_smcc_epochs}, we further analyse both the metrics by fixing the model and data; and by computing both the metrics for 10 times. 
We then considered the mean and variance in the performance, which is reflected as follows: SIS: {35.26} $\pm$ \textbf{6.46}, SMCC: {77.83} $\pm$ \textbf{0.36}. 
Here, the resulting variation is only the reflection of the metric, which clearly indicates the stability of SMCC over SIS.

\begin{table*}[!t]
\footnotesize
\centering
\caption{Quality of compositional image generation with FID measure}
\vspace{3pt}
\begin{tabular}{@{}lcccccc@{}}
\toprule
\textsc{Method} & \multicolumn{3}{c}{CLEVR} & \multicolumn{3}{c}{\textsc{Objects-Room}}
\\ 
\cmidrule(l){2-7} 
& FG-ARI $\uparrow$   & CFID $\downarrow$   & RFID $\downarrow$   & FG-ARI $\uparrow$  & CFID $\downarrow$   & RFID $\downarrow$ \\ 
\midrule
SA                     & $0.96 \pm 0.01$  & - & $41.81 \pm 2.82$ & 
                         $0.79 \pm 0.06$ & - & $16.49 \pm 3.34$ \\
SA-NoA                  & $0.62 \pm 0.02$  & - & $81.21 \pm 8.72$ & 
                        $0.41 \pm 0.12$ & - & $96.64 \pm 21.33$\\
\midrule
\textsc{PSA-NoA}        & $0.84 \pm 0.01$ & $36.42 \pm 8.53$ & $68.02 \pm 10.21$ 
                        & $0.78 \pm 0.04$ & $54.55 \pm 1.43$ & $21.37 \pm 1.03$ \\
\textsc{PSA}            & $0.85 \pm 0.02$ & $28.50 \pm 4.27$ & $52.70 \pm 1.74$  
                        & $0.78 \pm 0.02$ & $ 20.49 \pm 2.36$ & $21.00 \pm 1.43$ \\
\textsc{PSA-Proj}  & $0.95 \pm 0.00$ & $28.79 \pm 5.50$ & $39.42 \pm 8.29 $  
                        & $0.81 \pm 0.04$ & $34.58 \pm 4.32$ & $16.85 \pm 5.68$\\
\bottomrule
\end{tabular}
\label{table:compositionality_results}
\end{table*}
\begin{figure*}[!h]
    \centering
    \includegraphics[width=.95\textwidth, trim={40cm 29.5cm 0 49.5cm}, clip]{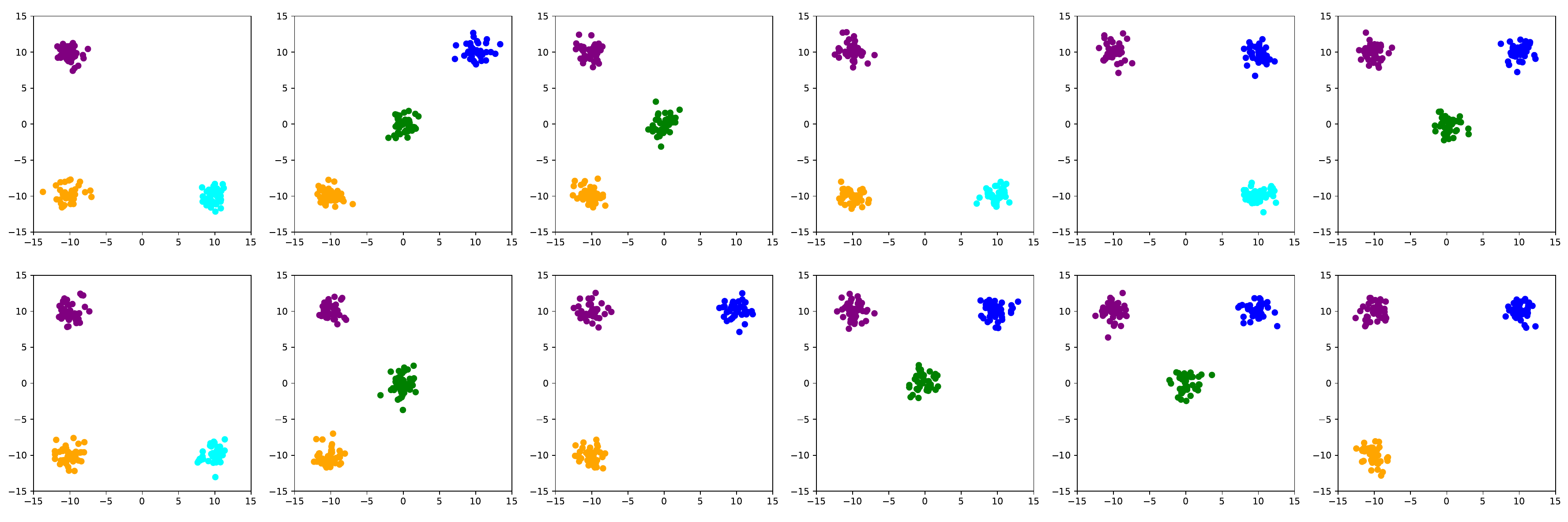}
    \caption{Random samples from the 2D synthetic dataset used in our aggregate posterior identifiability experiments. As outlined in the main text, there are in total five `object' clusters in the dataset, and each observation contains at most three of the clusters.}
    \label{fig:datasamples}
\end{figure*}

\section{Comparison with Autoencoding Variational Bayes}
\label{appendix:vae_discuss}

As explained in Section~\ref{sec:background}, applying slot attention to a deterministic encoding $\mathbf{z} = f_e(\mathbf{x}) \in \mathbb{R}^{N \times d}$ yields a set of $K$ object slot representations $\mathbf{s}_{1:K} \coloneqq \mathbf{s}_1,\dots,\mathbf{s}_K$. In combination, this process induces a stochastic encoder $q(\mathbf{s}_{1:K} \mid \mathbf{x})$, where the stochasticity comes from the random initialization of the slots in the first iteration: $\mathbf{s}^{t=0}_{1:K} \sim \mathcal{N}(\mathbf{s}_{1:K}; 0, \mathbf{I}) \in \mathbb{R}^{K \times d}$. Since each slot is a deterministic function of its previous state $\mathbf{s}^{t} \coloneqq f_s(\mathbf{z}, \mathbf{s}^{t-1})$ it is possible to randomly sample initial states $\mathbf{s}^0$ and obtain stochastic estimates of the slots. However, since each transition depends on $\mathbf{z}$, which in turn depends on the input $\mathbf{x}$, \textit{we do not get a generative model we can tractably sample from}. 

This can be remedied by placing a tractable prior over $\mathbf{z}$ and using the VAE framework along the lines of~\cite{pmlr-v202-wang23r}. Specifically, Wang et al.~\cite{pmlr-v202-wang23r} propose the Slot-VAE, which is a generative model that integrates slot attention and the VAE framework under a two-layer hierarchical latent model. 
However, under their formulation, there is a key challenge in calculating the KL term as the slot attention function is permutation equivariant meaning the slots have no fixed order across the posterior and prior. To compensate for this, the authors introduce a heuristic auxiliary loss and a parallel image processing path with an additional slot attention operation which is computationally costly.

In contrast, our approach does not suffer from such drawbacks. This was achieved by designing the slot attention operation itself as probabilistic, and proving that having a local (per-datapoint) GMM results in the \textit{aggregate slot posterior} and the concatenated slots being GMM distributed (Section~\ref{sec:expectation_maximization}). We then proved a new slot identifiability result using this insight (Section~\ref{sec:identifiability}). Our use of the aggregate posterior is inspired by but differs substantially in both method and application from previous works on VAEs~\cite{hoffman2016elbo,tomczak2018vae,van2017neural}. Our primary goal is not to learn a better generative prior but to obtain slot identifiable representations. Lastly, since the concatenated slots are provably GMM distributed, our model is reminiscent of a GMM-VAE~\cite{dilokthanakul2016deep}, but with a unique slot-based structure.

\section{Automatic Relevance Determination of Slots}
\label{appendix:ard_results}

For evaluation, we calculate the MAE between the estimated and ground truth numbers of slots and measure the reduction in FLOPs achieved using the proposed ARD method. We observe MAE values of 1.03 and 0.58, and significant savings of up to \textbf{41\%} and \textbf{62\%} in FLOPs on the CLEVR and Objects-Room datasets respectively, when compared to using a fixed number of slots $K$.

\begin{figure*}[t]
    \centering
    \subfloat{\includegraphics[width=.99\textwidth]{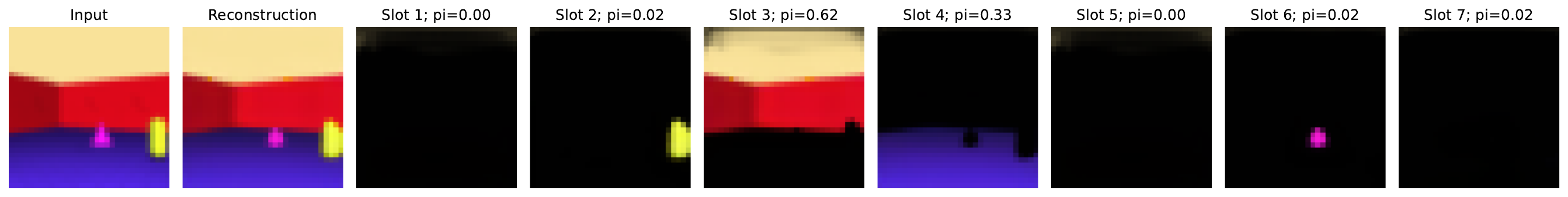}} \hfill
    \subfloat{\includegraphics[width=.99\textwidth]{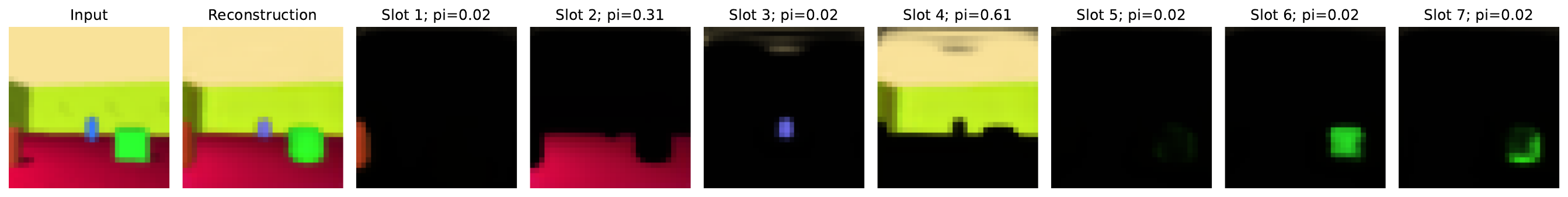}} \hfill
    \subfloat{\includegraphics[width=.99\textwidth]{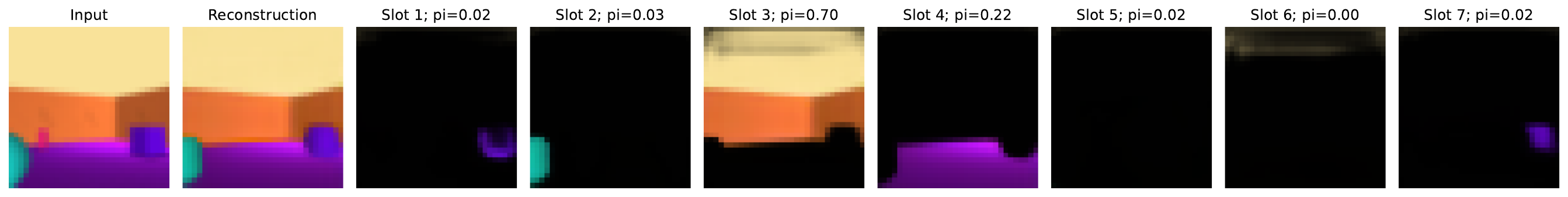}} \hfill
    \subfloat{\includegraphics[width=.99\textwidth]{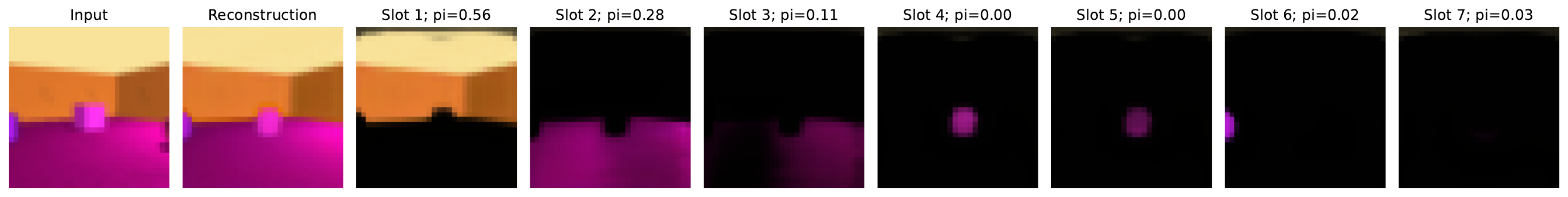}} \hfill
    \subfloat{\includegraphics[width=.99\textwidth]{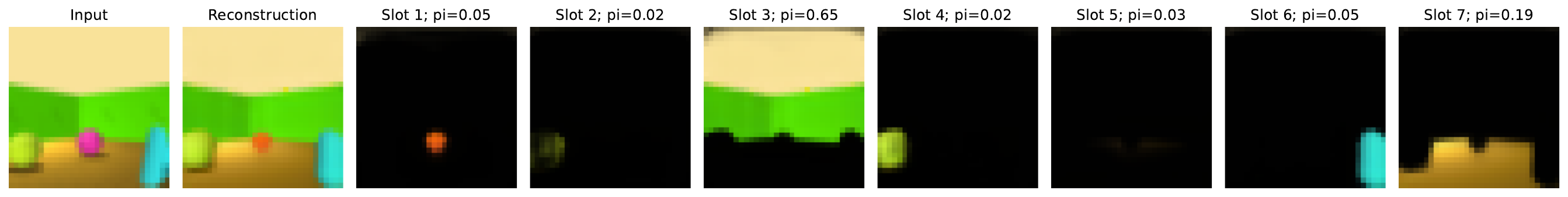}} \hfill
    \subfloat{\includegraphics[width=.99\textwidth]{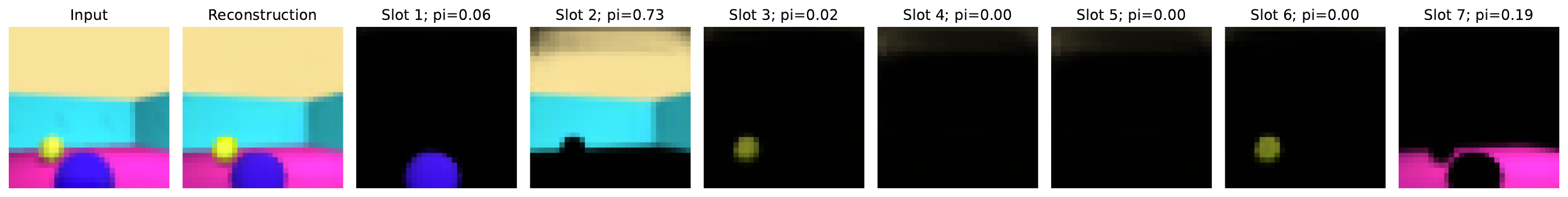}} \hfill
    \subfloat{\includegraphics[width=.99\textwidth]{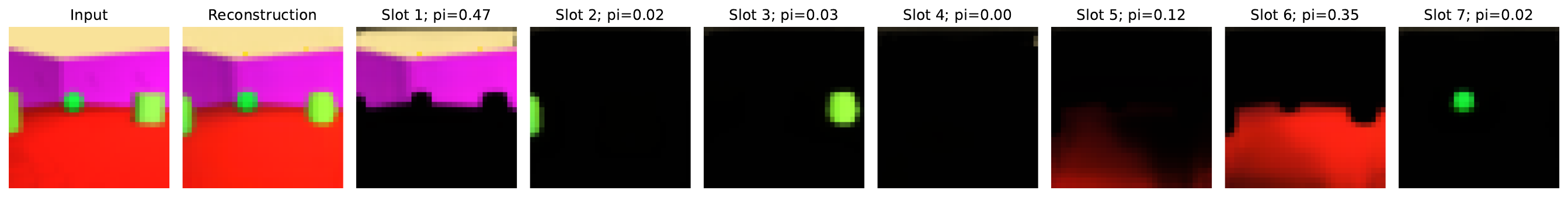}} \hfill
    \subfloat{\includegraphics[width=.99\textwidth]{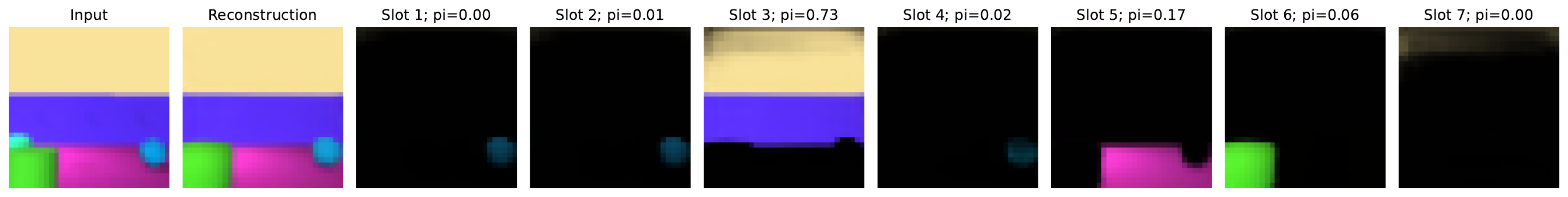}} \hfill
    \subfloat{\includegraphics[width=.99\textwidth]{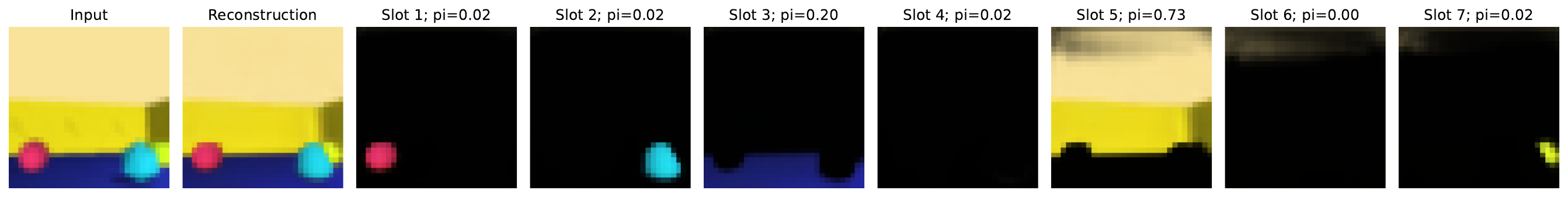}} \hfill
    \caption{Automatic Relevance Determination (ARD) of slots on the \textsc{ObjectsRoom} dataset. As shown, when using our proposed probabilistic slot attention (Algorithm~\ref{alg:the_alg}), the mixing coefficients $\boldsymbol{\pi}_{i} \in \mathbb{R}^K, \forall i$ automatically approach zero when slots are inactive.}
    \label{fig:ard_or_results}
\end{figure*}

\newpage
\section{Compositionality Results}
\label{appendix:compositionality_results}
\begin{figure*}[!h]
    \centering
    \subfloat{\includegraphics[width=1.0\textwidth]{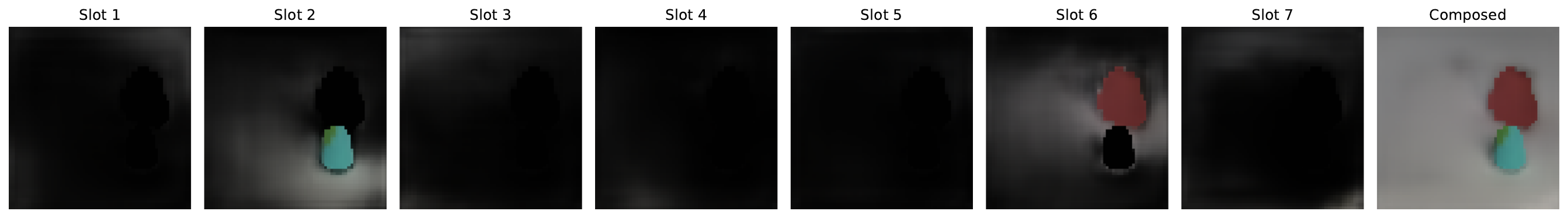}} \hfill
    \subfloat{\includegraphics[width=1.0\textwidth]{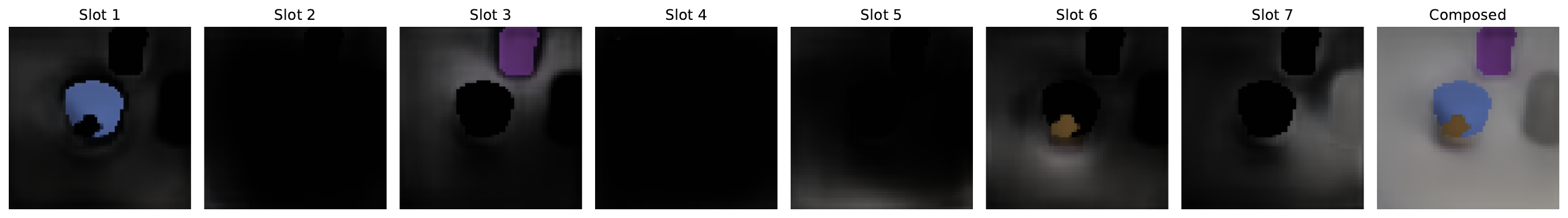}} \hfill
    \subfloat{\includegraphics[width=1.0\textwidth]{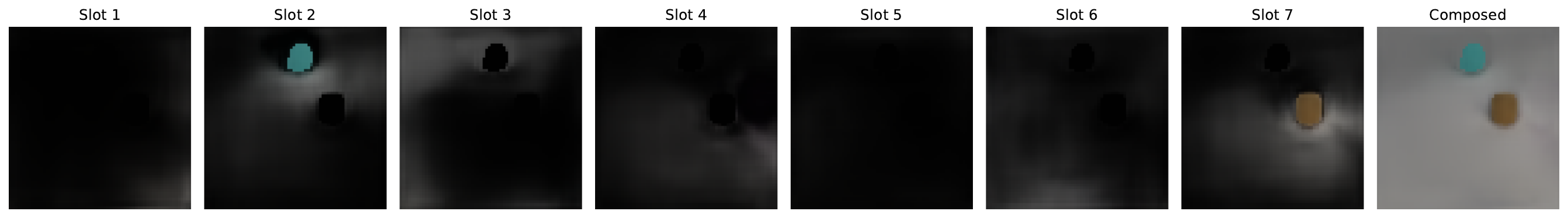}} \hfill
    \subfloat{\includegraphics[width=1.0\textwidth]{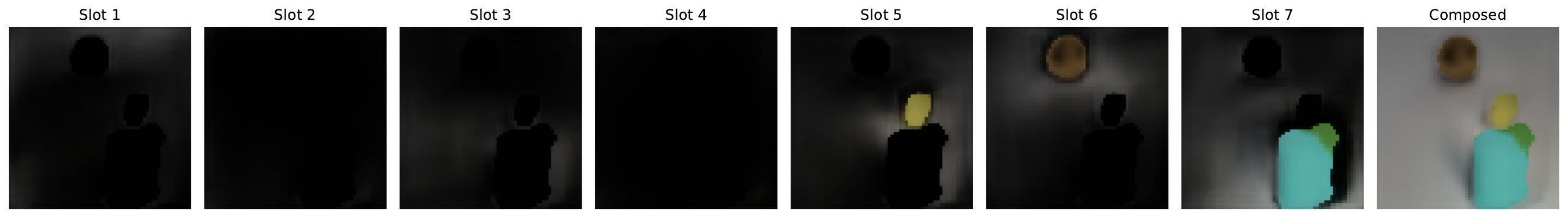}} \hfill
    \subfloat{\includegraphics[width=1.0\textwidth]{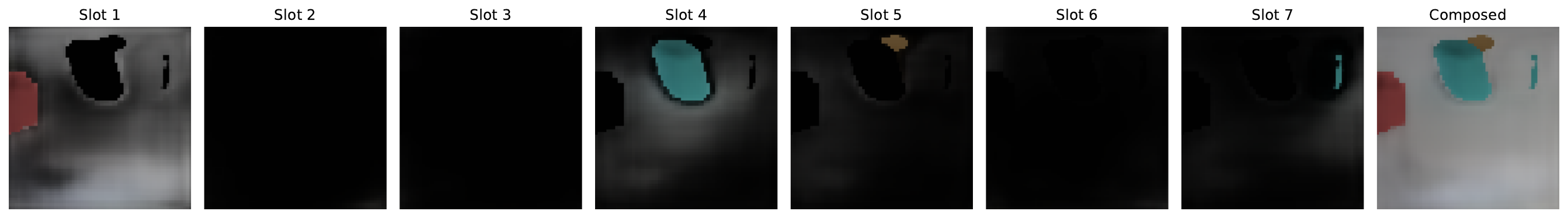}} \hfill
    \caption{Aggregate posterior sampling for image composition on CLEVR dataset.}
    \label{fig:composition_clevr_results}
\end{figure*}
\begin{figure*}[!h]
    \centering
    \subfloat{\includegraphics[width=1.0\textwidth]{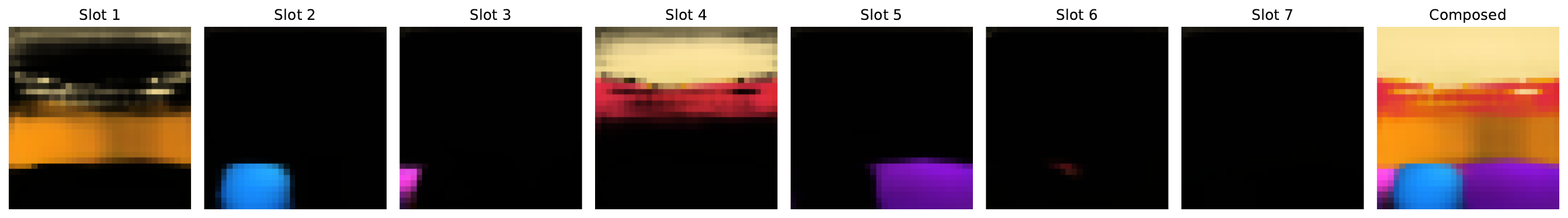}} \hfill
    \subfloat{\includegraphics[width=1.0\textwidth]{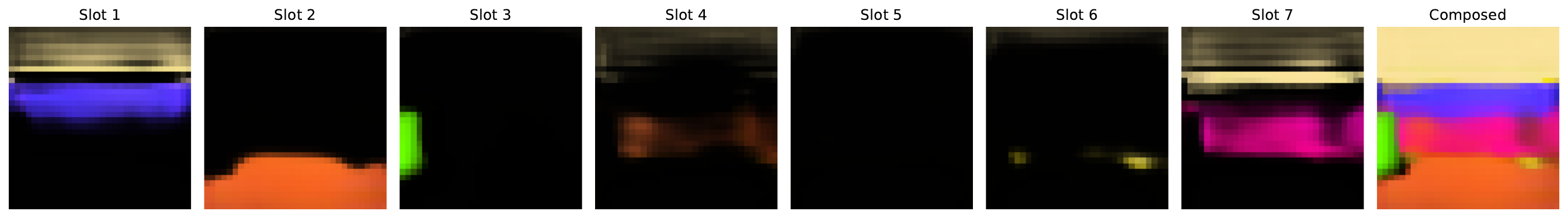}} \hfill
    \subfloat{\includegraphics[width=1.0\textwidth]{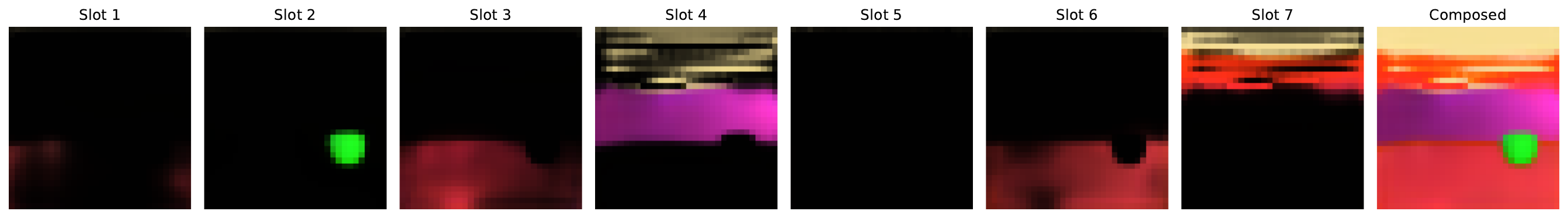}} \hfill
    \subfloat{\includegraphics[width=1.0\textwidth]{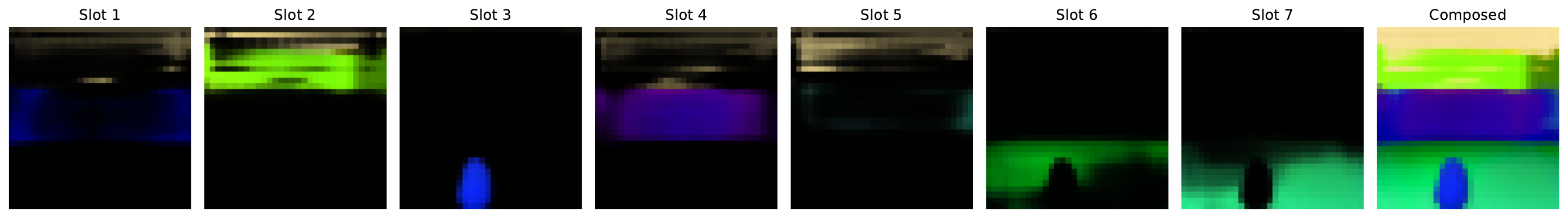}} \hfill
    \subfloat{\includegraphics[width=1.0\textwidth]{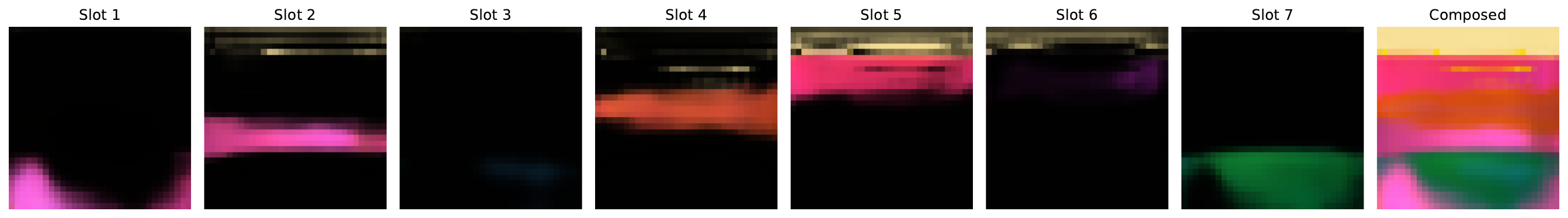}} \hfill
    \caption{Aggregate posterior sampling for image composition on \textsc{ObjectsRoom} dataset.}
    \label{fig:composition_or_results}
\end{figure*}
\newpage

\begin{figure*}[t]
    \centering
    \includegraphics[width=.99\textwidth]{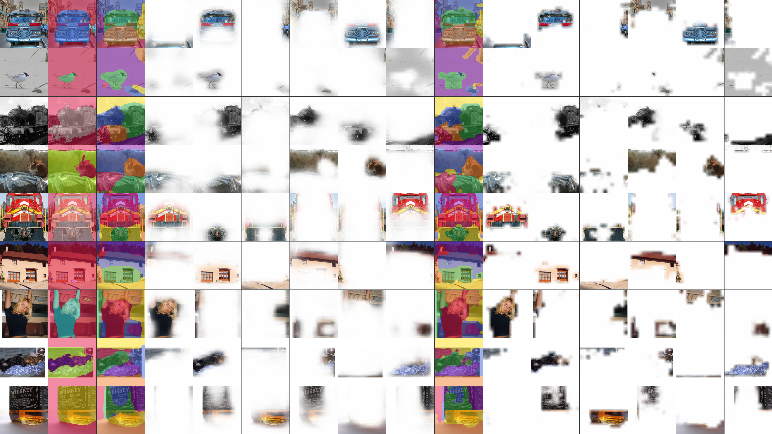} 
    \caption{Visualizations of attention and alpha masks on the Pascal VOC2012 dataset are shown, with alpha masks on the left and attention masks on the right, for a PSA-MLP model using a DINO feature extractor. In the figure above, the images from left to right represent: the original image, ground-truth segmentation, alpha mask segmentation, individual entities grouped in the alpha mask, slot attention segmentation mask, and individual entities grouped in the slot attention mask.}
    \label{fig:psa-mlp-results}
\end{figure*}

\begin{figure*}[t]
    \centering
    \includegraphics[width=.99\textwidth]{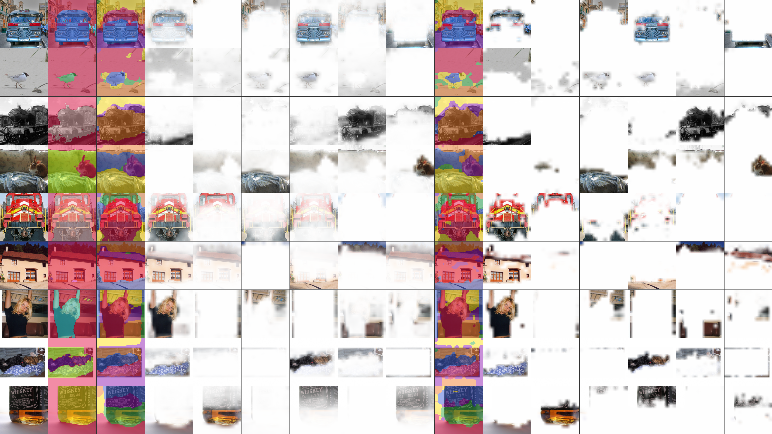} 
    \caption{Visualizations of attention and alpha masks on the Pascal VOC2012 dataset are shown, with alpha masks on the left and attention masks on the right, for a PSA-Transformer model using a DINO feature extractor. In the figure above, the images from left to right represent: the original image, ground-truth segmentation, alpha mask segmentation, individual entities grouped in the alpha mask, slot attention segmentation mask, and individual entities grouped in the slot attention mask.}
    \label{fig:psa-transformers-results}
\end{figure*}

%% file: main.bbl
\begin{thebibliography}{10}

\bibitem{ahuja2022weakly}
Kartik Ahuja, Jason~S Hartford, and Yoshua Bengio.
\newblock Weakly supervised representation learning with sparse perturbations.
\newblock {\em Advances in Neural Information Processing Systems}, 35:15516--15528, 2022.

\bibitem{ahuja2022interventional}
Kartik Ahuja, Yixin Wang, Divyat Mahajan, and Yoshua Bengio.
\newblock Interventional causal representation learning.
\newblock {\em arXiv preprint arXiv:2209.11924}, 2022.

\bibitem{battaglia2018relational}
Peter~W Battaglia, Jessica~B Hamrick, Victor Bapst, Alvaro Sanchez-Gonzalez, Vinicius Zambaldi, Mateusz Malinowski, Andrea Tacchetti, David Raposo, Adam Santoro, Ryan Faulkner, et~al.
\newblock Relational inductive biases, deep learning, and graph networks.
\newblock {\em arXiv preprint arXiv:1806.01261}, 2018.

\bibitem{behrens2018cognitive}
Timothy~EJ Behrens, Timothy~H Muller, James~CR Whittington, Shirley Mark, Alon~B Baram, Kimberly~L Stachenfeld, and Zeb Kurth-Nelson.
\newblock What is a cognitive map? organizing knowledge for flexible behavior.
\newblock {\em Neuron}, 100(2):490--509, 2018.

\bibitem{bengio2013representation}
Yoshua Bengio, Aaron Courville, and Pascal Vincent.
\newblock Representation learning: A review and new perspectives.
\newblock {\em IEEE transactions on pattern analysis and machine intelligence}, 35(8):1798--1828, 2013.

\bibitem{brady2023provably}
Jack Brady, Roland~S Zimmermann, Yash Sharma, Bernhard Sch{\"o}lkopf, Julius von K{\"u}gelgen, and Wieland Brendel.
\newblock Provably learning object-centric representations.
\newblock {\em arXiv preprint arXiv:2305.14229}, 2023.

\bibitem{brehmer2022weakly}
Johann Brehmer, Pim De~Haan, Phillip Lippe, and Taco~S Cohen.
\newblock Weakly supervised causal representation learning.
\newblock {\em Advances in Neural Information Processing Systems}, 35:38319--38331, 2022.

\bibitem{buchholz2022function}
Simon Buchholz, Michel Besserve, and Bernhard Sch{\"o}lkopf.
\newblock Function classes for identifiable nonlinear independent component analysis.
\newblock In Alice~H. Oh, Alekh Agarwal, Danielle Belgrave, and Kyunghyun Cho, editors, {\em Advances in Neural Information Processing Systems}, 2022.

\bibitem{burgess2019monet}
Christopher~P Burgess, Loic Matthey, Nicholas Watters, Rishabh Kabra, Irina Higgins, Matt Botvinick, and Alexander Lerchner.
\newblock Monet: Unsupervised scene decomposition and representation.
\newblock {\em arXiv preprint arXiv:1901.11390}, 2019.

\bibitem{chang2022object}
Michael Chang, Thomas~L Griffiths, and Sergey Levine.
\newblock Object representations as fixed points: Training iterative refinement algorithms with implicit differentiation.
\newblock {\em arXiv preprint arXiv:2207.00787}, 2022.

\bibitem{dilokthanakul2016deep}
Nat Dilokthanakul, Pedro~AM Mediano, Marta Garnelo, Matthew~CH Lee, Hugh Salimbeni, Kai Arulkumaran, and Murray Shanahan.
\newblock Deep unsupervised clustering with gaussian mixture variational autoencoders.
\newblock {\em arXiv preprint arXiv:1611.02648}, 2016.

\bibitem{eastwood2018a}
Cian Eastwood and Christopher K.~I. Williams.
\newblock A framework for the quantitative evaluation of disentangled representations.
\newblock In {\em International Conference on Learning Representations}, 2018.

\bibitem{elsayed2022savi++}
Gamaleldin Elsayed, Aravindh Mahendran, Sjoerd van Steenkiste, Klaus Greff, Michael~C Mozer, and Thomas Kipf.
\newblock Savi++: Towards end-to-end object-centric learning from real-world videos.
\newblock {\em Advances in Neural Information Processing Systems}, 35:28940--28954, 2022.

\bibitem{emami2022slot}
Patrick Emami, Pan He, Sanjay Ranka, and Anand Rangarajan.
\newblock Slot order matters for compositional scene understanding.
\newblock {\em arXiv preprint arXiv:2206.01370}, 2022.

\bibitem{engelcke2019genesis}
Martin Engelcke, Adam~R Kosiorek, Oiwi~Parker Jones, and Ingmar Posner.
\newblock Genesis: Generative scene inference and sampling with object-centric latent representations.
\newblock {\em arXiv preprint arXiv:1907.13052}, 2019.

\bibitem{engelcke2021genesis}
Martin Engelcke, Oiwi Parker~Jones, and Ingmar Posner.
\newblock Genesis-v2: Inferring unordered object representations without iterative refinement.
\newblock {\em Advances in Neural Information Processing Systems}, 34:8085--8094, 2021.

\bibitem{epstein2017cognitive}
Russell~A Epstein, Eva~Zita Patai, Joshua~B Julian, and Hugo~J Spiers.
\newblock The cognitive map in humans: spatial navigation and beyond.
\newblock {\em Nature neuroscience}, 20(11):1504--1513, 2017.

\bibitem{falck2021multi}
Fabian Falck, Haoting Zhang, Matthew Willetts, George Nicholson, Christopher Yau, and Chris~C Holmes.
\newblock Multi-facet clustering variational autoencoders.
\newblock {\em Advances in Neural Information Processing Systems}, 34:8676--8690, 2021.

\bibitem{gerstenberg2021counterfactual}
Tobias Gerstenberg, Noah~D Goodman, David~A Lagnado, and Joshua~B Tenenbaum.
\newblock A counterfactual simulation model of causal judgments for physical events.
\newblock {\em Psychological review}, 128(5):936, 2021.

\bibitem{gopnik2004theory}
Alison Gopnik, Clark Glymour, David~M Sobel, Laura~E Schulz, Tamar Kushnir, and David Danks.
\newblock A theory of causal learning in children: causal maps and bayes nets.
\newblock {\em Psychological review}, 111(1):3, 2004.

\bibitem{greff2019multi}
Klaus Greff, Rapha{\"e}l~Lopez Kaufman, Rishabh Kabra, Nick Watters, Christopher Burgess, Daniel Zoran, Loic Matthey, Matthew Botvinick, and Alexander Lerchner.
\newblock Multi-object representation learning with iterative variational inference.
\newblock In {\em International Conference on Machine Learning}, pages 2424--2433. PMLR, 2019.

\bibitem{greff2020binding}
Klaus Greff, Sjoerd Van~Steenkiste, and J{\"u}rgen Schmidhuber.
\newblock On the binding problem in artificial neural networks.
\newblock {\em arXiv preprint arXiv:2012.05208}, 2020.

\bibitem{heusel2017gans}
Martin Heusel, Hubert Ramsauer, Thomas Unterthiner, Bernhard Nessler, and Sepp Hochreiter.
\newblock Gans trained by a two time-scale update rule converge to a local nash equilibrium.
\newblock {\em Advances in neural information processing systems}, 30, 2017.

\bibitem{higgins2017betavae}
Irina Higgins, Loic Matthey, Arka Pal, Christopher Burgess, Xavier Glorot, Matthew Botvinick, Shakir Mohamed, and Alexander Lerchner.
\newblock beta-{VAE}: Learning basic visual concepts with a constrained variational framework.
\newblock In {\em International Conference on Learning Representations}, 2017.

\bibitem{hinton1979some}
Geoffrey Hinton.
\newblock Some demonstrations of the effects of structural descriptions in mental imagery.
\newblock {\em Cognitive Science}, 3(3):231--250, 1979.

\bibitem{hinton2022represent}
Geoffrey Hinton.
\newblock How to represent part-whole hierarchies in a neural network.
\newblock {\em Neural Computation}, pages 1--40, 2022.

\bibitem{hinton2018matrix}
Geoffrey~E Hinton, Sara Sabour, and Nicholas Frosst.
\newblock Matrix capsules with em routing.
\newblock In {\em International conference on learning representations}, 2018.

\bibitem{hoffman2016elbo}
Matthew~D Hoffman and Matthew~J Johnson.
\newblock Elbo surgery: yet another way to carve up the variational evidence lower bound.
\newblock In {\em Workshop in Advances in Approximate Bayesian Inference, NIPS}, volume~1, 2016.

\bibitem{hyvarinen2000independent}
A~Hyvarinen and E~Oja.
\newblock Independent component analysis: algorithms and applications.
\newblock {\em Neural Networks}, 13(4-5):411--430, 2000.

\bibitem{hyvarinen2016unsupervised}
Aapo Hyvarinen and Hiroshi Morioka.
\newblock Unsupervised feature extraction by time-contrastive learning and nonlinear ica.
\newblock {\em Advances in neural information processing systems}, 29, 2016.

\bibitem{hyvarinen1999nonlinear}
Aapo Hyv{\"a}rinen and Petteri Pajunen.
\newblock Nonlinear independent component analysis: Existence and uniqueness results.
\newblock {\em Neural networks}, 12(3):429--439, 1999.

\bibitem{hyvarinen2019nonlinear}
Aapo Hyvarinen, Hiroaki Sasaki, and Richard Turner.
\newblock Nonlinear ica using auxiliary variables and generalized contrastive learning.
\newblock In {\em The 22nd International Conference on Artificial Intelligence and Statistics}, pages 859--868. PMLR, 2019.

\bibitem{pmlr-v89-hyvarinen19a}
Aapo Hyvarinen, Hiroaki Sasaki, and Richard Turner.
\newblock Nonlinear ica using auxiliary variables and generalized contrastive learning.
\newblock In {\em Proceedings of the Twenty-Second International Conference on Artificial Intelligence and Statistics}, volume~89, pages 859--868. PMLR, 2019.

\bibitem{johnson2017clevr}
Justin Johnson, Bharath Hariharan, Laurens Van Der~Maaten, Li~Fei-Fei, C~Lawrence~Zitnick, and Ross Girshick.
\newblock Clevr: A diagnostic dataset for compositional language and elementary visual reasoning.
\newblock In {\em Proceedings of the IEEE conference on computer vision and pattern recognition}, pages 2901--2910, 2017.

\bibitem{multiobjectdatasets19}
Rishabh Kabra, Chris Burgess, Loic Matthey, Raphael~Lopez Kaufman, Klaus Greff, Malcolm Reynolds, and Alexander Lerchner.
\newblock Multi-object datasets.
\newblock https://github.com/deepmind/multi-object-datasets/, 2019.

\bibitem{kakogeorgiou2024spot}
Ioannis Kakogeorgiou, Spyros Gidaris, Konstantinos Karantzalos, and Nikos Komodakis.
\newblock Spot: Self-training with patch-order permutation for object-centric learning with autoregressive transformers.
\newblock In {\em Proceedings of the IEEE/CVF Conference on Computer Vision and Pattern Recognition}, pages 22776--22786, 2024.

\bibitem{khemakhem2020variational}
Ilyes Khemakhem, Diederik Kingma, Ricardo Monti, and Aapo Hyvarinen.
\newblock Variational autoencoders and nonlinear ica: A unifying framework.
\newblock In {\em International Conference on Artificial Intelligence and Statistics}, pages 2207--2217. PMLR, 2020.

\bibitem{Khemakhem2020_ice}
Ilyes Khemakhem, Ricardo Monti, Diederik Kingma, and Aapo Hyvarinen.
\newblock Ice-beem: Identifiable conditional energy-based deep models based on nonlinear ica.
\newblock In {\em Advances in Neural Information Processing Systems}, volume~33, 2020.

\bibitem{khemakhem2020ice}
Ilyes Khemakhem, Ricardo Monti, Diederik Kingma, and Aapo Hyvarinen.
\newblock Ice-beem: Identifiable conditional energy-based deep models based on nonlinear ica.
\newblock {\em Advances in Neural Information Processing Systems}, 33:12768--12778, 2020.

\bibitem{pmlr-v80-kim18b}
Hyunjik Kim and Andriy Mnih.
\newblock Disentangling by factorising.
\newblock In {\em Proceedings of the 35th International Conference on Machine Learning}, Proceedings of Machine Learning Research, 2018.

\bibitem{kingma2013auto}
Diederik~P Kingma and Max Welling.
\newblock Auto-encoding variational bayes.
\newblock {\em arXiv preprint arXiv:1312.6114}, 2013.

\bibitem{kipf2021conditional}
Thomas Kipf, Gamaleldin~F Elsayed, Aravindh Mahendran, Austin Stone, Sara Sabour, Georg Heigold, Rico Jonschkowski, Alexey Dosovitskiy, and Klaus Greff.
\newblock Conditional object-centric learning from video.
\newblock {\em arXiv preprint arXiv:2111.12594}, 2021.

\bibitem{kivva2022identifiability}
Bohdan Kivva, Goutham Rajendran, Pradeep Ravikumar, and Bryon Aragam.
\newblock Identifiability of deep generative models without auxiliary information.
\newblock {\em Advances in Neural Information Processing Systems}, 35:15687--15701, 2022.

\bibitem{kori2023grounded}
Avinash Kori, Francesco Locatello, Fabio De~Sousa Ribeiro, Francesca Toni, and Ben Glocker.
\newblock Grounded object centric learning.
\newblock {\em arXiv preprint arXiv:2307.09437}, 2023.

\bibitem{kulkarni2015deep}
Tejas~D Kulkarni, William~F Whitney, Pushmeet Kohli, and Josh Tenenbaum.
\newblock Deep convolutional inverse graphics network.
\newblock {\em Advances in neural information processing systems}, 28, 2015.

\bibitem{lachapelle2024nonparametric}
S{\'e}bastien Lachapelle, Pau~Rodr{\'\i}guez L{\'o}pez, Yash Sharma, Katie Everett, R{\'e}mi~Le Priol, Alexandre Lacoste, and Simon Lacoste-Julien.
\newblock Nonparametric partial disentanglement via mechanism sparsity: Sparse actions, interventions and sparse temporal dependencies.
\newblock {\em arXiv preprint arXiv:2401.04890}, 2024.

\bibitem{lachapelle2024additive}
S{\'e}bastien Lachapelle, Divyat Mahajan, Ioannis Mitliagkas, and Simon Lacoste-Julien.
\newblock Additive decoders for latent variables identification and cartesian-product extrapolation.
\newblock {\em Advances in Neural Information Processing Systems}, 36, 2024.

\bibitem{lake2017building}
Brenden~M Lake, Tomer~D Ullman, Joshua~B Tenenbaum, and Samuel~J Gershman.
\newblock Building machines that learn and think like people.
\newblock {\em Behavioral and brain sciences}, 40:e253, 2017.

\bibitem{li2018tell}
Kunpeng Li, Ziyan Wu, Kuan-Chuan Peng, Jan Ernst, and Yun Fu.
\newblock Tell me where to look: Guided attention inference network.
\newblock In {\em Proceedings of the IEEE conference on computer vision and pattern recognition}, pages 9215--9223, 2018.

\bibitem{lin2020space}
Zhixuan Lin, Yi-Fu Wu, Skand~Vishwanath Peri, Weihao Sun, Gautam Singh, Fei Deng, Jindong Jiang, and Sungjin Ahn.
\newblock Space: Unsupervised object-oriented scene representation via spatial attention and decomposition.
\newblock {\em arXiv preprint arXiv:2001.02407}, 2020.

\bibitem{locatello2019challenging}
Francesco Locatello, Stefan Bauer, Mario Lucic, Gunnar Raetsch, Sylvain Gelly, Bernhard Sch{\"o}lkopf, and Olivier Bachem.
\newblock Challenging common assumptions in the unsupervised learning of disentangled representations.
\newblock In {\em international conference on machine learning}, pages 4114--4124. PMLR, 2019.

\bibitem{locatello2020weakly}
Francesco Locatello, Ben Poole, Gunnar R{\"a}tsch, Bernhard Sch{\"o}lkopf, Olivier Bachem, and Michael Tschannen.
\newblock Weakly-supervised disentanglement without compromises.
\newblock In {\em International Conference on Machine Learning}, pages 6348--6359. PMLR, 2020.

\bibitem{locatello2020object}
Francesco Locatello, Dirk Weissenborn, Thomas Unterthiner, Aravindh Mahendran, Georg Heigold, Jakob Uszkoreit, Alexey Dosovitskiy, and Thomas Kipf.
\newblock Object-centric learning with slot attention.
\newblock {\em Advances in Neural Information Processing Systems}, 33:11525--11538, 2020.

\bibitem{lowe2024rotating}
Sindy L{\"o}we, Phillip Lippe, Francesco Locatello, and Max Welling.
\newblock Rotating features for object discovery.
\newblock {\em Advances in Neural Information Processing Systems}, 36, 2024.

\bibitem{mansouri2022object}
Amin Mansouri, Jason Hartford, Kartik Ahuja, and Yoshua Bengio.
\newblock Object-centric causal representation learning.
\newblock In {\em NeurIPS 2022 Workshop on Symmetry and Geometry in Neural Representations}, 2022.

\bibitem{mansouri2023object}
Amin Mansouri, Jason Hartford, Yan Zhang, and Yoshua Bengio.
\newblock Object-centric architectures enable efficient causal representation learning.
\newblock {\em arXiv preprint arXiv:2310.19054}, 2023.

\bibitem{marcus2003algebraic}
Gary~F Marcus.
\newblock {\em The algebraic mind: Integrating connectionism and cognitive science}.
\newblock MIT press, 2003.

\bibitem{marino2018iterative}
Joe Marino, Yisong Yue, and Stephan Mandt.
\newblock Iterative amortized inference.
\newblock In {\em International Conference on Machine Learning}, pages 3403--3412. PMLR, 2018.

\bibitem{pmlr-v97-mathieu19a}
Emile Mathieu, Tom Rainforth, N~Siddharth, and Yee~Whye Teh.
\newblock Disentangling disentanglement in variational autoencoders.
\newblock In {\em Proceedings of the 36th International Conference on Machine Learning}, 2019.

\bibitem{10.5555/525544}
Radford~M. Neal.
\newblock {\em Bayesian Learning for Neural Networks}.
\newblock Springer-Verlag, Berlin, Heidelberg, 1996.

\bibitem{ribeiro2022learning}
Fabio De~Sousa Ribeiro, Kevin Duarte, Miles Everett, Georgios Leontidis, and Mubarak Shah.
\newblock Learning with capsules: A survey.
\newblock {\em arXiv preprint arXiv:2206.02664}, 2022.

\bibitem{ribeiro2020capsule}
Fabio De~Sousa Ribeiro, Georgios Leontidis, and Stefanos Kollias.
\newblock Capsule routing via variational bayes.
\newblock In {\em Proceedings of the AAAI Conference on Artificial Intelligence}, volume~34, pages 3749--3756, 2020.

\bibitem{NEURIPS2020_47fd3c87}
Fabio De~Sousa Ribeiro, Georgios Leontidis, and Stefanos Kollias.
\newblock Introducing routing uncertainty in capsule networks.
\newblock In {\em Advances in Neural Information Processing Systems}, volume~33, pages 6490--6502, 2020.

\bibitem{rock1973orientation}
Irvin Rock.
\newblock Orientation and form.
\newblock 1973.

\bibitem{scholkopf2022statistical}
Bernhard Sch{\"o}lkopf and Julius von K{\"u}gelgen.
\newblock From statistical to causal learning.
\newblock {\em Proceedings of the International Congress of Mathematicians}, 2022.

\bibitem{seitzer2022bridging}
Maximilian Seitzer, Max Horn, Andrii Zadaianchuk, Dominik Zietlow, Tianjun Xiao, Carl-Johann Simon-Gabriel, Tong He, Zheng Zhang, Bernhard Sch{\"o}lkopf, Thomas Brox, et~al.
\newblock Bridging the gap to real-world object-centric learning.
\newblock {\em International Conference on Learning Representations}, 2023.

\bibitem{singh2021illiterate}
Gautam Singh, Fei Deng, and Sungjin Ahn.
\newblock Illiterate dall-e learns to compose.
\newblock {\em International Conference on Learning Representations}, 2022.

\bibitem{singh2022neural}
Gautam Singh, Yeongbin Kim, and Sungjin Ahn.
\newblock Neural block-slot representations.
\newblock {\em arXiv preprint arXiv:2211.01177}, 2022.

\bibitem{singh2022simple}
Gautam Singh, Yi-Fu Wu, and Sungjin Ahn.
\newblock Simple unsupervised object-centric learning for complex and naturalistic videos.
\newblock {\em Advances in Neural Information Processing Systems}, 35:18181--18196, 2022.

\bibitem{tenenbaum2011grow}
Joshua~B Tenenbaum, Charles Kemp, Thomas~L Griffiths, and Noah~D Goodman.
\newblock How to grow a mind: Statistics, structure, and abstraction.
\newblock {\em science}, 331(6022):1279--1285, 2011.

\bibitem{tomczak2018vae}
Jakub Tomczak and Max Welling.
\newblock Vae with a vampprior.
\newblock In {\em International Conference on Artificial Intelligence and Statistics}, pages 1214--1223. PMLR, 2018.

\bibitem{van2017neural}
Aaron Van Den~Oord, Oriol Vinyals, et~al.
\newblock Neural discrete representation learning.
\newblock {\em Advances in neural information processing systems}, 30, 2017.

\bibitem{van2020investigating}
Sjoerd Van~Steenkiste, Karol Kurach, J{\"u}rgen Schmidhuber, and Sylvain Gelly.
\newblock Investigating object compositionality in generative adversarial networks.
\newblock {\em Neural Networks}, 130:309--325, 2020.

\bibitem{vaswani2017attention}
Ashish Vaswani, Noam Shazeer, Niki Parmar, Jakob Uszkoreit, Llion Jones, Aidan~N Gomez, {\L}ukasz Kaiser, and Illia Polosukhin.
\newblock Attention is all you need.
\newblock {\em Advances in neural information processing systems}, 30, 2017.

\bibitem{von2021self}
Julius Von~K{\"u}gelgen, Yash Sharma, Luigi Gresele, Wieland Brendel, Bernhard Sch{\"o}lkopf, Michel Besserve, and Francesco Locatello.
\newblock Self-supervised learning with data augmentations provably isolates content from style.
\newblock {\em Advances in neural information processing systems}, 34:16451--16467, 2021.

\bibitem{wang2023slot}
Yanbo Wang, Letao Liu, and Justin Dauwels.
\newblock Slot-vae: Object-centric scene generation with slot attention.
\newblock {\em arXiv preprint arXiv:2306.06997}, 2023.

\bibitem{pmlr-v202-wang23r}
Yanbo Wang, Letao Liu, and Justin Dauwels.
\newblock Slot-{VAE}: Object-centric scene generation with slot attention.
\newblock In Andreas Krause, Emma Brunskill, Kyunghyun Cho, Barbara Engelhardt, Sivan Sabato, and Jonathan Scarlett, editors, {\em Proceedings of the 40th International Conference on Machine Learning}, volume 202 of {\em Proceedings of Machine Learning Research}, pages 36020--36035. PMLR, 23--29 Jul 2023.

\bibitem{spriteworld19}
Nicholas Watters, Loic Matthey, Sebastian Borgeaud, Rishabh Kabra, and Alexander Lerchner.
\newblock Spriteworld: A flexible, configurable reinforcement learning environment.
\newblock https://github.com/deepmind/spriteworld/, 2019.

\bibitem{watters2019spatial}
Nicholas Watters, Loic Matthey, Christopher~P Burgess, and Alexander Lerchner.
\newblock Spatial broadcast decoder: A simple architecture for learning disentangled representations in vaes.
\newblock {\em arXiv preprint arXiv:1901.07017}, 2019.

\bibitem{willetts2021don}
Matthew Willetts and Brooks Paige.
\newblock I don't need u: Identifiable non-linear ica without side information.
\newblock {\em arXiv preprint arXiv:2106.05238}, 2021.

\bibitem{yang2022nonlinear}
Xiaojiang Yang, Yi~Wang, Jiacheng Sun, Xing Zhang, Shifeng Zhang, Zhenguo Li, and Junchi Yan.
\newblock Nonlinear {ICA} using volume-preserving transformations.
\newblock In {\em International Conference on Learning Representations}, 2022.

\bibitem{yao2024multi}
Dingling Yao, Danru Xu, S{\'e}bastien Lachapelle, Sara Magliacane, Perouz Taslakian, Georg Martius, Julius von K{\"u}gelgen, and Francesco Locatello.
\newblock Multi-view causal representation learning with partial observability.
\newblock 2024.

\bibitem{yuille2006vision}
Alan Yuille and Daniel Kersten.
\newblock Vision as bayesian inference: analysis by synthesis?
\newblock {\em Trends in cognitive sciences}, 10(7):301--308, 2006.

\bibitem{zimmermann2021contrastive}
Roland~S Zimmermann, Yash Sharma, Steffen Schneider, Matthias Bethge, and Wieland Brendel.
\newblock Contrastive learning inverts the data generating process.
\newblock In {\em International Conference on Machine Learning}, pages 12979--12990. PMLR, 2021.

\end{thebibliography}
